\documentclass[journal]{IEEEtran}
\usepackage{amsmath,amsfonts}
\usepackage{algorithm}
\usepackage{array}
\usepackage[caption=false,font=normalsize,labelfont=sf,textfont=sf]{subfig}
\usepackage{textcomp}
\usepackage{stfloats}
\usepackage{url}
\usepackage{verbatim}
\usepackage{graphicx}
\usepackage{cite}
\usepackage{multicol}
\usepackage{multirow}
\usepackage{amsmath}
\usepackage{amssymb}
\usepackage{dsfont}
\usepackage{bbm}

\usepackage{graphicx}
\usepackage{amsfonts}
\usepackage{url}
\usepackage{array}
\usepackage{caption}
\usepackage{booktabs}
\usepackage{cite}
\usepackage{caption}
\usepackage{algorithm}
\usepackage{algpseudocode}
\usepackage{multicol}
\usepackage{multirow}
\usepackage{tabularx}
\usepackage{color}
\usepackage{rotating}
\usepackage{makecell}
\usepackage{colortbl}
\usepackage{ragged2e}
\usepackage[numbers, square, sort&compress]{natbib}
\usepackage{bbm}
\usepackage{bm}
\usepackage{float}  

\begin{document}

\title {Arbitrarily Shaped Scene Text Detection: A Decade of Advances and Systematic Analysis}
\author{Pengwen Dai, Feiyang He, Chaolang Li, Xugong Qin, Wenqi Ren, Xiaochun Cao,~\IEEEmembership{Senior Member,~IEEE}
\thanks{Pengwen Dai, Feiyang He, Chaolang Li, Wenqi Ren, and Xiaochun Cao are with the School of Cyber Science and Technology, Shenzhen Campus of Sun Yat-sen University, Shenzhen 518107, China(Email:daipw@mail.sysu.edu.cn,hefy8@mail2.sysu.edu.cn,lichlang5@\\-mail2.sysu.edu.cn,renwq3@mail.sysu.edu.cn,caoxiaochun@mail.sysu.edu.cn).} 
\thanks{Xugong Qin is with the School of Cyber Science and Engineering, Nanjing
University of Science and Technology, Nanjing 210094, China (e-mail:
qinxugong@njust.edu.cn).}
}

\markboth{Journal of \LaTeX\ Class Files,~Vol.~14, No.~8, August~2021}%
{Shell \MakeLowercase{\textit{et al.}}: A Sample Article Using IEEEtran.cls for IEEE Journals}


\maketitle

\begin{abstract}
Scene text detection has become an important research area in computer vision. However, dynamic changes in scenes and the complex diversity of text appearances make accurate scene text detection highly challenging. Although numerous arbitrary-shaped scene text detection methods have been proposed in recent years, with most claiming state-of-the-art performance, these performance comparisons are often unfair due to various inconsistent settings (e.g., training data, backbones, multi-scale feature fusion, evaluation protocols, etc.). Such discrepancies tend to obscure the strengths and weaknesses of the core techniques being proposed, further hindering progress in the field. In this paper, we first review the development of scene text detection in the deep learning era, systematically tracing and summarizing the technical evolution of the field. Then, we carefully examine and analyze the aforementioned inconsistent settings and propose unified frameworks for bottom-up and top-down scene text detection methods, respectively. Under the unified frameworks, we keep the settings of non-core modules consistent and focus on exploring representations of arbitrary-shaped scene text, aiming to standardize future research and ensure fair comparisons. Finally, we discuss valuable future research directions, with the goal of inspiring subsequent researchers. As the first comprehensive survey dedicated to arbitrary-shaped scene text detection, this paper seeks to eliminate the barriers to performance comparison among existing methods through investigation and detailed analysis, to reveal the strengths and weaknesses of prior models under fair comparisons, and thereby better promote the development of the field.
\end{abstract}


\section{Introduction}
\IEEEPARstart{R}{eading} text in the natural image is a hot topic, due to its wide practical applications, e.g., fine-grained classification \cite{CVPR22/KMST/Wang}, image retrieval \cite{CVPR22/ViSTA/Cheng}, image caption \cite{ECCV20/TextCaps/Sidorov}, visual question answering \cite{CVPR20/EST_VQA/Wang},  etc. 
A scene text system can combine any scene text detectors \cite{TIP18/TextBoxes++/Liao,DBNet++,TNNLS23/KPN/Zhang} and any recognizers \cite{TPAMI17/CRNN/Shi,TPAMI19/ASTER/Shi,TNNLS24/GLaLT/Zhang} into the two-stage models, or integrate them into an end-to-end trainable framework \cite{TPAMI23/ABINet++/Fang,CVPR23/Deepsolo/Ye,IJCV25/SwinTextSpotterv2/Huang}.  
Scene text detection, which aims to locate text instances in images, has attracted increasing attention in the deep learning community as a prerequisite for the text reading system.
However, uneven illumination, perspective distortion, and complex backgrounds in natural scenes cause difficulties in detecting scene texts. Meanwhile, the specific characteristics of texts, e.g., various scales, diverse aspect ratios, different fonts, and arbitrary-shape layouts, also increase the challenge of scene text detection.

With the development of deep learning, scene text detection has achieved significant breakthroughs. Early deep learning methods adopted a multi-stage processing paradigm similar to traditional methods, consisting of four stages: character candidate extraction, character candidate filtering, text instance construction, and text instance filtering. In the character candidate extraction stage, they typically used connectivity-based techniques (such as Stroke Width Transform (SWT)\cite{CVPR10/SWT/Boris}, Maximally Stable Extremal Regions (MSER)\cite{IVC04/MSER/Matas}, Stroke Band Keypoints (SBK)\cite{CVPR15/FASText/Busta}, and their variants\cite{TMM11/SDTL/Bouman,ICCV13/SFT/Huang,TIP14/Characterness/Li,CVPR16/Canny} or sliding window techniques\cite{CVPR15/Symmetry/Zhang}. Unlike traditional methods that design classifiers \cite{ICDAR09/Hanif,TITS23/HFENet/Liang} based on handcrafted features using machine learning approaches (e.g., AdaBoost, HOG) for character candidate filtering or text instance filtering, deep learning-based methods \cite{ECCV14/CNN_MSER/Huang,ICCV15/Tian/TextFlow,TMM18/Tang/SSFT,ACMMM16/MOTD_VAM} employ CNNs or RNNs for feature learning and classification. However, these methods suffer from limited localization accuracy of local character regions in complex scenes, error accumulation due to multi-stage processing, and limited capability in modeling the geometric layout of scene text with complex shapes.
 
\begin{figure}[t]
\centering
\includegraphics[width=1.0\linewidth]{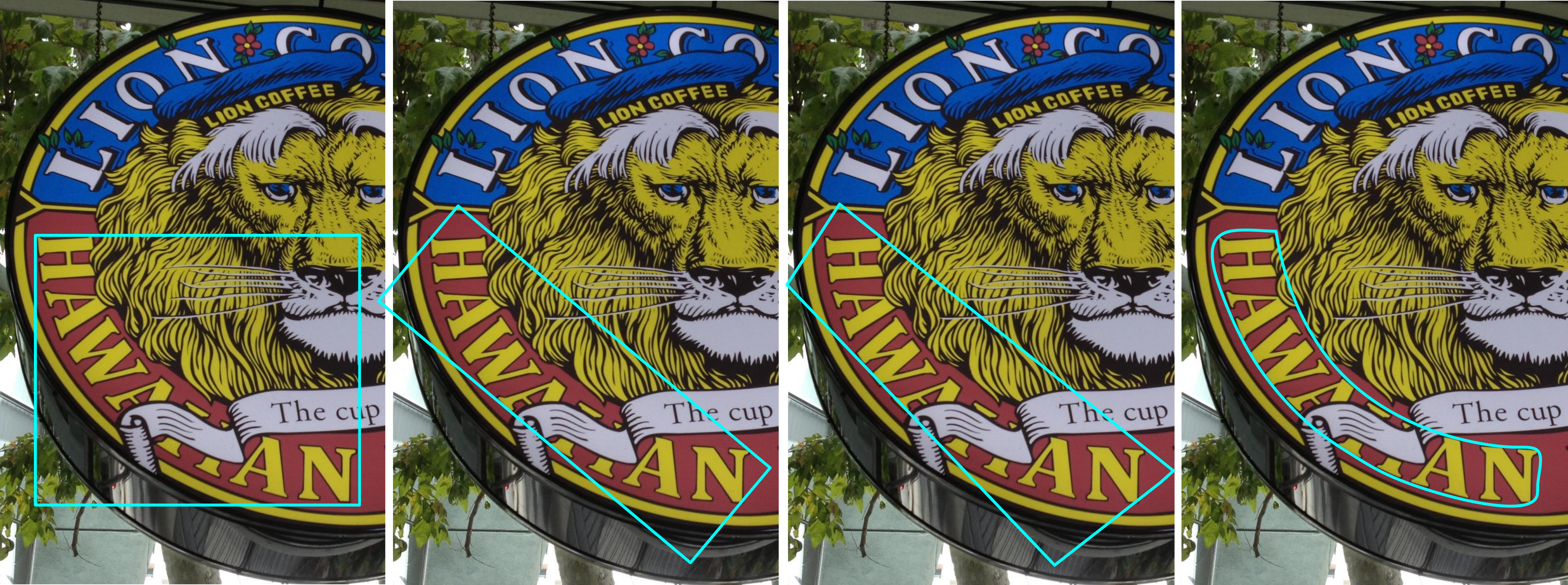}
\vspace{0.1cm} 
\setlength{\unitlength}{\linewidth}
\begin{picture}(1,0)
    \put(0.125,0){\makebox(0,0){\small(a)}}
    \put(0.375,0){\makebox(0,0){\small(b)}}
    \put(0.625,0){\makebox(0,0){\small(c)}}
    \put(0.875,0){\makebox(0,0){\small(d)}}
\end{picture}
\caption{Illustration of different representations for text instances. (a) Axis-aligned rectangle. (b) Rotated rectangle. (c) Quadrangle. (d) Arbitrary-shape contour.}
\label{fig:motivation}
\end{figure}

Unlike early CNN-based methods that were only used as classifiers, the mainstream approaches treat scene text as a special type of object, integrating classification and object location regression into a single deep learning model for multi-task learning. Existing methods \cite{ICDAR17/DSN/Wu,AAAI17/TextBox/Liao} have explored adopting axis-aligned rectangles, rotated rectangles, or quadrangles \cite{CVPR17/EAST/Zhou,ICCV17/DR/He,CVPR18/ITN/Wang,TIP18/TextBoxes++/Liao} to localize scene texts. Despite their great progress on horizontal or multi-oriented texts, these methods may fall short when handling arbitrary-shaped texts that are ubiquitous in real life, as shown in Figure \ref{fig:motivation} (a)(b)(c). To accurately localize arbitrary-shaped text contours in the natural image shown in Figure \ref{fig:motivation} (d), two types of methods are prevalent in the field of scene text detection. One is top-down based methods\cite{CVPR19/ATRR/Wang,PR19/CTD_CLOC/Liu,ECCV20/SDM/Xiao,TextBPN,PCR21,MM20/TextRay/Wang,AAAI24/Box2Poly/Chen}. They perform binary segmentation or the regression of arbitrary-shaped text contour points on the proposals. The other is bottom-up based methods \cite{CVPR19/CRAFT/Baek,ICCV19/PAN/Wang,ICCV19/TextDragon/Feng,CVPR20/DRRGN/Zhang,FSGNet,TextPMs,DBNet++,AAAI24/LRANet/Su,CVPR24/KACM/Zheng,AAAI25/ERRNet/Sun}. They first predict local units (\textit{e.g.}, pixels, connected components) and their auxiliary information, and then group them into different text instances.

However, it is difficult to determine whether and to what extent a newly proposed model shows improvement in performance and speed when compared with previous methods of the same type. This is because existing methods adopt different training settings and testing environments. Existing methods usually employ different backbone networks and utilize different external data for model pretraining. We also observe that different testing scales would distinctly influence the performance and speed of the model. Moreover, some methods even claim they have achieved state-of-the-art based on the reported performance using different evaluation protocols. These inconsistencies hinder fair comparisons between the proposed core techniques of existing scene text detection methods. Although some scholars \cite{ICCV19/textRec_fairComp/Baek} have made deep analyses and fair comparisons for scene text recognition models, the proposed strategies for scene text recognition are not suitable for scene text detection due to the differences in tasks. Especially, the task of scene text detection involves more complicated inconsistencies.

Although several surveys on scene text analysis have been published, there is still a lack of a systematic and comprehensive review specifically devoted to arbitrary-shaped scene text detection. Existing surveys on scene text recognition, such as \cite{CSUR21/TextRecWild/Chen,JFCST24/NaturalTextRecDL/Fanzhi,SIST26/SegFreeTextRec/Deepak}, mainly focus on recognition models, datasets, and evaluation protocols, and provide limited discussion of scene text detection. Some broader surveys, such as \cite{TPAMI15/TextDetRecSurvey/Ye,FCS16/TextDetRecAdv/Zhu,IJCV21/TextDetRecDeep/Long,TIP16/TextDetRecVideo/Author,TPAMI26/handwritten}, review scene text detection and recognition from a task-level perspective, but their discussions of detection are relatively general and do not specifically investigate arbitrary-shaped scene text detection. In addition, for surveys dedicated to scene text detection, early reviews such as \cite{IJCA15/TextLocSurvey/Chavre} and \cite{MLICOM16/TextDetSurvey/Wang} mainly focus on traditional scene text localization or detection pipelines, while more recent works such as \cite{AIR21/DeepTextDetReview/Khan} and \cite{MTAP24/MSERDetSurvey/Dutta} summarize deep learning-based scene text detection methods or specific technical lines such as MSER-based approaches. However, there is still no review that provides an in-depth investigation specifically focused on arbitrarily-shaped scene text detection.

To that end, we present a summary of prior surveys in Table \ref{tab:str_surveys} and systematically outline what sets our paper apart, emphasizing its distinct contributions. In this survey, we first provide a specialized and focused examination of arbitrary-shaped scene text detection, systematically reviewing end-to-end scene text detection methods in the deep learning era, from horizontal boxes to rotated boxes and then to arbitrary-shaped boundary boxes, and categorizing them based on learning paradigms. Second, to reveal the strengths and weaknesses of core techniques in existing methods, we construct unified frameworks for the two current mainstream paradigms of arbitrary-shaped scene text detection, i.e., top-down and bottom-up approaches, providing a common perspective for existing methods. Within this framework, we comprehensively uncover various inconsistencies that lead to unfair comparisons among arbitrary-shaped scene text detection methods. Furthermore, based on standardizing the experimental settings of non-core technical modules (e.g., identical image preprocessing, backbone networks, etc.), we focus on investigating the impact of core techniques (e.g., feature fusion learning, prediction networks) to elucidate the limitations and applicability boundaries of performance improvements in arbitrary-shaped scene text detection algorithms. Finally, we discuss several overlooked explorations and challenges in arbitrary-shaped scene text detection, offering richer guidance for future research.

\begin{table*}[t]
\caption{A summary of representative surveys on scene text detection and recognition}
\centering
\setlength{\tabcolsep}{2.6pt}
\renewcommand{\arraystretch}{1.08}
\scriptsize
\begin{tabular}{|>{\centering\arraybackslash}m{1.45cm}|
                >{\centering\arraybackslash}m{0.8cm}|
                >{\centering\arraybackslash}m{1.1cm}|
                >{\RaggedRight\arraybackslash}m{5.0cm}|
                >{\RaggedRight\arraybackslash}m{7.5cm}|}
\hline
Topic & Year & Venue & Title & Main Content \\ \hline

\multirow{4}{*}{Detection}
& 2015 & IJCA
& A Survey on Text Localization Method in Natural Scene Image\cite{IJCA15/TextLocSurvey/Chavre}
& This survey reviews text localization in natural scenes and its role in text extraction. However, it mainly focuses on traditional methods and gives limited benchmark analysis. \\ \cline{2-5}

& 2016 & MLICOM
& Text Detection in Natural Scene Image: A Survey\cite{MLICOM16/TextDetSurvey/Wang}
& This paper surveys scene text detection methods and compares different detection stages. However, it emphasizes taxonomy more than deep models and large-scale evaluations. \\ \cline{2-5}

& 2021 & AIR
& Deep Learning Approaches to Scene Text Detection: A Comprehensive Review\cite{AIR21/DeepTextDetReview/Khan}
& This review summarizes deep learning based scene text detection and introduces datasets and evaluation protocols. However, it gives less discussion of downstream recognition and tampering related tasks. \\ \cline{2-5}

& 2024 & MTAP
& Natural Scene Text Localization and Detection Using MSER and Its Variants\cite{MTAP24/MSERDetSurvey/Dutta}
& This survey focuses on MSER-based text localization and detection in complex scenes. However, its scope is relatively narrow and less connected to broader deep learning frameworks. \\ \hline

\multirow{3}{*}{Recognition}

& 2021 & CSUR
& Text Recognition in the Wild: A Survey\cite{CSUR21/TextRecWild/Chen}
& This survey summarizes key problems, representative methods, and resources for scene text recognition. However, it mainly focuses on recognition itself and less on joint modeling with detection. \\ \cline{2-5}

& 2024 & JFCST
& Survey on Natural Scene Text Recognition Methods of Deep Learning\cite{JFCST24/NaturalTextRecDL/Fanzhi}
& This paper reviews deep learning methods for scene text recognition and summarizes datasets and metrics. However, it is recognition-oriented and offers less cross-task analysis. \\ \cline{2-5}

& 2026 & SIST
& A Review of Deep Learning-Based Segmentation- Free Scene Text Recognition Methods\cite{SIST26/SegFreeTextRec/Deepak}
& This review focuses on segmentation-free scene text recognition and summarizes representative methods and results. However, it is relatively specialized and less comprehensive on the full scene text pipeline. \\ \hline

\multirow{4}{*}{Det.\ \& Rec.}
& 2015 & TPAMI
& Text Detection and Recognition in Imagery: A Survey\cite{TPAMI15/TextDetRecSurvey/Ye}
& This survey broadly reviews text detection and recognition in imagery and categorizes major systems. However, it predates modern deep learning advances and thus has limited coverage of newer benchmarks. \\ \cline{2-5}

& 2016 & FCS
& Scene Text Detection and Recognition: Recent Advances and Future Trends\cite{FCS16/TextDetRecAdv/Zhu}
& This review summarizes recent advances, resources, and future directions in scene text analysis. However, it is more transitional and provides limited discussion of later end-to-end models. \\ \cline{2-5}

& 2016 & TIP
& Text Detection, Tracking and Recognition in Video: A Comprehensive Survey\cite{TIP16/TextDetRecVideo/Author}
& This paper surveys text detection, tracking, and recognition in video and compares representative methods. However, it is video-oriented and less focused on still-image scene text benchmarks. \\ \cline{2-5}

& 2021 & IJCV
& Scene Text Detection and Recognition: The Deep Learning Era\cite{IJCV21/TextDetRecDeep/Long}
& This survey reviews major progress in scene text detection and recognition during the deep learning era. However, it discusses little about text tampering detection and authenticity analysis. \\ \hline

\end{tabular}
\label{tab:str_surveys}
\end{table*}

The main contributions of this paper are summarized as follows: 
\begin{itemize}
\item This review provides a systematic and comprehensive overview of end-to-end arbitrary-shape scene text detection methods in the deep learning era, serving as a valuable entry resource for beginners in the field and a navigation guide for established researchers to help them keep pace with the latest developments.

\item To the best of our knowledge, we are the first to reveal the limitations of the claimed state-of-the-art performance and the inconsistencies in experimental settings among existing arbitrary-shape scene text detection methods, helping researchers focus on breakthroughs in core technologies and promote the advancement of the field.

\item We establish unified frameworks for arbitrary-shape scene text detection to standardize future research and ensure fair and just comparisons.

\item We summarize the current bottlenecks in technological development and provide valuable directions for future exploration, which can serve as a reference for subsequent researchers.
\end{itemize}

The rest of the paper is structured as follows: Section \ref{sec:review} provides a classification and systematic review of end-to-end arbitrary-shaped scene text detection methods based on deep learning in a decade. Section \ref{sec:incons_ana} presents a fine-grained inconsistency analysis of current mainstream methods. Then, Section \ref{sec:framework} constructs two types of mainstream unified detection frameworks and provides detailed definitions. Meanwhile, Section \ref{sec:experiments} conducts extensive experiments under the unified framework to investigate the impact of non-core technical modules on performance. Section \ref{sec:directions} discusses future research directions. Finally, Section \ref{sec:conclusions} concludes this paper.

\section{Review of Related Work} \label{sec:review}
Currently, scene text detectors based on deep learning can be roughly divided into two categories: top-down scene text detectors and bottom-up scene text detectors.

\subsection{Top-down arbitrary-shape scene text detectors}
\begin{figure*}[t]
  \begin{center}
    \includegraphics[width=1.0\linewidth,height=0.29\linewidth]{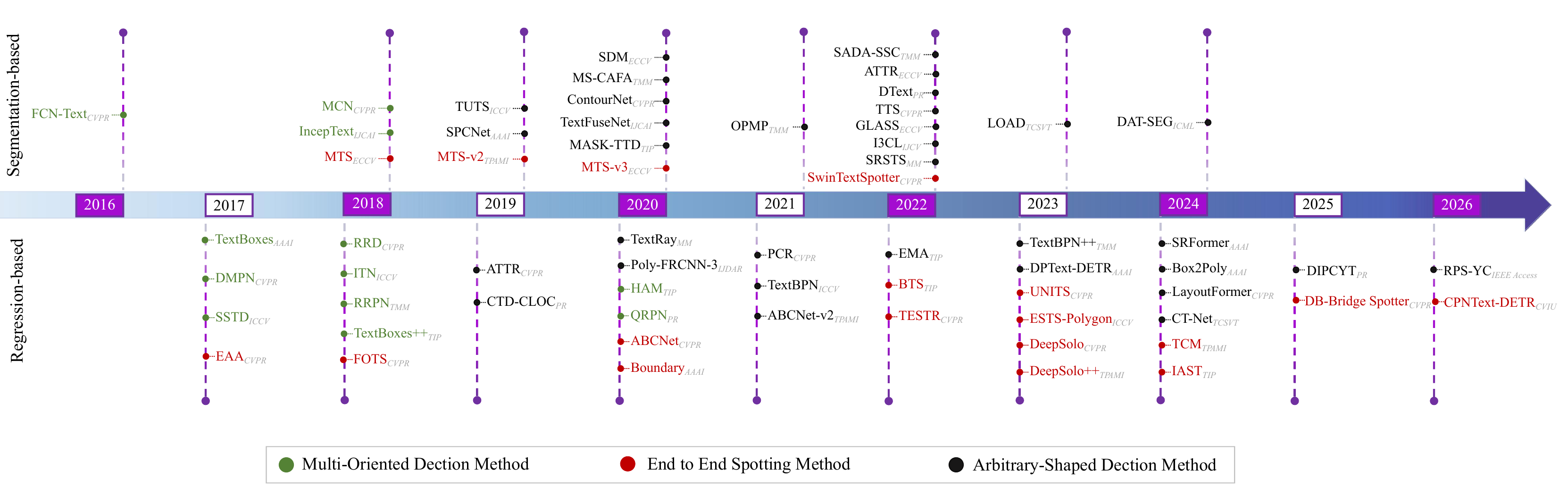}
  \end{center}
  \caption{Evolution of top-down based arbitrary-shape scene text detectors.}
  \label{fig:timeline_top-down}
  \vspace{-0.2cm}
\end{figure*}

Top-down scene text detection methods treat the text in a natural image as an object and extend general object detection technologies to localize the entire text instance. They segment the text instance or regress the key points based on candidate bounding boxes. As shown in Fig. \ref{fig:timeline_top-down}, the end-to-end top-down scene text detection method has achieved great progress in the past decade.

\subsubsection{\textbf{Segmentation-based Methods}}
Segmentation-based methods treat each pixel within a text region or its contracted region as foreground and formulate the task as a semantic segmentation problem to distinguish foreground text from complex backgrounds. 

In the early stage of deep learning, Scholars focused on detecting horizontal or multi-oriented scene text. FCN-Text \cite{CVPR16/FCN_Text/Zhang} directly segments text regions using a Fully Convolutional Network (FCN) \cite{CVPR15/FCN/Shelhamer} originally designed for object-level semantic segmentation, and within the segmented regions, the MSER algorithm is applied to extract character locations based on an inclined centerline. Differently, to robustly separate text instances from semantic segmentation regions, MCN \cite{CVPR18/MCN/Liu} converts an image into a Stochastic Flow Graph and then performs Markov Clustering on that graph. Inspired by the general object instance segmentation framework FCIS \cite{CVPR17/FCIS/Li}, IncepText \cite{IJCAI18/IncepText/Yang} designs an Inception-Text module to enhance the fused multi-scale features, and employs deformable position-sensitive ROI pooling to project arbitrary-orientation proposals to fixed-scale features. 

To localize arbitrary-shaped scene text, Total-Text \cite{ICDAR17/ToT} first directly extracts connected components of text by applying thresholding to the segmentation map. Promising performances are achieved based on the framework of instance segmentation, e.g., MaskRCNN \cite{ICCV17/Mask_RCNN/He}. 
SPCNet \cite{AAAI19/SPCN/Xie} designs a text context module to enrich the feature representation by predicting the mask to guide the feature learning in the process of pyramid feature fusion. MS-CAFA \cite{TMM19/MS-CAFA/Dai} exploits a pyramid Region-of-Interest (ROI) pooling attention mechanism to learn robust features for proposals with various scales. Mask-TTD \cite{TIP20/Mask-TTD/Liu} adopts a tightness prior to adjusting text proposals for better covering the entire text region, and utilizes the text frontier information to improve the text mask prediction. TextFuseNet \cite{IJCAI20/TextFuseNet/Ye} further introduces character-, word-, and global-level feature learning, and proposes a hierarchical fusion mechanism to perceive texts. Differently, SDM \cite{ECCV20/SDM/Xiao} designs a sequential deformation network to model the line-shape of the text instance, supervised by learning character counting. OPMP \cite{TMM21/OPMP/Zhang} proposes to generate arbitrary-shape proposals by a pyramid lengthwise and sidewise residual sequence modeling mechanism, followed by a multi-grain text classification subnetwork.  To achieve fracture detections among text instances at the gaps, I3CL \cite{IJCV22/I3CL/Du} obtains the discriminative representations for text instances via the intra-instance and inter-instance collaborative learning. Moreover, DText \cite{PR22/DText/Cai} employs dynamic convolution to guide the learning of mask prediction. LOAD \cite{TCSVT23/LOAD/Cheng} exploits an adaptive threshold module to filter the candidates and binarize the mask.

Instead of segmenting the entire text regions,  ContourNet \cite{CVPR20/ContourNet/Wang} also exploits to perform the segmentation for text contours by aggregating the features in horizontal and vertical orthogonal directions, based on the scale-insensitive adaptive proposals. CSE \cite {CVPR19/CSE/Liu} expands a seed point, randomly initialized within a candidate text region, into an arbitrarily-shaped text region. Besides, SADA-SSC \cite{TMM22/SADA_SSC/Dai} focuses on the scale-aware training sample online augmentation, achieving better model convergence and robust performance. Inspired by the general object detection based on the Transformer framework, e.g., DETR \cite{ECCV20/DETR/Carion}, some scholars also adapt it for arbitrary-shape scene text detection. ATTR \cite{ECCV22/ATTR/Zhou} utilizes the Transformer framework to aggregate multi-scale features and decode into a binary mask for each query guided by the text embedding. While DAT-SEG \cite{ICML24/DAT/Wan} not only introduces a cross-granularity interactive attention mechanism to enhance the learning capability for text instances at different granularities, but also builds a prompt-based segmentation module to further refine the detection results for text in arbitrary shapes and complex layouts.

In addition, some scholars integrate recognition networks on the basis of detection models to enhance the capability of detecting arbitrarily shaped scene text through collaborative learning. Built on MaskRCNN, Mask-TextSpotter (MTS)\cite{ECCV18/Mask_TextSpotter/Lyu} integrates a character-level segmentation branch in parallel with the text region segmentation to achieve recognition. To alleviate the constraint of relying on character position annotations for training and heuristic rules for grouping characters, Mask-TextSpotter-v2(MTS-v2) \cite{TPAMI19/Mask_TextSpotter++/Liao} utilizes a spatial attention module in the recognition part. Moreover, Mask-TextSpotter-v3(MTS-v3) \cite{ECCV20/MaskTextSpotterv3/Liao} designs a  Segmentation Proposal Network (SPN) instead of the frequently-used Region Proposal Network (RPN). Differently, SRSTS \cite{MM22/SRTS/Wu} predicts the anchor points to guide the detection and the recognition, mitigating the reliance on detection box accuracy for character-level recognition during the inference stage. Instead of performing character-level segmentation to achieve recognition, TUTS \cite{ICCV19/TUTS/Qin} performs the mask-level ROI pooling for the text instances, before being fed into a sequential recognition model. GLASS \cite{ronen2022glass} proposes a global-to-local
attention mechanism before the recognizer, greatly reduces the inherent difficulties associated with scale and word rotation. SwinTextSpotter \cite{SwinTextSpotter} and its extension SwinTextSpotterv2 \cite{IJCV25/SwinTextSpotterv2/Huang} employ a transformer encoder equipped with a dynamic head as the detector. To alleviate the expensive human labor of annotating arbitrary-shaped text locations in scene images,  TTS \cite{TTS22cvpr} constructs a Transformer-based framework to support both fully-supervised and weakly-supervised learning strategies, which learns a single latent representation for each word detection and designs a novel loss function based on the Hungarian loss.

\subsubsection{\textbf{Regression-based Methods}}
In the era of deep learning, early regression-based methods inherited the framework of general object detection and then made specialized designs for the characteristics of scene text. 

For horizontal bounding boxes, TextBox \cite{AAAI17/TextBox/Liao}, built upon the SSD \cite{ECCV16/SSD/Liu} framework for general object detection, utilizes rectangular convolutional kernels during feature learning to capture wider and taller text region features. FEN \cite{AAAI18/FEN/Zhang} utilizes a multi-level feature fusion strategy, adaptively weighted position-sensitive ROI pooling, and the mining strategy of positive proposals. 
To localize multi-oriented scene text, also built upon the framework of SSD, DMPN \cite{CVPR17/DMPN/Liu} designs the quadrilateral sliding windows for better regression. While SSTD \cite{ICCV17/SSD_RA/HePan} uses the global semantic segmentation as an attention map to assist the bounding box regression and classification. TextBoxes++ \cite{TIP18/TextBoxes++/Liao}  extends TextBox to detect oriented bounding boxes by optimizing the network architecture and training process. Similarly, built upon the generation object detection framework of Faster-RCNN, RRPN \cite{TMM17/RRPN/Ma} focuses on designing a rotated region proposal network, while QRPN \cite{PR20/QRPN/Wang} mainly designs the quadrilateral region proposal network to generate quadrilateral proposals. Instead, ITN \cite{CVPR18/ITN/Wang} directly predicts the rotation angle to rectify the features. While RRD \cite{CVPR18/RRD/Liao} utilizes the rotated convolution kernels to learn the feature representations of multi-oriented scene texts. To not suffer from the careful design of the anchors, HAM \cite{TIP20/HAM/Hou} develops a decoupled anchor mechanism that adaptively accommodates the variability of text bounding boxes and reduces the anchor count in the matching stage.

To regress the boundary of an arbitrarily-shaped scene text bounding box,  CTD-CLOC \cite{PR19/CTD_CLOC/Liu} predicts the offsets of key points to the top-left points, and utilizes the Long Short-Term Memory (LSTM) to smooth the offsets. While Poly-FRCNN \cite{IJDAR20/TOT/chng} regresses fixed key points based on the horizontal bounding boxes. Instead of the static regression, ATRR \cite{CVPR19/ATRR/Wang}  adaptively outputs the point pair using LSTM.  To avoid the labor-intensive design of anchors, TextRay performs top-down contour-based polar geometric modeling and geometric parameter learning within a single-stage anchor-free framework. While PCR \cite{PCR21} adopts an anchor-free framework to obtain a small number of horizontal bounding boxes, and then progressively evolves the key points of such horizontal candidates to the points along arbitrary text contours by aggregating contour information.  With a better front-end arbitrary-shape boundary initialization, TextBPN \cite{TextBPN} utilizes  Graph Convolutional Network (GCN) and a Recurrent Neural Network (RNN) to encode the boundary information, and then performs boundary deformation in an iterative way. TextBPN++ \cite{TextBPN++} introduces an energy minimization constraint, together with an energy monotonically decreasing constraint, to further optimize the boundary refinement learning while ensuring its stability, based on boundary transformers.
Similarly, CT-Net \cite{TCSVT24/CT-Net/Shao} designs an adaptive training strategy that allows contour transformers to explore a wider range of deformation possibilities, along with a re-scoring mechanism that effectively reduces false positives.

Also inspired by the framework of DETR, SRFormer \cite{AAAI24/SRFormer/Bu} constructs a mask-informed query enhancement module in the decoder stage of the Transformer architecture, which provides rich instance queries for subsequent progressive regression refinement. DPText-DETR \cite{AAAI23/DPText-DETR/Ye} constructs spatial position queries directly based on explicit point coordinates and evolves them dynamically through progressive updates by introducing a circular-shape guided factorized self-attention mechanism.  Instead of utilizing dense initial proposals and point-level feature embedding for individual vertices, Box2Poly \cite{AAAI24/Box2Poly/Chen}  adopts sparse proposal initialization to ease computational load and minimize memory consumption, and learns the instance-level feature embeddings for accurately predicting polygons via a cascade decoding pipeline and the PolyAlign layer. LayoutFormer \cite{CVPR24/LayoutFormer/Liang} introduces a framework for joint multi-level text detection and geometric layout prediction, which updates text queries using a few representative foreground features. Differently, DIPCYT \cite{PR24/DIPCYT/Roy} focuses on constructing the generalized arbitrary-shape scene text detection model in multiple domains based on the Transformer framework. EMA \cite{EMA} adopts a mixed supervision to train the model.

With the help of end-to-end scene text spotting learning, ABCNet \cite{CVPR20/ABCNet/Liu}, ABCNetv2 \cite{TPAMI21/ABCNetV2/Liu}, and FSGNet \cite{FSGNet}  adaptively fit the boundaries of arbitrarily-shaped text bounding boxes by predicting the control points of a Bezier curve, and design a BezierAlign layer to bridge the gap between detection and recognition. Boundary \cite{AAAI20/AYNIB/Wang} proposes a two-stage boundary point regression method by first finding the minimum oriented bounding box of each text instance, followed by predicting points within that box, while its extension Boundary-TextSpotter(BTS) \cite{TIP22/Boundary-TextSpotter/Lu} adopts a single-stage detection model to directly regress boundary points in an iterative way. TESTR \cite{TESTR} proposes a canonical representation of control points that is applicable to text instances annotated with both Bezier curves and polygons. ESTextSpotter \cite{ESTextSpotter} decomposes the conventional queries into task-aware queries for text polygon and content, which explicitly interact with each other while preserving discriminative patterns of text detection and recognition by a vision-language communication mechanism. DeepSoLo \cite{deepsolo} and DeepSoLo++ \cite{DeepSolo++} represent the character sequence of each text instance as ordered points and model them with learnable explicit point queries, simultaneously obtaining boundary points and text content through a single decoder. IAST \cite{TIP24/IAST/Zhang} divides the initial text boundary into two symmetric control points and iteratively refines the new text boundary using a lightweight boundary refinement technique integrated with a recurrent convolutional network. Differently, UNITS \cite{UNITS}  unifies arbitrary-shaped text detection and recognition into a sequence generation model. DG-BridgeTextSpotter \cite{CVPR25/DG-Brigde-Spotter/Huang} constructs a bridging module that connects a parameter-fixed detector and recognizer through an adapter. While FastTCM \cite{TPAMI/TCM/Yu} uses the potential of the CLIP model to improve scene text detection and spotting tasks.

\subsection{Bottom-up Arbitrary-shape Scene Text Detectors}
\begin{figure*}[t]
  \begin{center}
    \includegraphics[width=1.0\linewidth,height=0.29\linewidth]{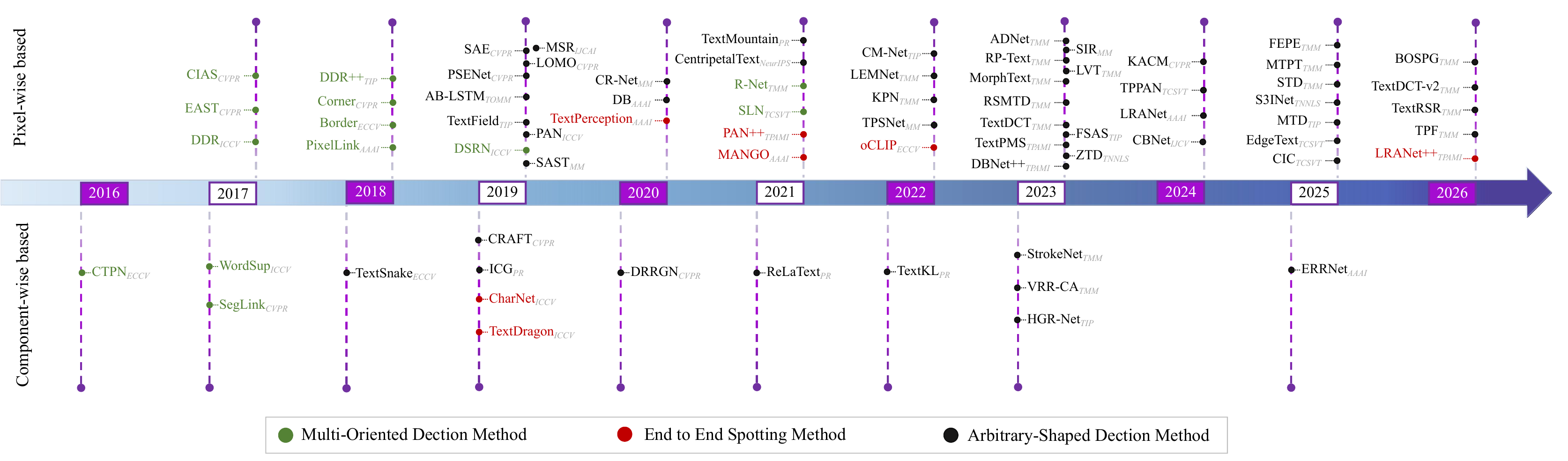}
  \end{center}
  \caption{Evolution of bottom-up arbitrary-shape scene text detection methods.}
  \label{fig:timeline_bottem-up}
  \vspace{-0.1cm}
\end{figure*}

The bottom-up scene text detection method first detects basic text components through the network and then aggregates the text components into a complete text instance through some post-processing methods. These detectors can be further divided into pixel-wise based methods and component-wise based methods. Fig. \ref{fig:timeline_bottem-up} illustrates that the past decade has witnessed substantial advancements in end-to-end bottom-up arbitrary-shape scene text detection methods.

\subsubsection{\textbf{Pixel-wise based Methods}}
These kinds of methods are based on instance segmentation and classify each pixel in the image to obtain a mask map of the foreground area. Then, the text pixels belonging to the same text are aggregated together through some pixel aggregation post-processing methods to obtain the final text instance bounding box.

Early methods focus on detecting multi-oriented scene texts. CIAS \cite{CVPR17/CIAS/He} leverages centerline segmentation to guide text instance segmentation. EAST \cite{CVPR17/EAST/Zhou} constructs rotated boxes or quadrilateral boxes by predicting, for each pixel within the shrunk scene text region, either the distances to the four sides of the bounding box or the offsets between the four corner points of the quadrilateral. While MOST \cite{CVPR21/MOST/He}, like EAST, obtains the rotated bounding boxes by adaptively adjusting the receptive fields of features. Based on the framework of EAST, to model the large variance of text scales, DSRN \cite{IJCAI19/DSRN/Wang} maps multi-scale convolutional features to a scale-invariant domain, resulting in equalized activation for text instances regardless of their sizes. While its extension R-Net \cite{TMM21/R-Net/Wang} introduces a scale transfer layer that projects large-scale feature maps onto a smaller scale, which has been shown to preserve the geometry of small-sized text effectively. While SLN \cite{TCSVT21/SLN/Cai} introduces a scale-residual learning approach that simultaneously eliminates concatenation-induced feature fusion residuals and up-sampling-induced scale transformation residuals. Similar to EAST, DDR \cite{ICCV17/DR/He} and its extension DDR++ \cite{TIP18/DDR++/He} also directly  predict the offsets from a pixel in the text region. Built on the direction regression network, PSS \cite{TCSVT20/PSS/Cheng} integrates the position-sensitive segmentation and pooling to learn the relative position information of various-scale texts. Instead of regressing distances or position offsets, PixelLink \cite{AAAI18/PixelLink/Deng} predicts the linkage relationships between each pixel and its eight surrounding pixels. Differently, Border \cite{ECCV18/Border/Xue} segments the four borders of each text instance, while  CornerPS \cite{CVPR18/cornerRS/Yao} predicts corner points and a position-sensitive segmentation map, and these corner points are used to sample and group into the candidate bounding boxes restored by the segmentation map during inference. 

To detect the arbitrary-shaped scene text, LOMO \cite{CVPR19/LOMO/Zhang} introduces an iterative refinement module and a shape expression module based on the DDR framework. By predicting text region kernels, PSENet \cite{CVPR19/PSENet/Wang} introduces a progressive scale expansion algorithm to fuse multi-shrunk.  Also, ADNet creates an adaptive dilation subnetwork to reconstruct the text region from a shrunk polygon. BOSPG predicts an instance map to facilitate rescoring. Besides, CentripetalText \cite{NeurIPS21/CentripetalText/Sheng} estimates the centripetal shift map, and  RSMTD \cite{TMM23/RSMTD/Yang} learns the reinforcement offset map and super-pixel window map. Moreover, PAN \cite{ICCV19/PAN/Wang} estimates the embedding vectors of pixels to measure the distances to different text kernels. RP-Net \cite{TMM23/RP-Text/Wang}, built on PAN, introduces a region context module and  a progressive fusion module to enrich the representations. SAE \cite{CVPR19/SAE/Tian} transforms pixels into an embedding space that promotes proximity among pixels from the same text. A prior constraint is introduced to guide the boundary prediction in a high-dimensional embedding space \cite{TMM21/LEMNet/Xing}. Besides, MSR \cite{IJCAI19/MSR/Xue} and TextField \cite{TIP19/TextField/Xu} learn the offset field of each pixel to the closest text boundary.  SAST \cite{MM19/SAST/Wang} predicts more geometric properties, including border offset, center offset, and vertex offset. Based on a similar framework of TextField, KAC \cite{CVPR24/KACM/Zheng} and KPN \cite{TNNLS23/KPN/Zhang}  focus on learning a dynamic convolution kernel. Based on KAC,  MTPT \cite{TMM25/MTPT/Xie} further introduces a self-supervised scene text pre-training technique using multimodal knowledge CLIP. 

By exploring uncertainty information, FSAS \cite{FSAS} particularly predicts the reliability of text and separatrix. TextPMS evaluates the uncertainty in text annotations explicitly using a probability map. Instead of multi-scale features, S3INet \cite{TNNLS25/S3INet/Wang} produces a single-level feature map that encapsulates multiscale features while preserving rich semantic information and highlighting the foreground. Like a camera focusing on the target, ZTD \cite{TNNLS23/ZTD/Yang} makes full use of the advantages of coarse and fine features to enhance the reliability of shrink-masks, while STD \cite{TMM25/STD/Han} exploits a spotlight calibration to concentrate efforts on the candidate region 
Further, DBNet\cite{AAAI20/DB/Liao}, DBNet++\cite{DBNet++}, SIR \cite{MM23/SIR/Qin}, and TPPAN \cite{TCSVT24/TPPAN/Xu} convert the binarization in post-processing into a differentiable operation via a learned threshold map. In CIC \cite{TCSVT25/CIC/Han}, the proposed differentiable adaptive gap operator enables the texts and kernels to collaboratively generate information with each other. Unlike previous methods that shrink the same distance everywhere, MTD \cite{TIP25/MTD/Han} focuses on the local contour to shrink different distances by a magnetic pull mechanism. 
Instead of the shrink-mask strategy, TPF \cite{TMM22/TPF/Yang} simulates a band-pass filter to directly segment the whole text, and CM-Net \cite{TIP22/CM-Net/Yang} focuses on constructing a concentric mask. MorphText \cite{TMM23/Morphtext/Xu} proposes two deep morphological modules to regularize text segments and determine the linkage between them. Instead of representing text instances in the image spatial domain using masks or sequences of contour points under Cartesian or polar coordinates, 
 LeafText \cite{TMM23/LVT/Yang} simulates the leaf growth process to rebuild text contours. Differently, TextDCT \cite{TMM23/TextDCT/Su} and TextDCTv2 \cite{TMM26/TextDCTv2/Pan} parameterize the mask. While FCENet \cite{CVPR21/FCENet/Zhu}, EdgeText \cite{TCSVT25/EdgeText/Yang} and TPSNet \cite{MM22/TPSNet/Wang} parameterize the contours.  LRANet \cite{AAAI24/LRANet/Su} and TextRSR \cite{TMM26/TextRSR/Shao} model structural relationships among various text shapes in a low-rank representation and robust subspace recovery, respectively. 

 \subsubsection{\textbf{Component-wise based Methods}}
The component-wise based methods \cite{ECCV16/CTPN/Tian,ECCV18/Textsnake/Long,CVPR19/CRAFT/Baek,PR19/ICG/Tang,CVPR20/DRRGN/Zhang,PR21/ReLaText/Ma,AAAI25/ERRNet/Sun} first generate the local components based on the pixel-wise predictions, and then focus on exploring the grouping strategies (\textit{e.g.}, heuristic rules, linkage estimation, relationship reasoning, etc.). 

For example, CTPN \cite{ECCV16/CTPN/Tian} employs fixed-width horizontal text components to process text instances exhibiting extreme aspect ratios. TextSnake \cite{ECCV18/Textsnake/Long} reconstructs entire text instances by sliding a circle along the central axis. TextDragon \cite{ICCV19/TextDragon/Feng} adopts a series of local quadrangles to describe the arbitrary shape of the scene text instance, and exploits a ROISlide mechanism to connect the detection and recognition.
Instead of the offline reconstruction, ICG \cite{PR19/ICG/Tang} regards the linkages between the estimated rotated squares as a binary classification, to dynamically formulate the entire text instance. Moreover, DRRGN \cite{CVPR20/DRRGN/Zhang} models associative relationships between text components for grouping via the graph convolutional
network (GCN), followed by applying the shortest path algorithm to determine sequential ordering within each group. Differently, ReLaText \cite{PR21/ReLaText/Ma} regards the component-wise based method as a scene graph generation task, and applies GCN to infer connections in predefined triplet structures. To eliminate the complex and tricky post-processing, ERRNet \cite{AAAI25/ERRNet/Sun} initially decomposes each text instance into ordered components, then models their spatial relationships as sequential motion trajectories, developing an end-to-end tracking decoder that eliminates post-processing in a component-wise based paradigm.

\begin{table*}[htbp]
\renewcommand\arraystretch{1}
  
  \caption{ Comparisons  between existing methods on the dataset CTW \cite{PR19/CTD_CLOC/Liu}. $\bigtriangleup$, $\diamond$, and $\star$ denote the short side, the long side, and the height, respectively, when resizing the image with keeping the aspect ratio. ($s_{min}$,\;$s_{max}$) indicates the short side is set to $s_{min}$ if it is less than $s_{min}$, and keep the longer side is not larger than $s_{max}$.  $\natural$ means using the recognition branch to optimize the detection in an end-to-end framework. `MS' represents the multi-scale testing. The red italic and the blue bold denote the optimal value for top-down based methods and bottom-up-based methods, respectively. }
  \label{table111}
  \vspace{-0.25cm}
  \begin{center}
     \resizebox{2.08\columnwidth}{!}{
     \begin{tabular}{|c|c|c|c|c|c|c|c|c|c|c|c|}
     \hline
     \multirow{2}{*}{Type} & \multirow{2}{*}{Method} & \multirow{2}{*}{Venue} & \multirow{2}{*}{Backbone} & \multirow{2}{*}{\begin{tabular}[c]{@{}c@{}}Pretrain\\ data\end{tabular}} & \multirow{2}{*}{\begin{tabular}[c]{@{}c@{}}training\\ data\end{tabular}} & \multirow{2}{*}{\begin{tabular}[c]{@{}c@{}}testing \\ size\end{tabular}} & \multicolumn{3}{c|}{IOU@0.5} & \multirow{2}{*}{FPS} & \multirow{2}{*}{GPUs (\#)} \\ \cline{8-10} & & & & & & & R (\%)  & P (\%)  & F (\%)  &  & \\
     \hline
     \hline
     \multirow{22}{*}{\begin{sideways} Top-down based methods\end{sideways}}   
     
     & ATRR \cite{CVPR19/ATRR/Wang} & CVPR'19 & SE-VGG16 & $\times$ & CTW & 720$\times$1280 & 80.2 & 80.1 &80.1  & 10.0  & Nvidia P40 (1) \\ 
     \cline{2-12}
     & CTC-CLOC \cite{PR19/CTD_CLOC/Liu} & PR'19 & ResNet50 &  SynText & CTW & (600,1000) & 69.8 & 77.4 & 73.4 & 13.3 & GTX 1080 (1) \\ 
     \cline{2-12} 
     & ABCNet \cite{CVPR20/ABCNet/Liu} & CVPR'20 & ResNet50 & Synth150K+COCO-Text+ICDAR-MLT& CTW & (560, 800) $\bigtriangleup$, 1333 $\diamond$ & N/A & N/A & 74.1 & 22.8 & Tesla V100 (4) \\ 
     \cline{2-12}
     & \multirow{2}{*}{TextFuseNet \cite{IJCAI20/TextFuseNet/Ye}}  & \multirow{2}{*}{IJCAI'20} & ResNet50  & \multirow{2}{*}{SynText} & \multirow{2}{*}{CTW} & \multirow{2}{*}{1000 $\bigtriangleup$} & 85.0 & 85.8 & 85.4 &7.3 & \multirow{2}{*}{Tesla V100 (4)} \\ \cline{4-4} \cline{8-11} & & &
        ResNet101 &  & &  & 85.4 & 87.8 & 86.6 & 3.7 &  \\
     \cline{2-12} 
     & ContourNet \cite{CVPR20/ContourNet/Wang} & CVPR'20 & ResNet50 & $\times$ & CTW & 720$\times$1280 &  84.1  & 83.7  & 83.9  & 4.5  & TiTan X (1) \\ 
     \cline{2-12} 
     & \multirow{2}{*}{SDM \cite{ECCV20/SDM/Xiao}}  & \multirow{2}{*}{ECCV'20} & \multirow{2}{*}{ResNet50} & \multirow{2}{*}{MLT17$\rightarrow$TOT} & \multirow{2}{*}{CTW} & 768 $\diamond$ &  82.3   & 85.8   & 84.0   & \multirow{2}{*}{N/A} & \multirow{2}{*}{N/A}\\ \cline{7-10} & & & & & & MS & 84.4 & 88.4 & 86.4 & & \\ 
     \cline{2-12}
     & Poly-FRCNN-3 \cite{IJDAR20/TOT/chng}  & IJDAR'20 & Inception-ResNet-v2 &  SynText$\rightarrow$COCO-Text & CTW & N/A & 62.0 & 86.0 & 72.0 & N/A & N/A \\ 
     \cline{2-12} 
     & \multirow{2}{*}{MS-CAFA \cite{TMM19/MS-CAFA/Dai}}  & \multirow{2}{*}{TMM'20} & \multirow{2}{*}{ResNet50} & \multirow{2}{*}{$\times$} & \multirow{2}{*}{CTW} & (600, 800) &  83.3 & 86.1 & 84.6 & 1.0 & \multirow{2}{*}{GTX 1080Ti (1)}\\ \cline{7-11} & & & & &  & MS & 85.1 & 85.7 &85.4 & N/A & \\ 
     \cline{2-12} 
     & Mask-TTD \cite{TIP20/Mask-TTD/Liu} & TIP'20 & ResNet50-FPN & $\times$ & CTW & N/A & 79.0 & 79.7 & 79.4  & N/A & Tesla P100 (1)
     \\ 
     \cline{2-12} 
     & ABCNetV2 \cite{TPAMI21/ABCNetV2/Liu} & TPAMI'21 & ResNet50 & Synth150K + ICDAR-MLT & CTW &  (640, 896) $\bigtriangleup$, 1600 $\diamond$ & 83.8 & 85.6 & 84.7 & 10 & Tesla V100 (4) \\ 
     \cline{2-12}
     & PCR \cite{PCR21} & CVPR'21 &  DLA34 & $\times$ & CTW & (416,640) $\bigtriangleup$  & 82.3 & 87.2 & 84.7  & N/A &  GTX 2080Ti (1)  \\
     \cline{2-12} 
     & TextBPN \cite{TextBPN} & ICCV'21 & ResNet50-FPN & MLT17 & CTW & N/A & 83.6 & 86.5 & 85.0  & 12.2 & RTX-3090 (1)  \\
     \cline{2-12} 
     & MSEM \cite{TextBPN} & TIP'21 & ResNet50 & SynText & CTW & 460 $\bigtriangleup$ & 82.1 & 86.1 & 84.1  & 32.3 & GTX Titan X  (1)  \\
     
     \cline{2-12}
     & R-Net \cite{RNet} & TMM'21 & VGG16 & SynText & CTW & N/A & 71.0 & 74.6 & 72.8  & N/A & TITANX Pascal (2)  \\
    \cline{2-12} 
    & OPMP \cite{OPMP} & TMM'21 & ResNet50-FPN-GN & MLT17 & CTW & N/A & 80.8 & 85.1 & 82.9  & N/A & TITAN X (2)  \\

     \cline{2-12} 
     & ATTR \cite{ECCV22/ATTR/Zhou} & ECCV'22 & ResNet18 & SynthText 150K & CTW & 640$\triangle$ & 80.1 & 80.2 & 80.1 & N/A & NVIDIA TITAN Xp (1) \\ 
     \cline{2-12}
     & SwinTextSpotter \cite{SwinTextSpotter} & CVPR'22 & SwinTranformer-FPN & SynthText +MLT17
      & CTW & N/A & N/A & N/A & 88.0  & N/A & N/A 
     \\ 
     \cline{2-12} 
     & TESTR \cite{TESTR} & CVPR'22 & ResNet50 & Synth150K+MLT17 & CTW & N/A & 82.6 & 92.0 & 87.1  & 5.3 & RTX 2080Ti (8)
     \\    
     \cline{2-12} 
     & I3CL \cite{du2022i3cl} & IJCV'22 & ResNet50 & Synth150K
     & CTW & N/A & 84.6 & 88.4 & 86.5  & N/A & Tesla V100 (4) 
     \\ 
     \cline{2-12} 
     & EMA  \cite{TIP20/Mask-TTD/Liu} & TIP'22 & DLA-34 & Synth150K
         & CTW & N/A & 82.1 & 86.1 & 84.1  & 32.3 &  GTX Titan X
     \\  
     \cline{2-12} 
     & ESTextSpotter-Polygon \cite{ESTextSpotter} & ICCV'23 & ResNet50 & Synth150K+MLT17
     & CTW & N/A & 88.6 & 91.5 & 90.0 & 4.3 &  RTX 3090 
     \\ 
     \cline{2-12} 
     & DeepSolo++  \cite{DeepSolo++} & TPAMI'23 & ResNet50 
     & \makecell{Synth150K+MLT17\\+IC13+IC15+TextOCR} 
     & CTW & N/A & 86.3 & \textcolor{red}{ \textbf{92.5}} & 89.3  & 20.0 &  N/A 
     \\ 
     \cline{2-12} 
     & TextBPN++  \cite{TextBPN++} & TMM'23 & ResNet50-DCN & MLT17
     & CTW & N/A & 84.7 & 88.3 & 86.5  & 16.5 &  RTX 3090 
     \\ 
     \cline{2-12} 
     & DPText-DETR  \cite{ye2022dptext} & AAAI'23 & ResNet50 & Synth150K+MLT19
     & CTW & N/A & 86.2 & 91.7 & 88.8  & N/A & Tesla A100 (4) 
     \\ 
     \cline{2-12} 
     & HGR-Net  \cite{HGR-Net} & TIP'23 &  ResNet50-FPN
     & \makecell{Synth150K+MLT17} 
     & CTW & (640,896) $\times$ 1600 & 85.9 & 86.1 & 86.0 & N/A 
     &  RTX 2080Ti (1) \\
     \cline{2-12} 
     & TextDCT\cite{TMM23/TextDCT/Su} & TMM'23 & ResNet50 & SynthText & CTW & (1056,1440) & 85.3 & 85.0 & 85.1 & 17.2 &  NVIDIA V100 (1) \\ 
     \cline{2-12} 
     & SRFormer \cite{AAAI24/SRFormer/Bu} & AAAI'24 & ResNet50 & SynthText150K + MLT17 +TotalText & CTW & (1000,1800) & 89.8 & 89.4 & 89.6 & N/A & NVIDIA 3090 (8) \\ 
     \cline{2-12}
     & CR2-Net \cite{TIPS24/CR2-Net/Wang} & TIPS'24 & ResNet50 & $\times$ & CTW & 400, 600, 720, 1000, 1200 $\triangle$, 2000 $\diamond$ & 78.6 & 86.2 & 82.2 & N/A & N/A \\ 
     \cline{2-12}
     & Box2Poly \cite{AAAI24/Box2Poly/Chen} & AAAI'24 & ResNet50 & $\times$ & CTW & (480 $\rightarrow$ 832,1600) & 87.5 & 88.8 & 88.1 & N/A & NVIDIA RTX 3090(8) \\ 
     \cline{2-12}
     &  LayoutFormer  \cite{CVPR24/LayoutFormer/Liang} & CVPR'24 &  ResNet50
     & \makecell{SynthText} 
     & CTW & 1120 $\star$ & 84.3 & 88.2 & 86.2 & N/A 
     &  N/A
     \\ 
     \cline{2-12} 
     &  LRANet  \cite{su2024lranet} & AAAI'24 &  ResNet50-DCN-FPN
     & \makecell{Synth150K} 
     & CTW & 704 $\bigtriangleup$ & 85.5 & 89.4 & 87.4 & \textcolor{red}{ \textbf{37.2}} 
     & RTX 3090 \\
     \cline{2-12} 
     &  DAT-SEG  \cite{ICML24/DAT/Wan} &ICML'24 &  SwinL
     & \makecell{$\times$} 
     & CTW & (800,1333) & \textcolor{red}{\textbf{90.9}}  & \textcolor{red}{\textbf{92.5}} & \textcolor{red}{\textbf{91.7}} & N/A 
     &  NVIDIA A100(8)
     \\ 
     \cline{2-12} 
     &   IAST  \cite{IAST2024} &TIP'24 & ResNet50-DCN
     & Synth150K+MLT-2017+ToT+ICDAR2015 
     & CTW & N/A &  84.8 &  89.2 & 86.9 & N/A 
     &  RTX 3090(2)
     \\ 
     \cline{2-12} 
     & CT-Net\cite{TCSVT24/CT-Net/Shao} &TCSVT'24 & ResNet50 & SynthText & CTW & N/A & 83.8 & 88.5 & 86.1 & 11.2 & NVIDIA V100(1)
     \\ 
     \cline{2-12} 
     &Zhang et al\cite{TMM24/Zhangetal/Zhang} &TMM'24 & ResNet50-DCN & MLT17 & CTW & (640,1024)  & 84.7 & 88.3 & 86.5 & 16.5 & RTX 3090(1)
     \\ 
     \cline{2-12}
     & DG-Bridge Spotter \cite{CVPR25/DG-Brigde-Spotter/Huang} & CVPR'25 & ResNet50 & SynthText & CTW & (1000,1824) & 86.2 & 92.1 & 89.0 & N/A & NVIDIA GeForce RTX 3090(1) \\

     \hline
     \hline
     \multirow{41}{*}{\begin{sideways} Bottom-up based methods\end{sideways}}   
    
     & TextSnake \cite{ECCV18/Textsnake/Long} & ECCV'18 & VGG16 & SynText & CTW & original size & 85.3 & 67.9 & 75.6  & N/A  &  TiTan X (2) \\ 
     \cline{2-12} 
     & MSR \cite{IJCAI19/MSR/Xue} & IJCAI'19 & ResNet50 & SynText & CTW & MS & 78.3 & 85.0 & 81.5 & N/A  &  GTX 1080Ti (2) \\ 
     \cline{2-12}
     & SegLink++ \cite{PR19/SegLink++/Tang} & PR'19 & VGG16 & SynthText & CTW & 512$\triangle$ & 79.8 & 79.8 & 81.3 & 7.1 & Tesla P100(1) \\ 
     \cline{2-12}
     & SAE \cite{CVPR19/SAE/Tian} & CVPR'19 & ResNet50 & SynText & CTW & 800 $\diamond$ & 77.8 & 82.7 & 80.1 & N/A  &  TiTan X (1) \\  
     \cline{2-12}
     & CRAFT \cite{CVPR19/CRAFT/Baek} & CVPR'19 & VGG16-BN & SynText & CTW & 1024 $\diamond$ & 81.1 & 86.0 & 83.5 & N/A  &  N/A \\  
     \cline{2-12}
     & \multirow{2}{*}{LOMO \cite{CVPR19/LOMO/Zhang}}  & \multirow{2}{*}{CVPR'19}  & \multirow{2}{*}{ResNet50} & \multirow{2}{*}{SynText} & \multirow{2}{*}{CTW} & 512 $\diamond$ & 69.6 & 89.2 & 78.4 & 4.4  & \multirow{2}{*}{Tesla K40m (4)} \\ \cline{7-11}
     & & & & & & MS & 76.5 & 85.7 & 80.8 & N/A                 & \\ 
     
     \cline{2-12} 
     & \multirow{2}{*}{PSENet-1s \cite{CVPR19/PSENet/Wang}} 
     & \multirow{3}{*}{CVPR'19} 
     & \multirow{3}{*}{ResNet50} & $\times$  
     & \multirow{3}{*}{CTW} 
     & \multirow{3}{*}{1280 $\diamond$}  & 75.6 & 80.6 & 78.0 & 3.9 
     & \multirow{3}{*}{N/A (4)} \\ 
     
     \cline{5-5} 
     \cline{8-11}
     &  & & & MLT17 &  & & 79.7 & 84.8 & 82.2 & 3.9 &  \\ 
     \cline{2-2} 
     \cline{5-5} 
     \cline{8-11}
     & PSENet-4s \cite{CVPR19/PSENet/Wang}   & & & MLT17 & & & 77.8 & 82.1 & 79.9 & 8.4 & \\

     \cline{2-12} 
     & \multirow{6}{*}{PAN \cite{ICCV19/PAN/Wang}}  
     & \multirow{6}{*}{ICCV'19} 
     & \multirow{6}{*}{ResNet18} 
     & \multirow{3}{*}{$\times$}  
     & \multirow{3}{*}{CTW} 
     & 320  $\bigtriangleup$ & 72.6    & 82.2    & 77.1    & \textcolor{blue}{\textbf{84.2}}  
     & \multirow{6}{*}{GTX 1080Ti (4)} \\ 
     \cline{7-11} 
     & & & & & & 512 $\bigtriangleup$ & 77.1  & 83.8 & 80.3 & 58.1 & \\ \cline{7-11}   
     & & & & & & 640 $\bigtriangleup$ & 77.7 & 84.6 & 81.0 & 39.8 &  \\ \cline{5-11}
     & & & 
     & \multirow{3}{*}{SynText} 
     & \multirow{3}{*}{CTW} & 320 $\bigtriangleup$ & 75.0 & 85.6 & 79.9 & \textcolor{blue}{\textbf{84.2}} & \\ 
     \cline{7-11}
     & & & & & & 512 $\bigtriangleup$ & 81.5  & 85.5  & 83.5  & 58.1 & \\ \cline{7-11}
     & & & & & & 640 $\bigtriangleup$ & 81.2 & 86.4 & 83.7 & 39.8 &  \\
     \cline{2-12}
     & TextDragon \cite{ICCV19/TextDragon/Feng} $\natural$ & ICCV'19 & VGG16 & SynText & CTW & N/A  & 82.8 & 84.5 & 83.6  & N/A  & GTX TiTan X (1) \\
     \cline{2-12} 
     & ICG \cite{PR19/ICG/Tang} & PR'19 & VGG16 &  SynText & CTW & 512 $\bigtriangleup$ & 79.8 & 82.8 & 81.3 & N/A & Tesla P100 (1) \\ 
     \cline{2-12}
     & AB-LSTM \cite{TOMM19/AB_LSTM/Liu} & TOMM'19 & VGG16 &  MLT17 & CTW & N/A & 81.6 & 83.0 & 82.3 & N/A & GTX 1080Ti (1) \\ 
     \cline{2-12}
     & TextField \cite{TIP19/TextField/Xu} & TIP'19 & VGG16 & SynText & CTW & 576$\times$576 & 79.8    & 83.0    & 81.4  & N/A  & TiTan Xp (1) \\  
     \cline{2-12}
     & \multirow{2}{*}{DB \cite{AAAI20/DB/Liao}}  & \multirow{2}{*}{AAAI'20} & ResNet18-DCN  & \multirow{2}{*}{SynText} & \multirow{2}{*}{CTW} & \multirow{2}{*}{1024 $\star$} & 77.5 & 84.8 & 81.0  & 55.0 & \multirow{2}{*}{GTX 1080Ti (1)} \\ \cline{4-4} \cline{8-11} & & &
      ResNet50-DCN &  & &  & 80.2 & 86.9 & 83.4 & 22.0 &  \\
      \cline{2-12} 
      & TextPerceptron\cite{AAAI20/TextPerceptron/Qiao} $\natural$ & AAAI'20 & ResNet50 & SynText & CTW & 1250 $\diamond$ & 81.9 & 87.5 & 84.6 & N/A & Tesla V100 (8) \\
     \cline{2-12} 
     & DRRG \cite{CVPR20/DRRGN/Zhang} & CVPR'20 & VGG16 & MLT17 & CTW & (512,1024)  & 83.0    & 85.9    & 84.5  & N/A  & RTX 2080Ti (1) \\ 
     \cline{2-12}
     & \multirow{2}{*}{CRNet \cite{MM20/CRNet/Zhou}}  & \multirow{2}{*}{MM'20} & \multirow{2}{*}{ResNet50} & \multirow{2}{*}{SynText} & \multirow{2}{*}{CTW} &  1536 $\diamond$ & 80.9    & 87.0    & 83.8 & \multirow{2}{*}{N/A} & \multirow{2}{*}{TiTan X (1)}\\ \cline{7-10} & & & & &  & MS &  82.0    & 86.6 & 84.2 & & \\ 
     \cline{2-12}
     & TextRay\cite{MM20/TextRay/Wang} & MM'20 & ResNet50 & $\times$ & CTW & (640,800)  & 80.4    & 82.8    & 81.6  & 2.0 & TiTan X (4) \\
     \cline{2-12}
     & TextFuseNet \cite{IJCAI20/TextFuseNet/Ye} & ICJAI'20 & ResNet50 & SynText & CTW & 1000$\triangle$  & 85.8 & 85.0 & 85.4 & 7.3 & NVidia Tesla V100 (1) \\ 
     \cline{2-12}
     & ReLaText \cite{PR21/ReLaText/Ma} & PR'21 & ResNet50 & SynText & CTW & 800 $\diamond$  & 83.3 & 86.2 & 84.8  & 10.6  & Nvidia V100 (4) \\
    \cline{2-12} 
     & FCENet \cite{FCENet} & CVPR'21 &  ResNet50-DCN-FPN &  $\times$ & CTW & (640,1280) & 83.4 & 87.6 & 85.5 & N/A & RTX 2080Ti(2)  \\ 
     \cline{2-12} 
     & TextMountain \cite{PR21/TextMountain/Zhu} & PR'21 & ResNet50 & ImageNet & CTW & 800$\diamond$ & 83.4 & 82.9 & 83.2 & N/A & GeForce GTX 1080Ti(1) \\ 
     \cline{2-12}
     & PAN++ \cite{TPAMI21/PAN++/Wang} & TPAMI'21 & ResNet18 & SynText & CTW & 640$\triangle$ & 81.1 & 87.1 & 84.0 & \textcolor{blue}{\textbf{84.2}} & NVidia V100 (4) \\ 
     \cline{2-12}
     & FSGNet  \cite{FSGNet} & CVPR'22 & ResNet50 & MLT+Bezier Curve Synthetic
     & CTW & N/A & 82.4 & 88.1 & 88.1  & N/A  &  N/A 
     \\ 
     \cline{2-12} 
     & TPSNet \cite{TPSNet2022} & ACMM'22 & ResNet50-DCN-FPN & Synth150K+MLT17  & CTW & N/A & 86.3 & 88.7 & 87.5 & 17.9 & RTX 3090 
     \\ 
     \cline{2-12} 
     & KPN  \cite{KPN} & TNNLS'22 & ResNet50 & MLT17
     & CTW & N/A & 86.4 & 84.0 & 85.2  & N/A &  GTX 1080Ti 
     \\ 
     \cline{2-12} 
     & SRSTS \cite{wu2022decoupling} & ACMM'22 & ResNet50-BiFPN 
     & \makecell{Synth800K\\+Bezier Curve Synthetic}
     & CTW & N/A & 83.3 & 88.9 & 86.0  & N/A & RTX 3090Ti
     \\ 
     \cline{2-12}
     & Boundary-TextSpotter \cite{Boundary-TextSpotter22} & TIP'22 & ResNet50 & SynText & CTW & N/A & 78.2 & 88.1 & 82.9  & 18.5 & Nvidia V100 (4) 
     \\ 
     \cline{2-12}
     & CM-Net \cite{TIP22/CM-Net/Yang} & TIP'22 & ResNet+FPN & SynthText & CTW & 640 $\triangle$ & 82.2 & 86 & 84.1 & 50.3 & 1080Ti (2) \\ 
     \cline{2-12}
     & ADNet  \cite{ADNet} & TMM'23 &  ResNet50-FPN
     & \makecell{SynText} 
     & CTW & 800 $\bigtriangleup$  & 83.1 & 88.2 & 85.6 & N/A
     &  GTX 1080Ti (1) \\ 
     \cline{2-12}
     & MorphText\cite{TMM23/Morphtext/Xu} & TMM'23 & ResNet50+FPN & SynthText & CTW & (640,640)  & 83.3  & \textcolor{blue}{\textbf{90.0}}  & 86.5 & 2.0  & RTX 3090 (1) \\
     \cline{2-12} 
     & RSMTD\cite{TMM23/RSMTD/Yang} & TMM'23 & ResNet50 & SynText & CTW & 640 $\bigtriangleup$ & 80.3  & 87.8  & 83.9 & 72.1  & GTX 1080Ti (2) \\
     \cline{2-12} 
     & SIR  \cite{SIRACMM23} & ACMM'23 & ResNet18 & Synth150K
     & CTW & N/A & 83.7 & 87.4 & 85.5  & 38.7 & RTX 2080Ti \\ 
     \cline{2-12} 
     & HFENet\cite{TITS23/HFENet/Liang} & TITS'23 & ResNet50 & SynthText
     & CTW & 800 $\star$  & 83.4 & 88.1 & 85.7  & 18.1 & GTX 1080Ti (1) \\ 
     \cline{2-12} 
     & \multirow{2}{*}{FSAS \cite{FSAS}}   
     & \multirow{2}{*}{TIP'23}  & ResNet18-FPN
     & \multirow{2}{*}{$\times$}
     & \multirow{2}{*}{CTW} 
     & \multirow{2}{*}{1200$\times$800}  & 77.7 & 84.6 & 81.0 & 35.2 
     & \multirow{2}{*}{Titan X Pascal (4)} \\ 
     \cline{4-4}
     \cline{8-11}
     & & & ResNet50-FPN & & & & 82.5 & 85.3 & 83.9 & 25.1 & \\

     \cline{2-12} 
     & \multirow{4}{*}{DBNet++ \cite{DBNet++}}
     & \multirow{4}{*}{TPAMI'23} 
     & \multirow{2}{*}{ResNet18-DCN}
     & \multirow{4}{*}{SynText} 
     & \multirow{4}{*}{CTW} & 800 $\star$  & 81.0 & 84.3 & 82.6 & 49 
     & \multirow{4}{*}{GTX 1080Ti} \\
     \cline{7-11}
     & & & & & & 1024 $\star$ & 81.3 & 86.7 & 83.9 & 40 & \\
     \cline{4-4}
     \cline{7-11}
     & & & \multirow{2}{*}{ResNet50-DCN}  & & & 800 $\star$ & 82.8 & 87.9 & 85.3 & 26 & \\
     \cline{7-11}
     & & & & & & 1024 $\star$ & 82.0 & 88.5 & 85.1 & 21 & \\
     
     \cline{2-12} 
     & ZTD\cite{TNNLS23/ZTD/Yang} & TNNLS'23 &  ResNet18
     & \makecell{SynText} 
     & CTW & 640 $\bigtriangleup$  & 80.2 & 88.4 & 84.1 & 76.9
     &  GTX 1080Ti (1)
     \\ 
     
     \cline{2-12} 
     & RP-Text \cite{TMM23/RP-Text/Wang} & TMM'23 & ResNet18
     & \makecell{SynText} 
     & CTW & 700 $\bigtriangleup$  & 81.6 & 87.8 & 84.7 & 23.8 & N/A \\ 
 
     \cline{2-12} 
     & HFENet \cite{TITS23/HFENet/Liang} & TIPS'23 & ResNet18 & SynthText & CTW & 640$\times$640 & 81.2 & 85.1 & 83.1 & 32.2 & GeForce GTX 1080Ti(1) \\ 
     \cline{2-12}
     & CBNet \cite{IJCV24/CBNet/Zhao} & IJCV'24 & ResNet18 & SynthText+ICDAR-MLT19 & CTW &  & 82.9 & 89.3 & 86.0 & 31.3 & NVIDIA Tesla P40 (2) \\   
     \cline{2-12}
     & \multirow{2}{*}{KAC \cite{zheng2024kernel}}
     & \multirow{2}{*}{CVPR'24}
     & ResNet18
     & \multirow{2}{*}{SynText}
     & \multirow{2}{*}{CTW}
     & \multirow{2}{*}{640×640}
     & 80.9 & 87.5 & 84.1 & 82.9
     & \multirow{2}{*}{NVIDIA TITAN RTX (1)} \\
     \cline{4-4}
     \cline{8-11}
     & & & ResNet50 & & & & 85.4 & 88.6 & 86.8 & 19.2 & \\
     \cline{2-12}
     & STD \cite{TMM25/STD/Han} & TMM'25 & ResNet50 & SynthText+ICDAR+MLT17 & CTW & 1152$\triangle$ & 84.9 & 88.5 & 86.7 & 12.1 & GTX 1080Ti(1) \\ 
     \cline{2-12}
     & CIC \cite{TCSVT25/CIC/Han} & TCSVT'25 & ResNet50 & SynthText+ICDAR+MLT17 & CTW & 736$\triangle$ & \textcolor{blue}{\textbf{88.9}} & 86.2 & 87.5 & 22.9 & GTX 1080Ti(1) \\ 
     \cline{2-12}
     & TextRSR \cite{TMM26/TextRSR/Shao} & TMM'26 & ResNet-DCN-FPN & SynthText 150K & CTW & 800,1600,1200,1000 & 87.5 & 89.5 & \textcolor{blue}{\textbf{88.5}} & 37.8 & NVIDIA RTX 3090 GPU (1) \\ 
     \hline
     \end{tabular}
     }
  \end{center}

\end{table*}


\begin{table*}[htbp]
\renewcommand\arraystretch{1}
  \caption{ Comparisons between existing methods on the dataset TOT \cite{IJDAR20/TOT/chng}.  $\spadesuit$ and $\clubsuit$ denote evaluating the performance with the metric \cite{IJDAR20/TOT/chng} IOU@0.5 and DetEval-v2 respectively.}
  \label{table222}
  \vspace{-0.25cm}
  \begin{center}
     \resizebox{2.03\columnwidth}{!}{
     \begin{tabular}{|c|c|c|c|c|c|c|c|c|c|c|c|}
     \hline
     \multirow{2}{*}{Type} & \multirow{2}{*}{Method} & \multirow{2}{*}{Venue} & \multirow{2}{*}{Backbone} & \multirow{2}{*}{\begin{tabular}[c]{@{}c@{}}Pretrain\\ data\end{tabular}} & \multirow{2}{*}{\begin{tabular}[c]{@{}c@{}}training\\ data\end{tabular}} & \multirow{2}{*}{\begin{tabular}[c]{@{}c@{}}testing \\ size\end{tabular}} & \multicolumn{3}{c|}{DetEval-v1} & \multirow{2}{*}{FPS} & \multirow{2}{*}{GPUs (\#)} \\ \cline{8-10} & & & & & & & R (\%)  & P (\%)  & F (\%)  &  & \\
     \hline
     \hline
     \multirow{30}{*}{\begin{sideways} Top-down based methods \end{sideways}}  
     & SPCNet \cite{AAAI19/SPCN/Xie} & AAAI'19 & ResNet50 & SynText$\rightarrow$MLT17 & TOT & N/A & 82.8 & 83.0 & 82.9 & N/A  & N/A (8)\\ 
     \cline{2-12}
     & ATRR \cite{CVPR19/ATRR/Wang} & CVPR'19 & SE-VGG16 & $\times$ & TOT & (720,1280) & 76.2 &80.9 & 78.5 & 10.0  & Nvidia P40 (1) \\

     \cline{2-12}
    & \multirow{3}{*}{Poly-FRCNN-3 \cite{IJDAR20/TOT/chng}}
    & \multirow{3}{*}{IJDAR'20}
    & \multirow{3}{*}{Inception-ResNet-v2}
    & \multirow{3}{*}{SynText$\rightarrow$COCO-Text}
    & \multirow{3}{*}{TOT}
    & \multirow{3}{*}{N/A}
    & 59.0 & 68.0 & 63.0 & 3.3 & \multirow{3}{*}{N/A} \\
    \cline{8-11}
    & & & & & &
    & 68.0 & 78.0 & 73.0 & 3.3 & \\
    \cline{8-11}
    & & & & & &
    & 70.0 & 80.0 & 75.0 & 3.3 & \\
    \cline{2-12}
    & \multirow{2}{*}{Mask-TextSpotter \cite{TPAMI19/Mask_TextSpotter++/Liao}}
    & \multirow{2}{*}{TPAMI'19}
    & \multirow{2}{*}{ResNet50-FPN}
    & \multirow{2}{*}{SynText}
    & \multirow{2}{*}{TOT}
    & \multirow{2}{*}{N/A}
    & 75.4 & 81.8 & 78.5 & N/A & \multirow{2}{*}{TiTan Xp (1)} \\
    \cline{8-11}
    & & & & & &
    & 82.4 & 88.3 & 85.2 & N/A & \\
    \cline{2-12}

     \cline{2-12}
     & TUTS \cite{ICCV19/TUTS/Qin} & ICCV’19 & ResNet50 & ImageNet & TOT & 480→800$\triangle$ & 85 & 87.8 & 86.4 & N/A & Tesla V100(15) \\ 
     \cline{2-12}
     & Boundary \cite{AAAI20/AYNIB/Wang} $\natural$ & AAAI'20 & ResNet50 & SynText & TOT & 1100 $\diamond$ &  85.0  & 88.9  & 87.0  & N/A  & TiTan Xp ($>$ 1) \\ 
     \cline{2-12} 
     & ContourNet \cite{CVPR20/ContourNet/Wang} & CVPR'20 & ResNet50 & $\times$ & TOT & (720,1280) &  83.9  & 86.9  & 85.4  & 3.8  & TiTan X (1) \\ 
     
     \cline{2-12} 
     & \multirow{2}{*}{TextFuseNet \cite{IJCAI20/TextFuseNet/Ye}}  
     & \multirow{2}{*}{IJCAI'20} 
     & ResNet50  
     & \multirow{2}{*}{SynText} 
     & \multirow{2}{*}{TOT} 
     & \multirow{2}{*}{1000 $\bigtriangleup$} & 83.2 & 87.5 & 85.3 & 7.1 
     & \multirow{2}{*}{Tesla V100 (4)} \\ 
     \cline{4-4} 
     \cline{8-11} & & 
     & ResNet101  &  & &  
     & 85.3 & 89.0 & 87.1 & 3.3 &  \\
     
     \cline{2-12} 
     & \multirow{2}{*}{SDM \cite{ECCV20/SDM/Xiao}}  
     & \multirow{2}{*}{ECCV'20} 
     & \multirow{2}{*}{ResNet50} 
     & \multirow{2}{*}{MLT17} 
     & \multirow{2}{*}{TOT} & 1024 $\diamond$ &  84.7  & 89.2  & 86.9  
     & \multirow{2}{*}{N/A} 
     & \multirow{2}{*}{N/A}\\ 
     \cline{7-10} & & & & &  & MS & 86.0 & 90.9 & 88.4 & & \\ 
     
     \cline{2-12}
     & \multirow{2}{*}{MS-CAFA \cite{TMM19/MS-CAFA/Dai}} 
     & \multirow{2}{*}{TMM'20} 
     & \multirow{2}{*}{ResNet50} & $\times$ 
     & \multirow{2}{*}{TOT} 
     & \multirow{2}{*}{(600,800)} & 74.7 & 83.5 & 78.9 
     & \multirow{2}{*}{0.8} 
     & \multirow{2}{*}{GTX 1080Ti (1)} \\ 
     \cline{5-5} 
     \cline{8-10}
     & & & & SynText & & & 78.6 & 84.6 & 81.5 & & \\
     \cline{2-12} 
     & Mask-TTD \cite{TIP20/Mask-TTD/Liu} & TIP'20 & ResNet50-FPN & $\times$ & TOT & N/A & 74.5 & 79.1 & 76.7  & N/A & Tesla P100 (1)  \\
    \cline{2-12} 
     & PCR \cite{PCR21} & CVPR'21 &  DLA34 & $\times$ & TOT & (416,640) $\bigtriangleup$  & 82.0 & 88.5 & 85.2  & N/A &  GTX 2080Ti \\ 
     \cline{2-12} 
     & TextBPN \cite{TextBPN} & ICCV'21 & ResNet50-FPN & MLT17 & TOT & N/A & 85.2 & 90.7 & 87.9  & 10.7 & RTX-3090 (1)  \\
     \cline{2-12} 
     & MSEM \cite{TextBPN} & TIP'21 & ResNet50 & SynText & TOT & 640 $\bigtriangleup$ & 83.3 & 88.2 & 85.6  & \textcolor{red}{\textbf{24.2}} & GTX Titan X  (1)  \\

     \cline{2-12} 
     & SwinTextSpotter \cite{SwinTextSpotter} & CVPR'22 & SwinTranformer-FPN & SynthText +MLT17
      & TOT & N/A & N/A & N/A & 88.0  & 1.0 & N/A 
     \\ 
     \cline{2-12} 
     & GLASS \cite{ronen2022glass} & ECCV'22 & ResNet50 & Synth800K & TOT & N/A & 85.5 & 90.8 & 88.1  & 3.0 & N/A
     \\ 
     \cline{2-12} 
     & TESTR \cite{TESTR} & CVPR'22 & ResNet50 & Synth150K+MLT17 & TOT & N/A & 81.4 & 93.4 & 86.9  & 5.3 & RTX 2080Ti (8)
     \\ 
     \cline{2-12}
     & Boundary-TextSpotter \cite{TIP22/Boundary-TextSpotter/Lu} & TIP'22 &  & SynthText & TOT & 800$\times$800 & 83.9 & 88.1 & 86.0 & 26.0 & NVIDIA V100(4) \\ 
     \cline{2-12} 
     & TTS \cite{TTS22cvpr} & CVPR'22 & ResNet50 
     & \makecell{Synth800K+COCO-Text\\+IC13+IC15+SCUT}
     & TOT & N/A & 81.2 & 89.6 & 85.2  & 13.4 & Nvidia V100 (4) 
     \\ 
     \cline{2-12} 
     & I3CL \cite{du2022i3cl} & IJCV'22 & ResNet50 & Synth150K
     & TOT & N/A & 83.7 & 89.2 & 86.3  & N/A & Tesla V100 (4) 
     \\
     \cline{2-12} 
     & EMA  \cite{EMA} & TIP'22 & DLA-34 & Synth150K
     & TOT & N/A & 88.3 & 88.2 & 85.6  & 24.2 &  GTX Titan X
     \\ 
     \cline{2-12} 
     & ESTextSpotter-Polygon \cite{ESTextSpotter} & ICCV'23 & ResNet50 & Synth150K+MLT17
     & TOT & N/A & 88.1 & 92.0 & 90.0  & 4.3 &  RTX 3090 
     \\ 
      \cline{2-12} 
     & DeepSolo \cite{deepsolo} & CVPR'23 & ResNet50 
     & \makecell{Synth150K+MLT17\\+IC13+IC15}
     & TOT & N/A & 82.1 & 93.1 & 87.3  & 17.0 & N/A 
     \\
     \cline{2-12} 
     & STSEA \cite{STSEA} & TIP'23 &  ResNet50 & LBTS-10K & TOT & N/A & 82.1 & 89.3  & 85.6  & N/A  & RTX 3090 (1) \\ 
     \cline{2-12} 
     & DPText-DETR  \cite{ye2022dptext} & AAAI'23 & ResNet50 & Synth150K+MLT19
     & TOT & N/A & 86.4 & 91.8 & 89.0  & 17.3 & Tesla A100 (4) 
     \\ 
     \cline{2-12} 
     & UNITS  \cite{UNITS} & CVPR'23 & SwinTransformer 
     & \makecell{SynText+IC13+IC15\\+MLT19+TextOCR+HierText} 
     & TOT & N/A & N/A & N/A & 89.8 & N/A &  Tesla A100 (16)
     \\ 
     \cline{2-12} 
     & NLIDC  \cite{NLIDC} & TIP'23 &  VGG16 
     & \makecell{SynText} 
     & TOT & N/A & 86.9 & 86.7 & 86.8 & N/A &  RTX 2080 (1)
     \\ 
     \cline{2-12} 
     & HGR-Net  \cite{HGR-Net} & TIP'23 &  ResNet50-FPN
     & \makecell{Synth150K+MLT17} 
     & TOT & (640,896) $\times$ 1600 & 86.5 & 88.6 & 87.5 & N/A 
     &  GTX 2080Ti (1) \\ 
     
     \cline{2-12} 
     & TextDCT\cite{TMM23/TextDCT/Su} & TMM'23 & ResNet50 & SynthText
     & TOT & (1024,2000) & 82.7 & 87.2 & 84.9 & 15.1 &  NVIDIA V100 (1) \\ 

     \cline{2-12} 
     &  LayoutFormer  \cite{LayoutFormer} & CVPR'24 &  ResNet50
     & \makecell{SynthText} 
     & TOT & 800 $\star$ & 85.1 & 89.3 & 87.1 & N/A 
     &  N/A
     \\ 
     \cline{2-12} 
     &  LRANet  \cite{su2024lranet} & AAAI'24 &  ResNet50-DCN-FPN
     & \makecell{Synth150K} 
     & TOT & 1000 $\bigtriangleup$ & 87.8 & 90.3 & 89.0 & N/A
     & RTX 3090 \\
     \cline{2-12} 
     &  Box2Poly  \cite{chen2024box2poly} & AAAI'24 &  ResNet50-FPN
     & \makecell{Synth150K+TOT+ICDAR19 MLT} 
     & TOT & N/A & 86.6 & 90.2 & 88.4 & N/A 
     &  RTX 3090 (8)
     \\ 
     \cline{2-12} 
     &  DAT-SEG  \cite{ICML24/DAT/Wan} &ICML'24 &  SwinL
     & \makecell{$\times$} 
     & TOT & (800,1333) &  \textcolor{red}{\textbf{89.2}} &  \textcolor{red}{\textbf{95.0}} & \textcolor{red}{\textbf{92.0}} & N/A 
     &  NVIDIA A100(8)
     \\ 
     \cline{2-12} 
     &   IAST  \cite{IAST2024} &TIP'24 & ResNet50-DCN
     & Synth150K+MLT-2017+ToT+ICDAR2015 
     & TOT & N/A &  85.2 & 94.7 & 89.7 & 7.8
     &  RTX 3090(2)
     \\ 
     \cline{2-12} 
     & CT-Net\cite{TCSVT24/CT-Net/Shao} &TCSVT'24 & ResNet50 & SynthText & TOT & N/A &  85.0 & 90.8 & 87.8 & 10.1 & NVIDIA V100(1)
     \\ 
     \cline{2-12} 
     &Zhang et al\cite{TMM24/Zhangetal/Zhang} &TMM'24 & ResNet50-DCN & MLT17 & TOT & (640,1024)  & 87.9 & 92.4 & 90.1 & 13.2 & RTX 3090(1)
     \\ 
     \cline{2-12}
     & ABCNet \cite{CVPR20/ABCNet/Liu} & CVPR'20 & ResNet50 & Synth150K+COCO-Text+ICDAR-MLT & TOT & 800$\triangle$ & N/A & N/A & 61.9 & 22.8 & Tesla V100 GPU(4) \\ 
     \cline{2-12}
     & ABCNetV2 \cite{TPAMI21/ABCNetV2/Liu} & TPAMI'21 & ResNet50 & Synth150K + ICDAR-MLT & TOT & (640 $\rightarrow$ 896, 1600) & 84.1 & 90.2 & 87 & 10.0 & Tesla V100 GPU(1) \\ 
     \cline{2-12}
     & SRFormer \cite{AAAI24/SRFormer/Bu} & AAAI'24 & ResNet50 & SynthText150K+MLT17+TotalText & TOT & (1000,1800) & 87.9 & 91.5 & 89.7 & N/A & NVIDIA 3090 (8) \\ 
     \cline{2-12}
     & DG-Bridge Spotter \cite{CVPR25/DG-Brigde-Spotter/Huang} & CVPR 25 & ResNet50 & SynthText & TOT & (1000,1824) & 86.5 & 92.0 & 89.2 & 6.7 & NVIDIA GeForce RTX 3090(1) \\

     \hline
     \hline
     \multirow{37}{*}{\begin{sideways} Bottom-up based methods \end{sideways}}   
     & SegLink++ \cite{PR19/SegLink++/Tang} & PR'19 & VGG16 & SynText & TOT & 768$\triangle$ & 80.9 & 82.1 & 81.5 & 7.1 & Tesla P100(1) \\ 
     \cline{2-12} 
     & CRAFT \cite{CVPR19/CRAFT/Baek} & CVPR'19 & VGG16-BN & SynText & TOT & 1280 $\diamond$ & 79.9 & 87.6 & 83.6 & N/A  & N/A \\  
     \cline{2-12}
     & \multirow{2}{*}{LOMO \cite{CVPR19/LOMO/Zhang}}  & \multirow{2}{*}{CVPR'19}  & \multirow{2}{*}{ResNet50} & \multirow{2}{*}{SynText} & \multirow{2}{*}{TOT} & 512 $\diamond$ & 75.7 & 88.6 &  81.6  & N/A  & \multirow{2}{*}{Tesla K40m (4)} \\ \cline{7-11}
     & & & & & & MS & 79.3 & 87.6 & 83.3 & N/A                 & \\ 
     
     \cline{2-12} 
     & \multirow{2}{*}{PSENet-1s \cite{CVPR19/PSENet/Wang} $\spadesuit$ } & \multirow{3}{*}{CVPR'19} 
     & \multirow{3}{*}{ResNet50} & $\times$  & \multirow{3}{*}{TOT} & \multirow{3}{*}{1280 $\diamond$}  & 75.1 & 81.8 & 78.3 & 3.9 & \multirow{3}{*}{N/A (4)} \\ \cline{5-5} \cline{8-11}
     &  & & & MLT17 &  & & 78.0 & 84.0 & 80.9 & 3.9 &  \\ 
     \cline{2-2} \cline{5-5} \cline{8-11}
     & PSENet-4s \cite{CVPR19/PSENet/Wang} $\spadesuit$ & & & MLT17 & & & 75.2 & 84.5 & 79.6 & 8.4 & \\ 
     \cline{2-12} 
     & MSR \cite{IJCAI19/MSR/Xue} & IJCAI'19 & ResNet50 & SynText & TOT & MS & 74.8 & 83.8 & 79.0 & N/A  &  GTX 1080Ti (2) \\ 
     \cline{2-12}
     & \multirow{6}{*}{PAN \cite{ICCV19/PAN/Wang}}  & \multirow{6}{*}{ICCV'19} & \multirow{6}{*}{ResNet18} & \multirow{3}{*}{$\times$}  & \multirow{3}{*}{TOT} 
     & 320 $\bigtriangleup$ & 71.3    & 84.0    & 77.1    & \textcolor{blue}{\textbf{82.4}}   & \multirow{6}{*}{GTX 1080Ti (4)} \\ \cline{7-11} 
     & & & & & & 512 $\bigtriangleup$ & 78.4  & 86.7 & 82.4 & 57.1 & \\ \cline{7-11}   
     & & & & & & 640 $\bigtriangleup$ & 79.4 & 88.0 & 83.5 & 39.6 &  \\ \cline{5-11}
     & & & & \multirow{3}{*}{SynText} & \multirow{3}{*}{TOT} & 320 $\bigtriangleup$ & 75.0 & 85.6 & 79.9 & \textcolor{blue}{\textbf{82.4}} & \\ \cline{7-11}
     & & & & & & 512 $\bigtriangleup$ & 79.7  & 89.4  & 84.3  & 57.1 & \\ \cline{7-11}
     & & & & & & 640 $\bigtriangleup$ & 81.0 & 89.3 & 85.0 & 39.6 &  \\ \cline{2-12}
     & TextDragon \cite{ICCV19/TextDragon/Feng} $\natural$ & ICCV'19 & VGG16 & SynText & TOT & N/A  & 75.7 & 85.6 & 80.3  & N/A  & GTX TiTan X (1) \\
     \cline{2-12}
     & \multirow{2}{*}{CharNet \cite{ICCV19/CharNet/Xing} $\natural$}  & \multirow{2}{*}{ICCV'19} & Hourglass-57  & \multirow{2}{*}{SynText} & \multirow{2}{*}{TOT} & \multirow{2}{*}{N/A} & 81.0 & 88.6 & 84.6 & N/A & \multirow{2}{*}{N/A (8)} \\ \cline{4-4} \cline{8-11} & & &
     Hourglass-88 &  & &  & 81.7 & 89.9 & 85.6 & N/A &  \\ \cline{2-12} 
     & \multirow{2}{*}{SAST \cite{MM19/SAST/Wang}}  & \multirow{2}{*}{MM'19} & \multirow{2}{*}{ResNet50} & \multirow{2}{*}{SynText} & \multirow{2}{*}{TOT} & 512 $\diamond$ &  76.9 & 83.8 & 80.2 & \multirow{2}{*}{N/A} & \multirow{2}{*}{TiTan Xp (4)}\\ \cline{7-10} & & & & &  & MS &75.5 & 85.6 & 80.2 & & \\ 

     \cline{2-12}
     & TextField \cite{TIP19/TextField/Xu}  $\spadesuit$  & TIP'19 & VGG16 & SynText & TOT & 768$\times$768 & 79.9 & 81.2    & 80.6  & N/A  & TiTan Xp (1) \\ 
     \cline{2-12} 
     & \multirow{2}{*}{DB \cite{AAAI20/DB/Liao}}  & \multirow{2}{*}{AAAI'20} & ResNet18-DC  & \multirow{2}{*}{SynText} & \multirow{2}{*}{TOT} & \multirow{2}{*}{800 $\star$} & 77.9 & 88.3 & 82.8  & 50.0 & \multirow{2}{*}{GTX 1080 Ti (1)} \\ \cline{4-4} \cline{8-11} & & &
      ResNet50-DC &  & &  & 82.5 & 87.1 & 84.7 & 32.0 &  \\ \cline{2-12} 
      & TextPerceptron\cite{AAAI20/TextPerceptron/Qiao} $\natural$  & AAAI'20 & ResNet50 & SynText & TOT & 1350 $\diamond$  & 81.8    & 88.8    & 85.2  & N/A  & Tesla V100 (8) \\
     \cline{2-12} 
     & DRRGN \cite{CVPR20/DRRGN/Zhang} & CVPR'20 & VGG16 & MLT17 & TOT & (512,1280)  & 84.9   & 86.5    & 85.7  & N/A  & RTX 2080Ti (1) \\ 
     \cline{2-12} 
     & \multirow{2}{*}{CRNet \cite{MM20/CRNet/Zhou}}  & \multirow{2}{*}{MM'20} & \multirow{2}{*}{ResNet50} & \multirow{2}{*}{SynText} & \multirow{2}{*}{TOT} & 1536 $\diamond$ & 82.5    & 85.8    & 84.1  & \multirow{2}{*}{N/A} & \multirow{2}{*}{TiTan X (1)}\\ \cline{7-10} & & & & &  & MS & 84.2    & 85.1    & 84.6  & & \\ 
     \cline{2-12}
     & TextRay\cite{MM20/TextRay/Wang} $\spadesuit$ & MM'20 & ResNet50 & $\times$ & TOT & 960$\times$960 & 77.9    & 83.5  & 80.6  & N/A  & TiTan X (4) \\ 
    \cline{2-12} 
     & FCENet \cite{FCENet} & CVPR'21 &  ResNet50-DCN-FPN &  $\times$ & TOT & (960,1280) & 82.5 & 89.3 & 85.8 & N/A & RTX 2080Ti(2)  \\ 
     \cline{2-12} 
     & SRSTS \cite{wu2022decoupling} & MM'22 & ResNet50-BiFPN 
     & \makecell{Synth800K\\+Bezier Curve Synthetic}
     & TOT & N/A & 82.3 & \textcolor{blue}{\textbf{92.0}} & 87.2  & 18.7 & RTX 3090Ti
     \\ 
     \cline{2-12} 
     & FSGNet  \cite{FSGNet} & CVPR'22 & ResNet50 
     & MLT+Bezier Curve Synthetic
     & TOT & N/A & 85.7 & 90.7 & 88.1  & 12.9 &  N/A 
     \\ 
     \cline{2-12} 
     & Boundary-TextSpotter \cite{Boundary-TextSpotter22} & TIP'22 & ResNet50 & SynText & TOT & N/A & 81.2 & 89.6 & 85.2  & 13.4 & Nvidia V100 (4) \\
     \cline{2-12} 
     & TPSNet \cite{TPSNet2022} & MM'22 & ResNet50-DCN-FPN & Synth150K+MLT17  & TOT & N/A & 86.8 & 88.5 & \textcolor{blue}{\textbf{90.2}}  & 9.3 & RTX 3090 
     \\ 
     \cline{2-12} 
     & \multirow{2}{*}{FSAS \cite{FSAS}}   
     & \multirow{2}{*}{TIP'23}  & ResNet18-FPN
     & \multirow{2}{*}{$\times$}
     & \multirow{2}{*}{TOT} 
     & \multirow{2}{*}{1333$\times$800}  & 77.0 & 85.8 & 81.1 & 33.5 
     & \multirow{2}{*}{Titan X Pascal (4)} \\ 
     \cline{4-4}
     \cline{8-11}
     & & & ResNet50-FPN & & & & 79.9 & 88.7 & 84.1 & 24.3 & \\
     \cline{2-12} 
     & ADNet  \cite{ADNet} & TMM'23 &  ResNet50-FPN
     & \makecell{SynText} 
     & TOT & 800 $\bigtriangleup$  & 84.4 & 90.6 & 87.4 & 18.0 
     &  GTX 1080Ti (1)      \\
     \cline{2-12}
     & MorphText\cite{TMM23/Morphtext/Xu} & TMM'23 & ResNet50-FPN & SynthText & TOT & (800,800)  & 85.2  & 90.6  & 87.8 & 1.8  & RTX 3090 (1) \\
     \cline{2-12} 
     & RSMTD\cite{TMM23/RSMTD/Yang} & TMM'23 & ResNet50 & SynText & TOT & 640 $\bigtriangleup$ & 83.8  & 88.5  & 86.1 & 70.9  & GTX 1080Ti (2) \\
     \cline{2-12}
     & HFENet \cite{TITS23/HFENet/Liang} & TIPS’23 & ResNet18 & SynText & TOT & 640$\times$640 & 81.7 & 85.7 & 83.7 & 22.0 & GeForce GTX 1080Ti(1) \\ 
     \cline{2-12} 
     & SIR  \cite{SIRACMM23} & MM'23 & ResNet18 & Synth150K
     & TOT & N/A & 83.1 & 91.8 & 87.2  & 58.0 & RTX 2080Ti \\ 
     \cline{2-12} 
     & HFENet\cite{TITS23/HFENet/Liang} & TITS'23 & ResNet50 & SynthText
     & TOT & 1024 $\star$  & 84.0 & 89.0 & 86.4  & 12.2 & GTX 1080Ti (1) \\ 
     \cline{2-12} 
     
     \cline{2-12} 
     & \multirow{2}{*}{DBNet++ \cite{DBNet++}}
     & \multirow{2}{*}{TPAMI'23} & ResNet18-DCN
     & \multirow{2}{*}{SynText} 
     & \multirow{2}{*}{TOT} 
     & \multirow{2}{*}{800 $\star$ }  & 79.6 & 87.4 & 83.3 & 48.0 
     & \multirow{2}{*}{1080ti} \\
     \cline{4-4}
     \cline{8-11}
     & & & ResNet50-DCN & & &  & 83.2 & 88.9 & 86.0 & 28.0 & \\

     \cline{2-12} 
     & \multirow{2}{*}{ZTD \cite{TNNLS23/ZTD/Yang}}  & \multirow{2}{*}{TNNLS'23} & \multirow{2}{*}{ResNet18} & \multirow{2}{*}{SynText} & \multirow{2}{*}{TOT} & 512 $\bigtriangleup$ &  80.6  & 90.5  & 85.3 & 93.2 & \multirow{2}{*}{GTX 1080Ti(1)}\\ \cline{7-11} & & & & &  & 640 $\bigtriangleup$ & 82.3 & 90.1 & 86.0 & 75.2 & \\

     \cline{2-12} 
     & RP-Text \cite{TMM23/RP-Text/Wang} & TMM'23 & ResNet18
     & \makecell{SynText} 
     & TOT & 640 $\bigtriangleup$  & 82.8 & 89.4 & 86.0 & 31.7
     & N/A\\ 

     \cline{2-12}
     & \multirow{2}{*}{KAC \cite{zheng2024kernel}}
     & \multirow{2}{*}{CVPR'24}
     & ResNet18
     & \multirow{2}{*}{SynText}
     & \multirow{2}{*}{CTW}
     & \multirow{2}{*}{640×640}
     & 83.4 & 90.2 & 86.7 & 56.6
     & \multirow{2}{*}{NVIDIA TITAN RTX (1)} \\
     \cline{4-4}
     \cline{8-11}
     & & & ResNet50 & & & & 85.4 & 88.6 & 86.8 & 19.2 & \\
     
     \cline{2-12} 
     & CBNet \cite{IJCV24/CBNet/Zhao} & IJCV'24 & ResNet18 & SynthText+ICDAR-MLT19 & TOT & N/A & 79.1 & 90.1 & 84.5 & N/A & NVIDIA Tesla P40 (2) \\ 
     \cline{2-12}
     & CIC \cite{TCSVT25/CIC/Han} & TCSVT'25 & ResNet50 & SynthText+ICDAR+MLT17 & TOT & 736$\triangle$ & \textcolor{blue}{\textbf{90.2}} & 85.5 & 87.8 & 16.7 & GTX 1080Ti(1) \\ 
     \cline{2-12}
     & TextRSR \cite{TMM26/TextRSR/Shao} & TMM'26 & ResNet-DCN-FPN & SynthText 150K & TOT & 800,1600,1200,1000 & 86.5 & 91.7 & 89.1 & 23.1 & NVIDIA RTX 3090 GPU (1) \\ 

     \hline
     \end{tabular}
     }
  \end{center}
\end{table*}

\section{Inconsistencies Analyses} \label{sec:incons_ana}

\subsection{Data Augmentation}
Scene text detection methods with deep neural networks are data-driven. Data augmentation thus plays a significant role in learning robust models. However, different data augmentation strategies may affect the performance  in the model in complex scenarios. For example,  the top-down based methods usually resize the original input image such that the shorter side to different sizes (e.g., [800,1000,1200] \cite{IJCV22/I3CL/Du}, [480, . . . , 832] \cite{AAAI24/Box2Poly/Chen}, etc.) while the longer side is set within a fixed scale (e.g., 1800 \cite{IJCV22/I3CL/Du}, 1600 \cite{AAAI24/Box2Poly/Chen}). Differently, the bottom-up based methods randomly crop the sub-image from the input image with different sizes (\textit{e.g.}, 512$\times$512 \cite{MM19/SAST/Wang,CVPR24/ODM/Duan}, 640$\times$640 \cite{CVPR19/SAE/Tian,CVPR19/PSENet/Wang,AAAI20/DB/Liao,CVPR24/KACM/Zheng}, 768$\times$768 \cite{TIP19/TextField/Xu}, 960$\times$960 \cite{MM20/TextRay/Wang}, etc.). This strategy makes the model have different generalization abilities to the scales of texts. 

Meanwhile, some existing methods randomly rotate the input image with different angles (\textit{e.g.}, \{0$^{\circ}$, 90$^{\circ}$, 180$^{\circ}$, 270$^{\circ}$\} \cite{MM19/SAST/Wang}, [-10$^{\circ}$, 10$^{\circ}$] \cite{CVPR19/PSENet/Wang,AAAI20/DB/Liao,TIP20/MTTD/Liu}, [0$^{\circ}$, 360$^{\circ}$) \cite{AAAI24/LRANet/Su,AAAI25/ERRNet/Sun}, etc.),  pursuing the robustness to the rotated arbitrary-shape scene texts. 
In addition, some existing methods also randomly add noise and change the brightness with different hyperparameters, which also influence the performance. These additional data augmentation strategies often bring great performance improvements, which hinder our exploration of novel research paradigms.

\subsection{Training Data}
Some existing methods directly train the model on real-world training data. Differently, some methods first utilize the synthetic dataset SynText \cite{CVPR16/FCN_Text/Zhang} or the large-scale real-world dataset (\textit{e.g.}, MLT17 \cite{MLT17/competition/Nayef}, COCO-Text \cite{ICDAR17/COCO_Text/Gomez}, etc.) to pre-train the model, and then the model is fine-tuned on the real-world training data. Table \ref{table111} and Table \ref{table222} show that pre-training on external data can bring obvious performance improvements. 

For example, pre-training on MLT17 achieves the improvement of 4.2\% (82.2\% vs. 78.0\%) in $\textit{F-measure}$ on CTW datasets for PSENet-1s \cite{CVPR19/PSENet/Wang} (Table \ref{table111}).\;Similarly,  pre-training on SynText has improved the \textit{F-measure} of about 1.5\% (85.0\% vs. 83.5\%) in PAN \cite{ICCV19/PAN/Wang} (Table \ref{table222}). When using the synthetic curved scene texts Synth150k \cite{CVPR20/ABCNet/Liu}, the top-down based methods (e.g., Box2Poly \cite{AAAI24/Box2Poly/Chen}, SRFormer \cite{AAAI24/SRFormer/Bu}, etc.) and the bottom-up based methods (e.g., LRANet \cite{AAAI24/LRANet/Su}, ERRNet \cite{AAAI25/ERRNet/Sun}, etc.) can both achieve better performance compared with those using the original synthetic dataset SynText \cite{gupta2016synthetic}. Note that SynText \cite{gupta2016synthetic} contains 800k synthetic images, and most of the text instances in each image are strict text, labeled by the quadrilateral bounding box. Synth150k \cite{CVPR20/ABCNet/Liu} diversifies and enriches the oriented or curved scene text via the same synthetic approach as SynText \cite{gupta2016synthetic}, generating 150k synthesized images, in which 94,723 images mainly contain straight texts, and 54,327 images mostly contain curved text.
Besides, existing methods also adopt different pre-training epochs, which may also result in obviously different performance, but existing methods usually do not care about these inconsistencies, when compared with other methods. This will precisely ignore the performance comparison between methods and lose the rationality of the comparison.

\subsection{Testing Scales}
In the inference stage, various testing sizes of the input image can obviously affect the performance and speed of the model, as displayed in Table \ref{table111} and Table \ref{table222}. For example, for typical top-down methods, when adjusting the long side to 1024, SDM \cite{ECCV20/SDM/Xiao} achieves a 1.5\% rise in $\textit{F-measure}$ on the MS dataset; while for MS-CAFA \cite{TMM19/MS-CAFA/Dai} with the short side setting to 600 and keeping the longer side not larger than 800, its $\textit{F-measure}$ on the MS dataset is improved by 1.2\% (85.4\% vs. 84.6\%). For the well-known bottom-up based methods, PAN \cite{ICCV19/PAN/Wang} that does not pre-train on SynText, as the short size of the input image increases, the $\textit{F-measure}$ increases by 3.9\% (81.0\% vs. 77.1\%) and 6.4\% (83.5\% vs. 77.1\%) on the datasets CTW and TOT, respectively. 

Moreover, some methods even use the multi-scale testing strategy to promote performance. However, they may only have a slight improvement in performance, but consequently bring a decrease in speed. For example, the recent work CRNet \cite{MM20/CRNet/Zhou} with the multi-scale testing strategy (Table \ref{table111}) only increases 0.3\% (84.2\% vs. 83.9\%) in $\textit{F-measure}$, compared with the previous method ContourNet \cite{CVPR20/ContourNet/Wang} using the single-scale testing strategy. At the same time, compared with its own fixed size, it only increased by 0.4\% (84.2\% vs. 83.8\%). These show that pure multi-scale testing only brings minimal improvements, and they also burden the inference speed.

\begin{figure*}[htbp]
  \begin{center}
\includegraphics[width=1\linewidth,height=0.42\linewidth]{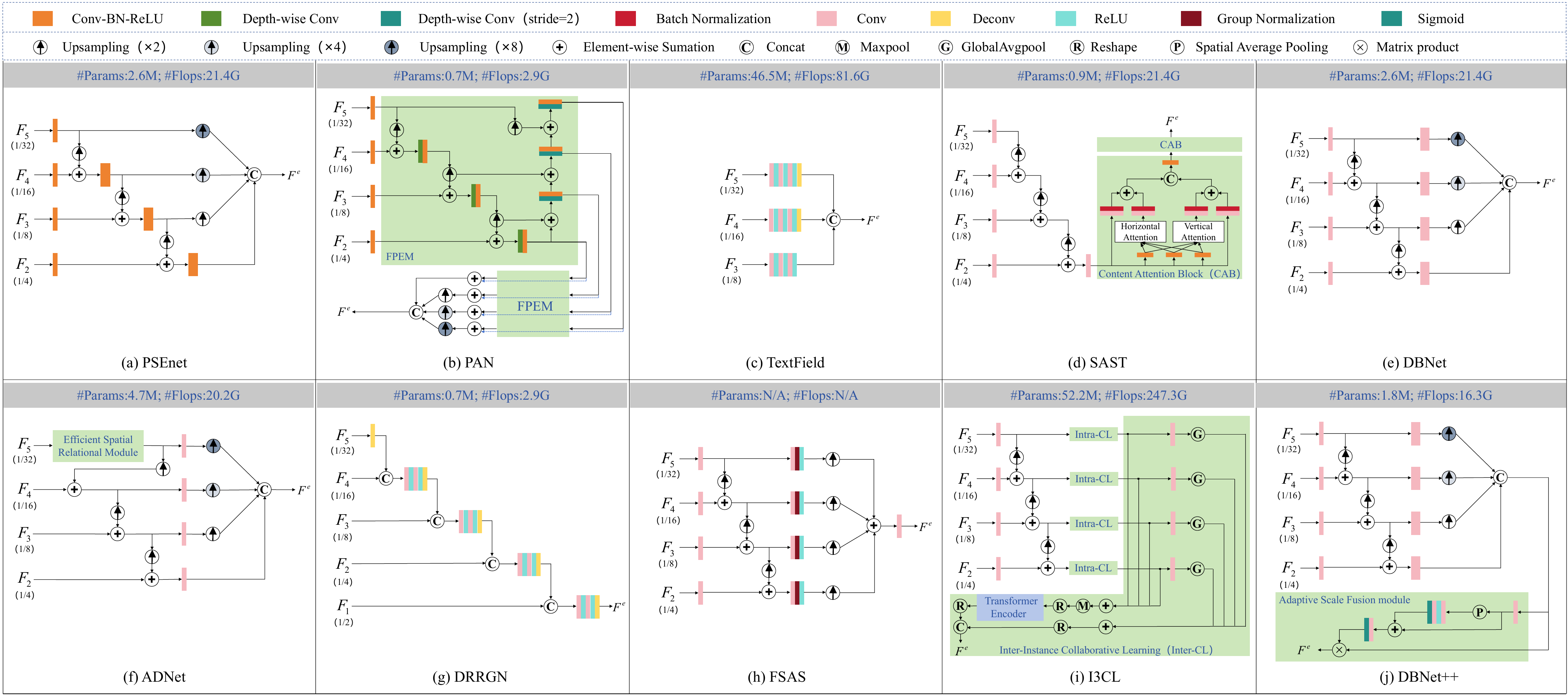}
  \end{center}
  \caption{Illustration of the multi-scale feature fusion modules for (a) PSENet \cite{CVPR19/PSENet/Wang}, (b) PAN \cite{ICCV19/PAN/Wang}, (c) TextField \cite{TIP19/TextField/Xu}, (d) SAST \cite{MM19/SAST/Wang}, (e) DBNet \cite{AAAI20/DB/Liao} and (f) ADNet \cite{tmm23adnet}, (g) DRRGN \cite{CVPR20/DRRGN/Zhang}, (h) FSAS \cite{tip23fsa}, (i) I3CL \cite{du2022i3cl}, (j) DBNet++ \cite{22dbnet++}. The number of parameters (\#Parameters) and the number of floating-point operations per second (\#Flops) are reported based on the input size of $640\times640\times3$ and the multi-scale features generated by the backbone network ResNet50. N/A indicates a method with unpublished code.}
  \label{fig:ffm}
\end{figure*}

\subsection{Evaluation Protocols}
When evaluating on TOT \cite{IJDAR20/TOT/chng}, it involves three kinds of evaluation protocols (\textit{e.g.}, IOU@0.5, DetEval-v1, and DetEval-v2). They are defined as follows.

\textbf{DetEval-v1} is a classic text detection evaluation metric based on area coverage constraints.

The main differences between  DetEval-v1 and DetEval-v2 are: i) The latter does not care about the texts annotated by `\#' in evaluation. ii) In the matching process of the detection and the ground truth, they adopt different matching thresholds. In addition, the evaluation metric IOU@0.5 in this dataset is also not concerned with the ground-truth texts annotated by `\#'. 

Some of the existing methods, however, do not indicate which kind of evaluation protocol they use, which makes the comparison very confusing. Moreover, as shown in Table \ref{table222}, prior works report the performances under different evaluation protocols and make comparisons with each other, and the performances reported under different evaluation protocols have obvious differences. Thus, these comparisons are unreasonable, making the assertion of achieving state-of-the-art unbelievable for some methods.

\subsection{Backbones}
As shown in Table \ref{table111} and Table \ref{table222}, previous methods utilize different backbone networks (\textit{e.g.}, VGG16, ResNet50, ResNet101, etc.) pre-trained on ImageNet \cite{IJCV15/ImageNet/Russakovsky} to extract visual features.\; Stronger ones usually bring better performance. For the top-down based methods, some of them \cite{TMM19/MS-CAFA/Dai, TMM22/SADA_SSC/Dai,TIP20/Mask-TTD/Liu,IJCV22/I3CL/Du} usually leverage the ResNet-FPN as the backbone, and then generate the predictions at each feature level. For the bottom-up based methods, DBnet \cite{AAAI20/DB/Liao}, using ResNet50 with deformable convolution networks (termed `ResNet50-DCN') as the backbone, achieves a 2.4\% improvement(83.4\% vs. 81.0\%) in $\textit{F-measure}$ on CTW datasets, compared with ResNet18-DCN. Similarly, using ResNet101 also increases the $\textit{F-measure}$ by 1.2\% (86.6\% vs. 85.4\%) compared to using ResNet50 in TextFuseNet \cite{IJCAI20/TextFuseNet/Ye} on CTW datasets. Even though some works adopt the same backbone networks, different versions of networks may also influence the performance. For example, PAN \cite{ICCV19/PAN/Wang} uses the ResNet-v2 \cite{ECCV16/ResNet_v2/He}\footnote{We found this from their official code: \url{https://github.com/zhongqianli/pan_pp.pytorch}} while other methods usually adopt ResNet-v1 \cite{CVPR16/ResNet_v1/He}.

\subsection{Multi-scale Feature Fusion}
Arbitrary-shape scene text detectors usually involve fusing multi-scale features generated by the backbone network, before feeding them into the prediction head. This fusion between low-level and high-level features can enrich the feature representations, facilitating the model to detect texts with various scales. However, existing methods usually introduce different fusion techniques, which further make it difficult to fairly investigate the representations of describing texts. These different fusion techniques not only influence the performance, but also result in different memory and speed.

Besides, even though the module adopts similar fusion strategies, it could still bring different performance gains, due to different resolutions of the output maps. For example, in PSENet \cite{CVPR19/PSENet/Wang},  when the resolution of the output map is 1/4 of the input image (termed as `PSENet-4s'), it decreases by 2.3\% (82.2\% vs. 79.9\%) and 1.3\% (80.9\% vs. 79.6\%) in $\textit{F-measure}$ for the datasets CTW and TOT, respectively, compared with `PSENet-1s'.  As shown in Figure \ref{fig:ffm}, we display the feature fusion modules of several methods. We can observe that these modules not only have different numbers of parameters but also involve various numbers of floating-point operations per second. Due to the elaborate design, PAN \cite{ICCV19/PAN/Wang} and SAST \cite{MM19/SAST/Wang} have fewer parameters. Especially for PAN \cite{ICCV19/PAN/Wang}, it has the fewest parameters (0.7M) and floating-point operations (2.9G). Although TextField \cite{TIP19/TextField/Xu} has a simpler network structure, it involves larger filters and a larger number of filters in the convolution layers and deconvolution layers, resulting in more parameters and floating-point operations. Generally speaking, these differences could influence the performance and speed of models, which would hinder our explorations of the core techniques (that is, the representations of describing scene texts) of existing models.

\begin{figure*}[t]
  \begin{center}
     \includegraphics[width=1.0\linewidth,height=0.29\linewidth]{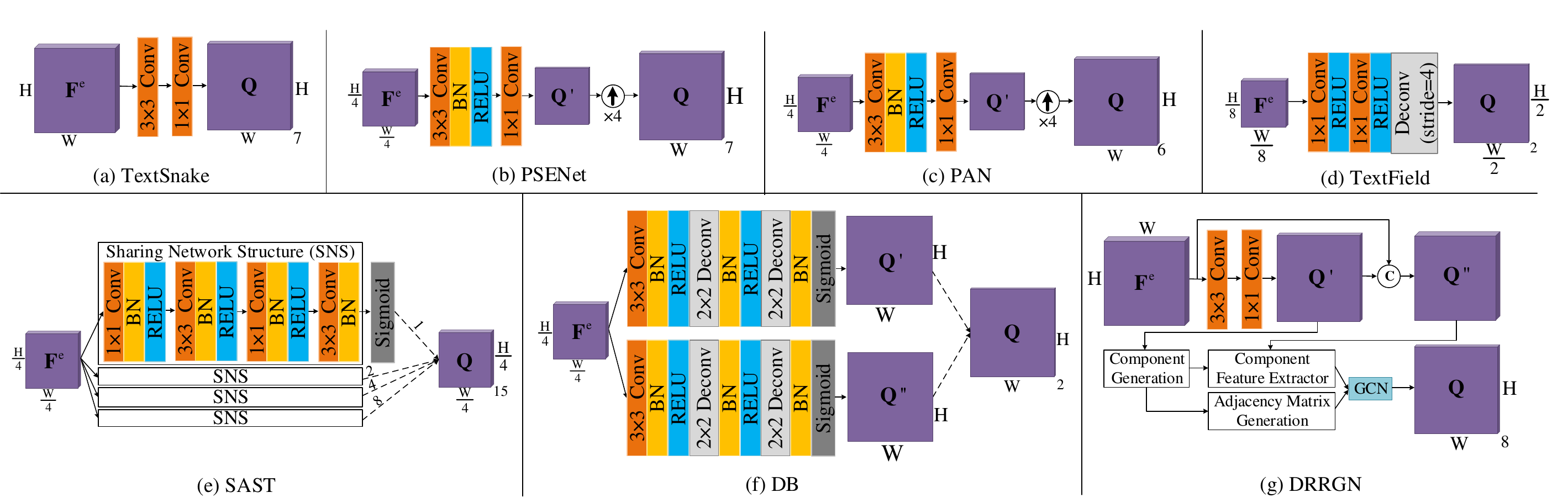}
  \end{center}
  \caption{Illustration of the prediction heads for (a) TextSnake \cite{ECCV18/Textsnake/Long}, (b) PSENet \cite{CVPR19/PSENet/Wang}, (c) PAN \cite{ICCV19/PAN/Wang}, (d) TextField \cite{TIP19/TextField/Xu}, (e) SAST \cite{MM19/SAST/Wang}, (f) DBNet \cite{AAAI20/DB/Liao} and (g) DRRGN\cite{CVPR20/DRRGN/Zhang}.}
  \vspace{-0.2cm}
  \label{fig:prediction_head}
\end{figure*}

\subsection{Prediction Heads}\label{sec:detail_pd}
The core techniques of existing arbitrary-shape scene text detection models are mainly reflected in the prediction heads. The prediction head module of the bottom-up-based method focuses on estimating relevant elements for localizing different text instances. 

As shown in Figure \ref{fig:prediction_head}, different methods estimate different elements, which directly indicate the ability to localize the arbitrarily-shaped scene texts. For example, the prediction head output $\textbf{Q} \in \mathbb{R}^{H \times W \times 7}$ of TextSnake \cite{ECCV18/Textsnake/Long} consists of the classification logits of text regions (the channel $C=2$),  the classification logits of text center lines ($C=2$) and the geometry attributes (\textit{e.g.}, the radius of disk, the $sine$ operation of disk angles, and the $cosine$ operation of disk angles). In PSENet \cite{CVPR19/PSENet/Wang}, $\textbf{Q} \in \mathbb{R}^{H \times W \times 7}$ consists of 7 kernel maps, which are shrunk text region maps with different scales. In PAN \cite{ICCV19/PAN/Wang}, $\textbf{Q} \in \mathbb{R}^{H \times W \times 6}$ consists of the classification logits of text regions ($C=2$), the classification logits of text kernels ($C=2$), and the similarity embeddings ($C=2$). In TextField \cite{TIP19/TextField/Xu}, $\textbf{Q} \in \mathbb{R}^{\frac{H}{2} \times \frac{W}{2} \times 2}$ denotes the direction field that encodes both binary text mask and direction information that can be used to separate adjacent text instances. In SAST \cite{MM19/SAST/Wang}, $\textbf{Q} \in \mathbb{R}^{\frac{H}{4} \times \frac{W}{4} \times 15}$ consists of the shrunk text center segmentation map ($C=1$), the offsets between the pixels in the text center segmentation map and the object-level centers ($C=2$), the offsets between the pixels in the text center line and four vertices of bounding quadrangle of text region ($C=8$), and the offsets of the pixels in text center lines to the upper borders and lower borders ($C=4$). In DBNet \cite{AAAI20/DB/Liao}, $\textbf{Q} \in \mathbb{R}^{H \times W \times 2}$ consists of the probability map and the threshold map. In DRRGN \cite{CVPR20/DRRGN/Zhang}, $\textbf{Q}' \in \mathbb{R}^{H \times W \times 8}$ is similar to that of TextSnake \cite{ECCV18/Textsnake/Long}, except that DRRGN \cite{CVPR20/DRRGN/Zhang} estimates the distances to the top border and the bottom border instead of the radius in TextSnake \cite{ECCV18/Textsnake/Long}. Based on $\textbf{Q}'$, DRRGN \cite{CVPR20/DRRGN/Zhang} generates text components. Note that the text components are directly obtained based on the ground truth. After that, DRRGN \cite{CVPR20/DRRGN/Zhang} employs the graph convolution network (GCN) to induce deep relationships between text components. The final output $\textbf{Q} \in \mathbb{R}^{H \times W \times 8}$  reveals the linkage logits between each text component and its four neighbors.

\section{Proposed Unified Frameworks}\label{sec:framework}
\begin{figure*}[htbp]
  \begin{center}
  \includegraphics[width=1\linewidth,height=0.36\linewidth]{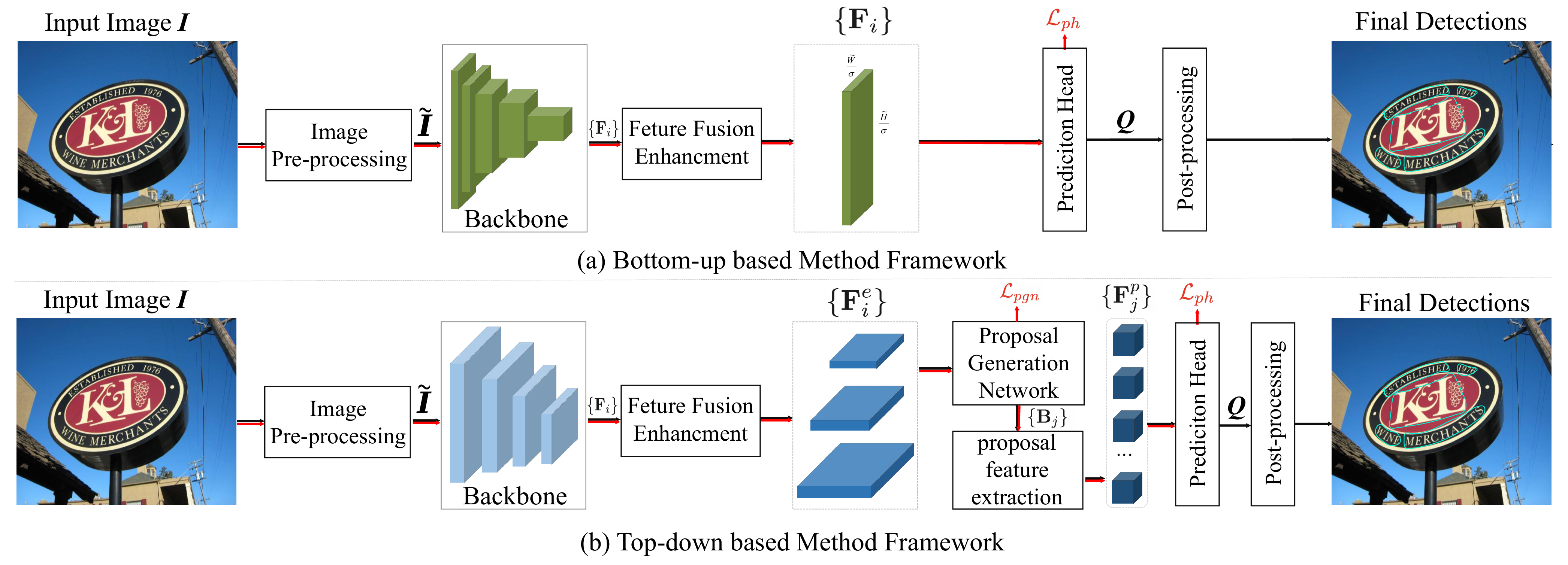}
  \end{center}
  \caption{Illustration of the unified frameworks for the arbitrary-shape scene detection models. }
  \label{fig:framework}
  \vspace{-0.2cm}
\end{figure*}

In this section, we propose a unified framework for the bottom-up and top-down based arbitrary-shape scene text detection methods, respectively, as shown in Figure \ref{fig:framework}. This bottom-up based unified framework mainly consists of five modules: image pre-processing, backbone, multi-scale feature fusion, prediction head, and post-processing. Differently, the top-down based unified framework additionally contains the proposal generation network and the ROI pooling operation.

Specifically, the original image $\textbf{I} \in \mathbb{R}^{H \times W \times 3}$  is first fed into the image pre-processing module  to generate the network input $\widetilde{\textbf{I}} \in \mathbb{R}^{\widetilde{H} \times \widetilde{W} \times 3}$, formulated as:
\begin{equation}
  \widetilde{\textbf{I}} =\mathcal{N}{(\mathcal{P} (\textbf{I}))},
\end{equation}
where $\mathcal{N}$ means image normalization; $\mathcal{P}$ refers to the data augmentation and the test image resizing strategy for the training and testing stages, respectively. Due to the differences in $\mathcal{P}$ in existing methods, we may unify the settings of $\mathcal{P}$ for fair comparisons following \cite{CVPR20/DRRGN/Zhang} for the bottom-up based method. In the training stage, the data augmentation mainly involves four steps: i)  Randomly scale the original image via the aspect ratio ranging in [0.75, 2.5]; ii) Randomly crop the image patch with the scale of 640 $\times$ 640; iii) Randomly rotate the cropped image patch with the angle of [-90$^{\circ}$, 90$^{\circ}$]. iv) Randomly flip the image in the horizontal direction with a probability of 0.5. In the testing stage, the test image resizing strategy indicates that the short side of  $\widetilde{\textbf{I}}$ is set to $s$ if it is less than $s$, while the longer side is not larger than $2s$. In the experiments, $s$ is set to 512 by default. For the top-down based methods, different from the above settings, we may resize the original image while keeping the original aspect ratios.

In the backbone module, we utilize the backbone network pre-trained on ImageNet \cite{IJCV15/ImageNet/Russakovsky} to extract multi-scale visual feature representations, which can be formulated as: 
\begin{equation}
\{\textbf{F}_{i}\}_{i=1}^{L} = \mathcal{B}({\widetilde{\textbf{I}};\;\Theta_{b}} ),
\end{equation}
where $\textbf{F}_{i} \in \mathbb{R}^{(\widetilde{H}/2^{i+\mathbb{1}(i=1)})\times(\widetilde{W}/2^{i+\mathbb{1}(i=1)})\times D_{i}}$ denotes the feature map generated by the $i$-th stage of the backbone network $\mathcal{B}$ with the pre-trained weights of $\Theta_{b}$; $L$ is the number of multi-scale features; $\mathbbm{1}$ means the indicator function. $D_{i}$ is the dimension of the feature. In the framework, we adopt the frequently-used backbone network ResNet50 \cite{CVPR16/ResNet_v1/He} or ResNet50-FPN \cite{CVPR17/ResNet50-FPN/Lin}, like most existing scene text detectors.

Next, the feature fusion enhancement network $\phi$ combines low-level and high-level features to generate a more representative feature $\textbf{F}^{e} \in \mathbb{R}^{(\widetilde{H}/\sigma)\times(\widetilde{W}/\sigma)\times{D_{e}}}$ for the bottom-up based methods. It can be expressed as, 
\begin{equation}
  \textbf{F}^{e} = \phi(\{\textbf{F}_{i}\}_{i=1}^{L};\;\Theta_{f}),
\end{equation}
where $\sigma$ and $D_{e}$ denote the downsampling factor and dimension of feature maps;  $\Theta_{f}$ indicates the learnable parameters in $\phi$. Additionally, $\phi$ usually adopts a top-down fusion strategy. That is, it gradually fuses feature maps from deep semantic representations to shallow local cues.\; However, as shown in Figure \ref{fig:ffm}, $\phi$ in most previous methods has obvious differences, which can result in different performances. To avoid the influence of different $\phi$, we set the frequently-used fusion strategy like those in \cite{ECCV18/Textsnake/Long,CVPR20/DRRGN/Zhang}, where $\sigma$ is equal to 1. For the top-down based methods, $\phi$ usually outputs multi-scale features $\{\textbf{F}^{e}_{i}\}_{i=1}^{N}$ that are used to capture different-scale scene texts, where $N$ is the number of multi-scale features.

After that, for the bottom-up based methods, the enhanced feature $\textbf{F}^{e}$ is fed into a single prediction head module $\varphi$ to estimate the local unit categories and their auxiliary information, termed as $\textbf{Q}\in \mathbb{R}^{\frac{\sigma'\widetilde{H}}{\sigma} \times \frac{\sigma' \widetilde{W}}{\sigma} \times C_{o}}$, formulated as : 
\begin{equation}
  \textbf{Q} =\mathcal{U}_{\sigma'}(\varphi(\textbf{F}^{e}; \Theta_{p})),
\end{equation}
where $\mathcal{U}_{\sigma'}$ indicates upsampling the resolution of the output by a factor of $\sigma'$.\;$\Theta_{p}$ denotes the learnable parameters in $\varphi$.\;$C_{o}$ is the number of predictions (\textit{e.g.}, categories and auxiliary information). For example, in PSENet \cite{CVPR19/PSENet/Wang}, $\textbf{Q}$ consists of $C_{o}$ segmentation masks for the text instances at different shrunk scales. It is worth noting that PSENet-1s ($\sigma'$/$\sigma$ = 4/4 = 1) achieves better performance than PSENet-4s ($\sigma'$/$\sigma$ = 1/4), due to different resolutions of the outputs. Similarly, $\textbf{Q}$ in DBNet \cite{AAAI20/DB/Liao} contains the predicted shrunk text region map and the estimated threshold map, and its resolution is also upsampled to the same as that of the network input ($\sigma'$/$\sigma$ = 4/4 = 1).  For a fair comparison, we set $\sigma'$ = $\sigma$ in experiments. 

Differently, for the top-down based methods, the unified framework employs the proposal generation network to obtain proposals, followed by the ROI pooling to project the various-scale proposals into fixed features, which are formulated as,
\begin{equation}
    \{\mathcal{B}_{j}\}_{j=1}^{M} = \phi_{pgn}(\{\textbf{F}_{i}^{e}\},\Theta_{pgn}),
\end{equation}
where $\mathcal{B}_{j}$ means the $j$-th predicted bounding box. $\phi_{pgn}$ is the proposal generation network with the learnable parameters $\Theta_{pgn}$. Then, the fixed-scale feature of each proposal is formulated as,
\begin{equation}
   \textbf{F}_{j}^{b} = \Phi_{b}(\mathcal{B}_{j}, \{\textbf{F}_{i}\}_{i=1}^{N}),\;j=1, ..., M,
\end{equation}
where $\Phi_{b}$ is the proposal feature extraction operation, denoted as $\textbf{Q}=\{\textbf{F}^{b}_{j}\}_{j=1}^{M}$. For example, some methods (e.g., MS-CAFA \cite{TMM19/MS-CAFA/Dai}, MTTD \cite{TIP20/MTTD/Liu}, I3CL \cite{IJCV22/I3CL/Du}, etc.) utilize the ROI pooling or ROI Align techniques as $\Phi_{b}$, while  some methods (e.g., PCR \cite{PCR21}, FCENet \cite{CVPR21/FCENet/Zhu}, TextBPN++ \cite{TextBPN++}, etc.) utilize $\Phi_{b}$ to extract the contour feature representation of the proposal.

Finally,  after obtaining the network output $\textbf{Q}$, it is directly fed into the loss function to calculate the loss in the training stage, or is post-processed to generate final detection results in the testing stage. Generally, most existing scene text detectors mainly focus on the exploration of representations of describing text instances, which is reflected in the prediction head module $\varphi$ and its corresponding loss function.

\section{Experiments and Analyses} \label{sec:experiments}
To verify the influences of the inconsistent settings, we emphatically investigate the prediction head and fusion mechanism of several well-known bottom-up based methods, to profoundly disclose the advances and shortcomings of current research on arbitrary-shape scene text detection.

\subsection{Datasets and Evaluation Metrics}

\begin{table}[t]
\caption{Fair Comparisons of performances on CTW. The speed denotes the average over three runs. }
\label{table3}
\centering
\resizebox{\columnwidth}{!}{%
\begin{tabular}{l|lllllll}
\toprule
 & \multicolumn{3}{c}{IOU@0.5} & \multicolumn{3}{c}{DetEval-v2} &  \\ \cline{2-7}
\multirow{-2}{*}{Method} & \multicolumn{1}{c}{R (\%)} & \multicolumn{1}{c}{P (\%)} & \multicolumn{1}{c}{F (\%)} & \multicolumn{1}{c}{R (\%)} & \multicolumn{1}{c}{P (\%)} & \multicolumn{1}{c}{F (\%)} & \multirow{-2}{*}{\begin{tabular}[c]{@{}l@{}}Speed\\ (FPS)\end{tabular}} \\ \toprule
{TextSnake} & 63.56 & 68.98 & 66.16 & 74.23 & 69.59 & 71.83 & 2.9 \\
{PSENet} & 74.9 & 82.63 & 78.58 & 70.03 & 75.47 & 72.65 & 1.9  \\
{PAN} & 75.95 & \textbf{88.59} & 81.78 & 75.81 & \textbf{87.40} & 81.20 & 13.2 \\
{DBNet} & 65.12 & 81.42 & 72.37 & 54.02 & 67.80 & 60.13 & 43.1 \\
{DBNet++} & 67.47 & 76.27 & 71.60 & 55.85 & 63.91 & 59.61 & \textbf{52.3}  \\
{DRRGN} & 72.56  & 76.10 & 74.29 & 67.61 & 72.12 & 69.80 & 1.9 \\
SRFormer & 78.68  & 86.47 & \textbf{82.39} & 78.44 & 81.85 & 80.11  &  9.3 \\ \hline
{TextSnake*} & 63.40 & 71.35 & 67.14 & 73.88 & 72.82 & 73.35 & 2.9 \\
PSENet* & 72.91 & 83.69 & 77.93 & 68.77 & 78.82 & 73.45 & 1.8 \\
PAN* & \textbf{88.23} & 74.54 & 80.81 & \textbf{88.55} & 75.50 & \textbf{81.51} & 32.2 \\
DBNet* & 71.54 & 78.96 & 75.07 & 65.29 & 73.10 & 68.98 & 40.0 \\
DBNet++* & 71.51  & 81.20 & 76.05 & 64.76 & 73.89 & 69.02 & 39.1 \\
\toprule
\end{tabular}%
}

\end{table}

\begin{table}[ht]
\caption{Fair Comparisons of performances on TOT. The speed denotes the average over three runs.}
\label{table4}
\centering
\resizebox{\columnwidth}{!}{%
\begin{tabular}{l|lllllll}
\toprule
 & \multicolumn{3}{c}{IOU@0.5} & \multicolumn{3}{c}{DetEval-v2} &  \\ \cline{2-7}
\multirow{-2}{*}{Method} & \multicolumn{1}{c}{R (\%)} & \multicolumn{1}{c}{P (\%)} & \multicolumn{1}{c}{F (\%)} & \multicolumn{1}{c}{R (\%)} & \multicolumn{1}{c}{P (\%)} & \multicolumn{1}{c}{F (\%)} & \multirow{-2}{*}{\begin{tabular}[c]{@{}l@{}}Speed\\ (FPS)\end{tabular}} \\ \toprule
{TextSnake} & 71.73 & 72.26 & 71.99 & 75.85 & 71.51 & 73.62 & 1.1 \\
{PSENet} & 71.37 & 80.38 & 75.61 & 71.82 & 76.72 & 74.19 & 1.3 \\
{PAN} & 72.41 & 83.43 & 77.53 & 76.42 & \textbf{84.12} & 80.09 & 10.0 \\
{DBNet} & 76.95 & 86.22 & 81.32 & 74.70 & 82.60 & 78.45 & \textbf{40.8} \\
{DBNet++} & 78.04 & 82.02 & 79.98 & 77.40 & 80.52 & 78.93 & 37.6 \\
DRRGN & 71.87 & 85.07 & 77.91 & 62.80 & 74.00 & 67.94 & 1.5 \\
SRFormer & 80.44 & \textbf{89.27} & \textbf{84.63} & 77.15  & 82.10  & 79.55  & 5.7 \\ \hline
TextSnake* & 65.83 & 72.91 & 69.19 & 71.30 & 72.29 & 71.79 & 1.5 \\
PSENet* & 70.42 & 80.67 & 75.19 & 71.20 & 79.60 & 75.17 & 1.3 \\
PAN* & \textbf{86.27} & 72.41 & 78.74 & \textbf{87.45} & 74.85 & \textbf{80.66} & 20.9  \\
DBNet* & 76.32 & 83.02 & 79.53 & 74.52 & 80.36 & 77.33 & 35.6 \\
DBNet++* & 75.36 & 83.24 & 79.11 & 76.09 & 81.90 & 78.89 & 36.8\\
\toprule
\end{tabular}%
}
\end{table}

\subsubsection{CTW \cite{PR19/CTD_CLOC/Liu}}
This dataset is a prevalent arbitrary-shape scene text detection benchmark. It contains 1,500 images (1,000 images for training and 500 images for testing) in total. The annotation of the text instance is line-level and is labeled by a polygon with 14 fixed key points.

\subsubsection{TOT \cite{IJDAR20/TOT/chng}}
This dataset is also a frequently-used benchmark for arbitrary-shape scene text detection. It consists of 1,255 training images and 300 testing images. The text instance in the image is annotated by a word-level polygon with an unfixed number of key points. 

For the dataset CTW, we adopt the official evaluation metric, IOU@0.5, to evaluate the performance of the methods. Meanwhile, to evaluate the performance on the dataset TOT, three official evaluation protocols (i.e., IOU@0.5, DetEval-v1, and DetEval-v2) are employed. In \cite{IJDAR20/TOT/chng}, the authors have illustrated that DetEval-v2 is more reasonable than DetEval-v1. To better reveal the performance difference, we thus adopt two evaluation protocols (IOU@0.5 and DetEval-v2) for each dataset. With these two evaluation protocols, we adopt the Recall (R), Precision (P), and the F-measure (F) to describe the performance of each method.

\subsection{Implementation details}
Based on the official open-source codes, we adopt the same settings for the image pre-processing, backbone, and evaluation strategy to make a fair comparison. Other settings are kept the same as the corresponding original methods. In addition, during training, the size of the input image is set to 640 $\times$ 640, and during the testing phase, the shortest side of the image is fixed to 512 pixels. At the same time, the model is directly trained for 600 epochs on the training set of the corresponding dataset. The batch size is fixed to 8 in the training stage. When testing, the batch size is set to 1 in a single thread. The optimizer and learning rate follow the settings of the original official codes. All experiments are conducted with the deep learning framework PyTorch 1.12, and on a workstation with a single RTX 3090 GPU, a 2.90GHz $\times$ 32 Intel(R) Xeon(R) Gold 6226R CPU, and 64 GB RAM.

\begin{figure*}[t]
  \centering

  \includegraphics[width=1.0\linewidth,height=0.25\linewidth]{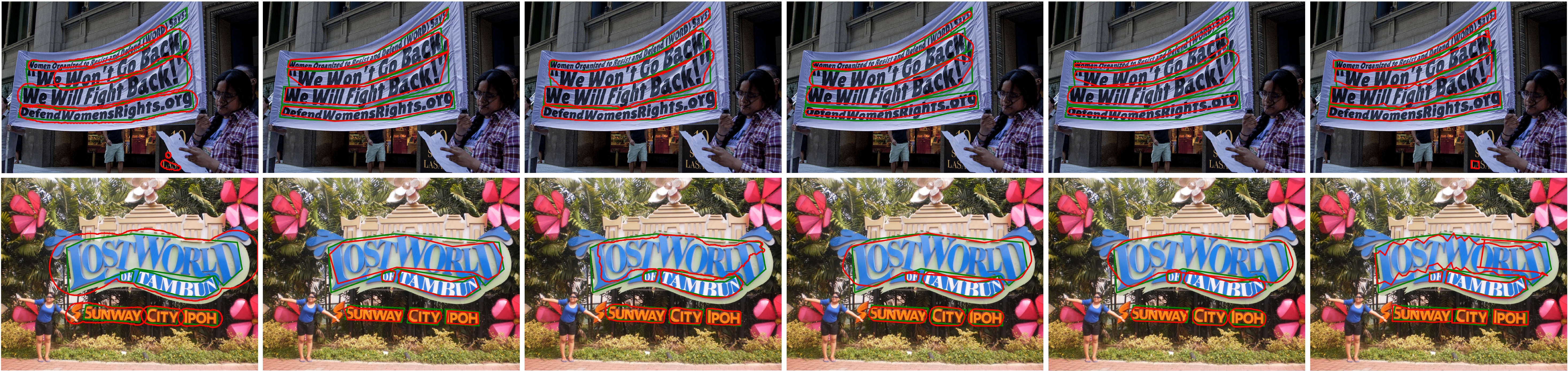}
  \setlength{\unitlength}{0.166666\linewidth} 
  \begin{picture}(6,0)
    \put(0.5,0){\makebox(0,0){\scriptsize (a) TextSnake \cite{ECCV18/Textsnake/Long}}}
    \put(1.5,0){\makebox(0,0){\scriptsize (b) PSENet \cite{CVPR19/PSENet/Wang}}}
    \put(2.5,0){\makebox(0,0){\scriptsize (c) PAN \cite{ICCV19/PAN/Wang}}}
    \put(3.5,0){\makebox(0,0){\scriptsize (d) DBNet \cite{AAAI20/DB/Liao}}}
    \put(4.5,0){\makebox(0,0){\scriptsize (e) DBNet++\cite{22dbnet++}}}
    \put(5.5,0){\makebox(0,0){\scriptsize (f) DRRGN \cite{CVPR20/DRRGN/Zhang}}}
  \end{picture}

  \vspace{0.3cm}

  \includegraphics[width=1.0\linewidth,height=0.25\linewidth]{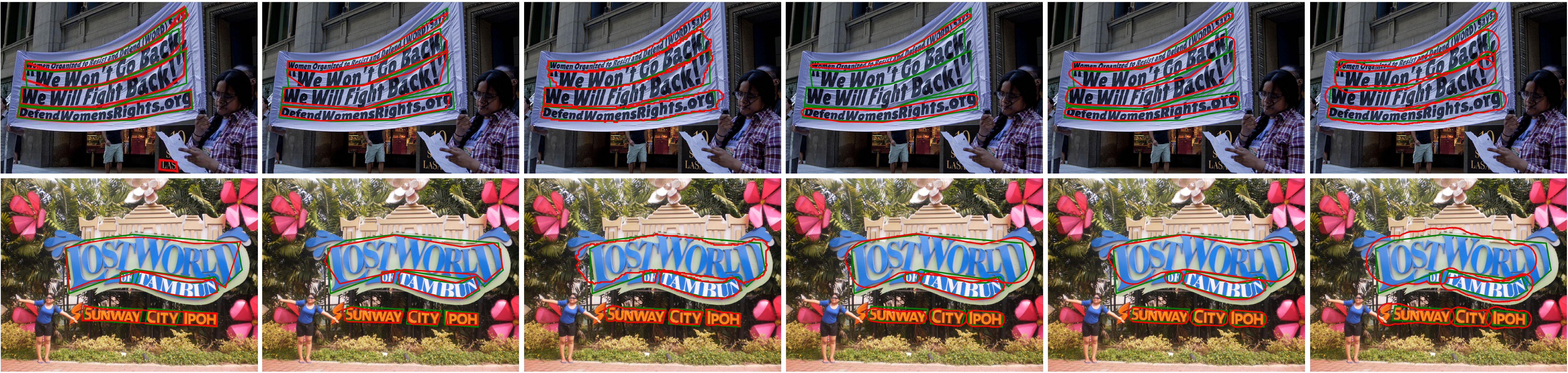}
  \setlength{\unitlength}{0.166666\linewidth}
  \begin{picture}(6,0)
    \put(0.5,0){\makebox(0,0){\scriptsize (g) SRFormer \cite{bu2023srformer}}}
    \put(1.5,0){\makebox(0,0){\scriptsize (h) TextSnake* \cite{ECCV18/Textsnake/Long}}}
    \put(2.5,0){\makebox(0,0){\scriptsize (i) PSENet* \cite{CVPR19/PSENet/Wang}}}
    \put(3.5,0){\makebox(0,0){\scriptsize (j) PAN* \cite{ICCV19/PAN/Wang}}}
    \put(4.5,0){\makebox(0,0){\scriptsize (k) DBNet* \cite{AAAI20/DB/Liao}}}
    \put(5.5,0){\makebox(0,0){\scriptsize (l) DBNet++* \cite{22dbnet++}}}
  \end{picture}

  \caption{Qualitative detection results. The samples in the first row and the second row are from the datasets CTW and TOT, respectively. The red bounding boxes denote detection results, while the green bounding boxes are the ground truth. “*” indicates the comparison of the fusion performance of various methods under a unified detection head. 
  }
  \label{fig:vis_det_results}
\end{figure*}

\begin{table}[t]
\caption{Comparison of generalization capabilities from CTW to TOT datasets.}
\label{table5}
\centering
\resizebox{\columnwidth}{!}{%
\begin{tabular}{l|llllll}
\toprule
 & \multicolumn{3}{c}{IOU@0.5} & \multicolumn{3}{c}{DetEval-v2} \\ \cline{2-7} 
\multirow{-2}{*}{Method} & \multicolumn{1}{c}{R (\%)} & \multicolumn{1}{c}{P (\%)} & \multicolumn{1}{c}{F (\%)} & \multicolumn{1}{c}{R (\%)} & \multicolumn{1}{c}{P (\%)} & \multicolumn{1}{c}{F (\%)} \\ \toprule
{TextSnake} & 27.72 & 40.22 & 32.82 & 62.03 & 54.67 & 58.12 \\
{PSENet} & 31.85 & 56.98 & 40.86 & 62.26 & 66.57 & 64.34 \\
{PAN} & 58.71 & 28.45 & 38.33 & 79.59 & 67.93 & \textbf{73.30} \\
{DBNet} & 28.81 & \textbf{59.62} & 38.85 & 44.85 & 65.63 & 53.29 \\
{DBNet++} & 31.26 & 56.24 & 40.19 & 47.66 & 60.41 & 53.28 \\
DRRGN & 34.98 & 58.94 & 43.91 & 62.62 & 66.50 & 64.50 \\
SRFormer & \textbf{62.20}  & 37.64  & \textbf{46.90} & \textbf{65.18}   & 70.16 & 67.58  \\ \hline
TextSnake* & 31.31 & 46.03 & 37.27 & 63.00 & 57.99 & 60.50 \\
PSENet* & 29.36 & 56.41 & 38.62 & 60.78 & 70.34 & 65.21 \\
PAN* & 28.99 & 57.99 & 38.66 & 66.86 & \textbf{77.02} & 71.58 \\
DBNet* & 31.03 & 55.93 & 33.92 & 52.88 & 63.24 & 57.59 \\
DBNet++* & 31.22 & 57.00 & 40.34 & 52.11 & 64.41 & 57.61 \\
\toprule
\end{tabular}%
}
\end{table}

\subsection{Comparisons of Performance and Speed}
\begin{figure}[t]
\centering

{\includegraphics[width=0.235\textwidth]{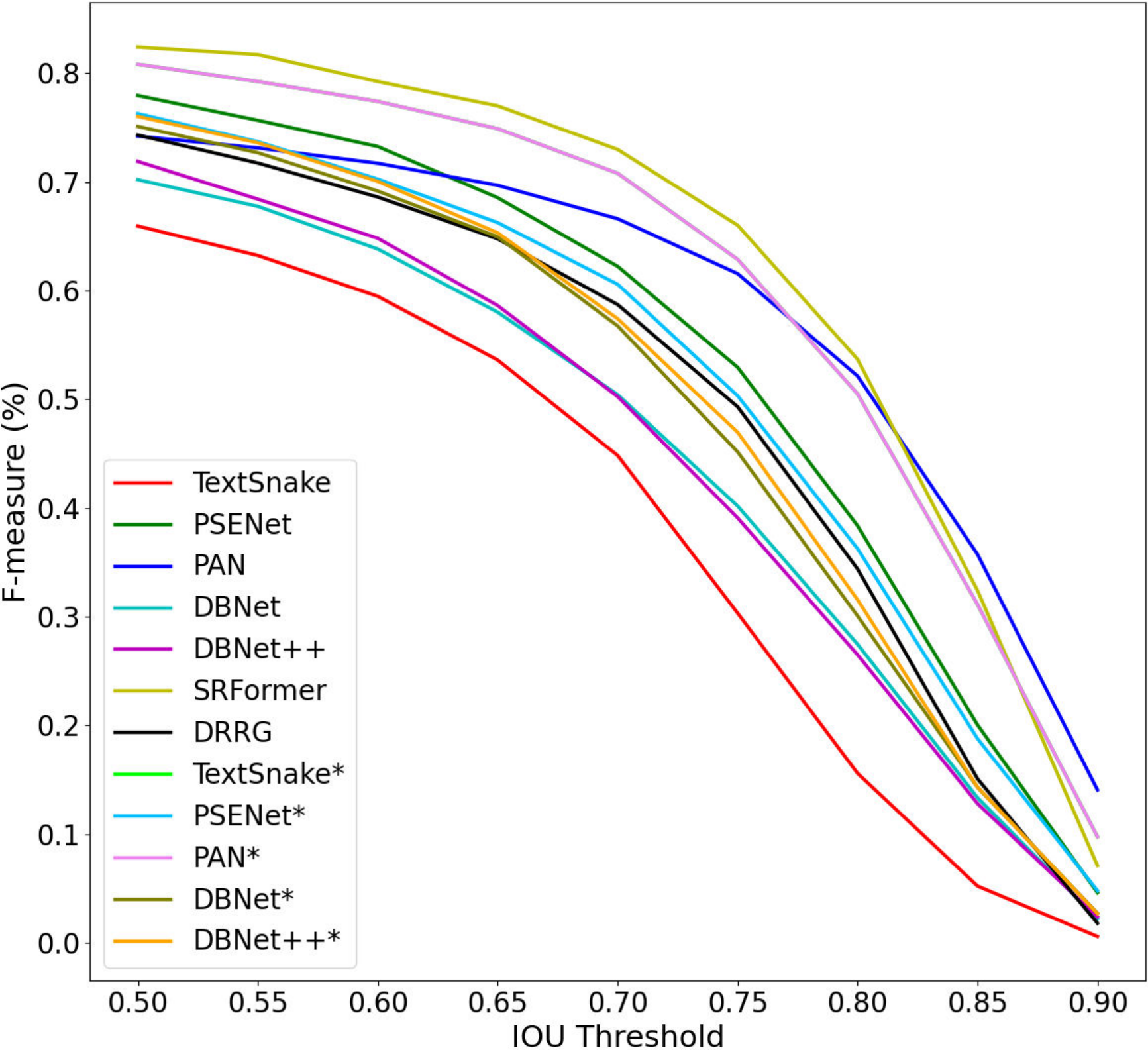}}  
{\includegraphics[width=0.24\textwidth]{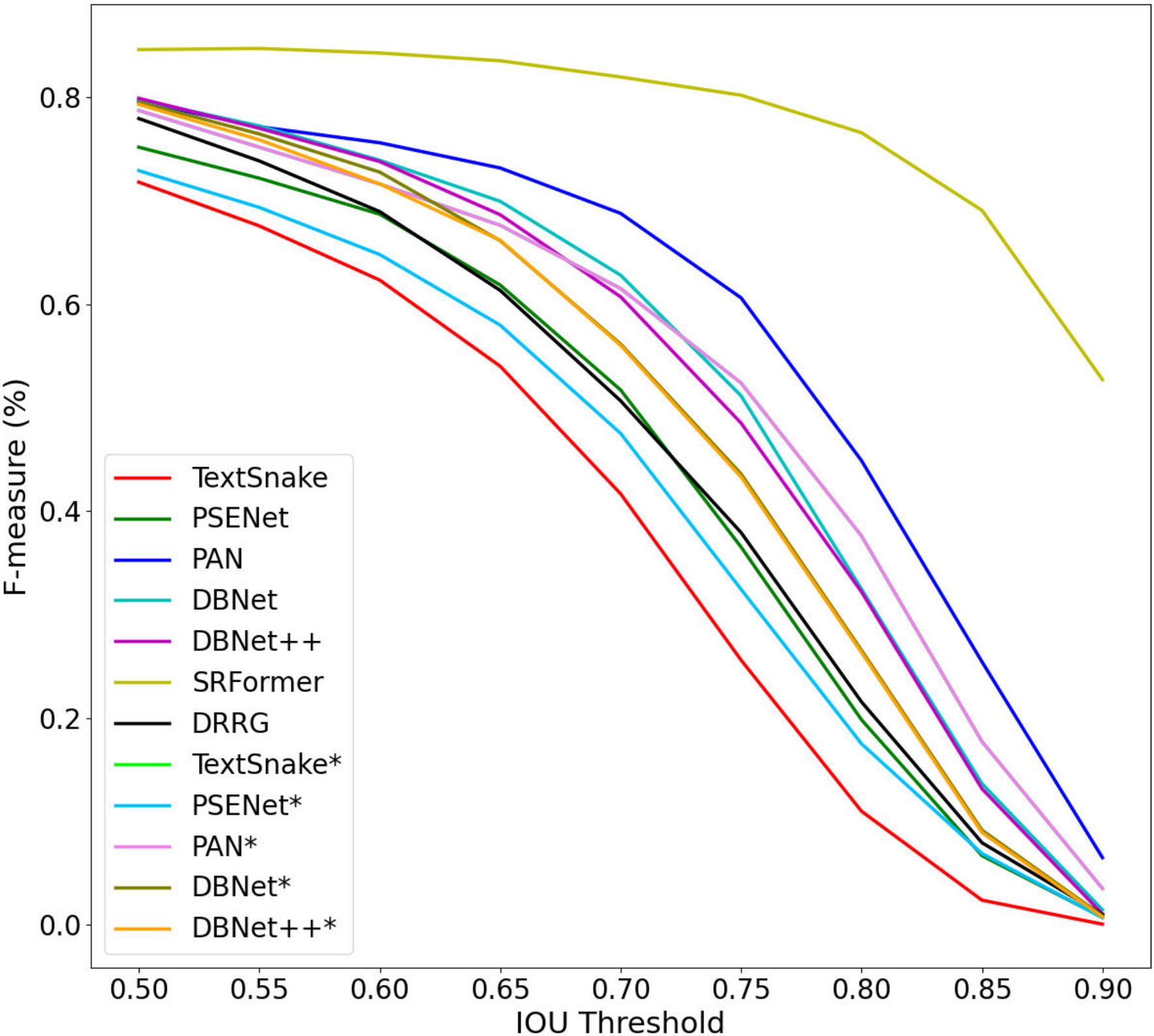}} 
\leftline{\small{\hspace{1.6cm} (a) CTW \hspace{3cm} (b) TOT}}
\caption{Performance changes against the IOU threshold in the evaluation protocol. 
Best viewed in color.
}
\label{fig6}
\vspace{-0.2cm}
\end{figure}

Table \ref{table3} shows that SRFormer \cite{bu2023srformer} and PAN* \cite{ICCV19/PAN/Wang} achieve the optimal $\textit{F-measure}$ on CTW under the evaluation protocols IOU@0.5 and DetEval-v2, respectively.\; As shown in Table \ref{table4}, on the dataset TOT, SRFormer \cite{bu2023srformer} and PAN* \cite{ICCV19/PAN/Wang} have obtained the best $\textit{F-measure}$ under IOU@0.5 and DetEval-v2, respectively. In terms of speed, DBNet++ \cite{DBNet++} significantly outperforms other methods. Meanwhile, the speed of the component-wise based methods TextSnake \cite{ECCV18/Textsnake/Long} and DRRGN \cite{CVPR20/DRRGN/Zhang} is distinctly slower than the pixel-wise based methods (\textit{e.g.}, PSENet \cite{CVPR19/PSENet/Wang}, PAN \cite{ICCV19/PAN/Wang}, DBNet \cite{AAAI20/DB/Liao}, etc.). Some visualized detection results are presented in Figure \ref{fig:vis_det_results}. We can observe that some state-of-the-art methods can not localize the accurate text contours. Some text contours even contain multiple scene text instances under complicated scenarios. 
These quantitative and qualitative experimental results further indicate that an older method can be better than a newer method in performance or speed under fair comparisons. It is because the improvements in the original methods mainly come from some tricks (\textit{e.g.}, more training data, stronger backbone, well-designed feature fusion strategies, etc.), and the proposed text-shape representation may not work well. To further verify the ability to accurately localize the text contours, we use stricter thresholds under the evaluation protocol IOU. As shown in Figure \ref{fig6}, PAN \cite{ICCV19/PAN/Wang} and SRFormer \cite{bu2023srformer} are more robust to the change of IOU threshold than other methods on CTW and TOT, respectively. Similarly, the qualitative results in Figure \ref{fig:vis_det_results} have also shown that  PAN \cite{ICCV19/PAN/Wang} and SRFormer \cite{bu2023srformer} can achieve more accurate text contours.  

\begin{table}[t]
\caption{Comparison of generalization capabilities from TOT to CTW datasets.}
\label{table6}
\centering
\resizebox{\columnwidth}{!}{%
\begin{tabular}{l|llllll}
\toprule
 & \multicolumn{3}{c}{IOU@0.5} & \multicolumn{3}{c}{DetEval-v2} \\ \cline{2-7} 
\multirow{-2}{*}{Method} & \multicolumn{1}{c}{R (\%)} & \multicolumn{1}{c}{P (\%)} & \multicolumn{1}{c}{F (\%)} & \multicolumn{1}{c}{R (\%)} & \multicolumn{1}{c}{P (\%)} & \multicolumn{1}{c}{F (\%)} \\ \toprule
{TextSnake} & 53.32 & 37.77 & 44.22 & 70.8 & 67.23 & 68.97 \\
{PSENet} & 51.92 & 36.65 & 42.97 & 61.07 & 59.78 & 60.42 \\
{PAN} & 36.28 & 43.05 & 39.54 & \textbf{71.92} & 64.57 & 68.05 \\
{DBNet} & 53.65 & 36.09 & 43.15 & 64.22 & 70.09 & 67.03 \\
{DBNet++} & \textbf{54.30} & 34.39 & 42.11 & 67.10 & 69.28 & 68.17 \\
DRRGN & 48.34 & 35.59 & 41.00 & 53.17 & 56.50 & 54.79   \\
SRFormer & 32.70 & \textbf{60.59} & 42.48 & 61.77 & 68.50 & 64.96 \\ \hline
TextSnake* & 53.75 & 47.75 & \textbf{47.62} & 69.60 & 70.79 & 70.19 \\
PSENet* & 50.68 & 35.70 & 41.89 & 60.27 & 61.98 & 61.11 \\
PAN* & 37.62 & 51.86 & 43.61 & 69.44 & 63.40 & 66.28 \\
DBNet* & 56.16 & 35.92 & 43.81 & 68.74 & 72.02 & 70.34 \\
DBNet++* & 55.93 & 35.05 & 43.09 & 69.31 & \textbf{72.19} & \textbf{70.72} \\
\toprule
\end{tabular}%
}
\vspace{-0.1cm}

\end{table}

\subsection{Influence of Testing Scale}
\begin{figure}[t]
   \begin{center}
      \includegraphics[width=0.45\linewidth]{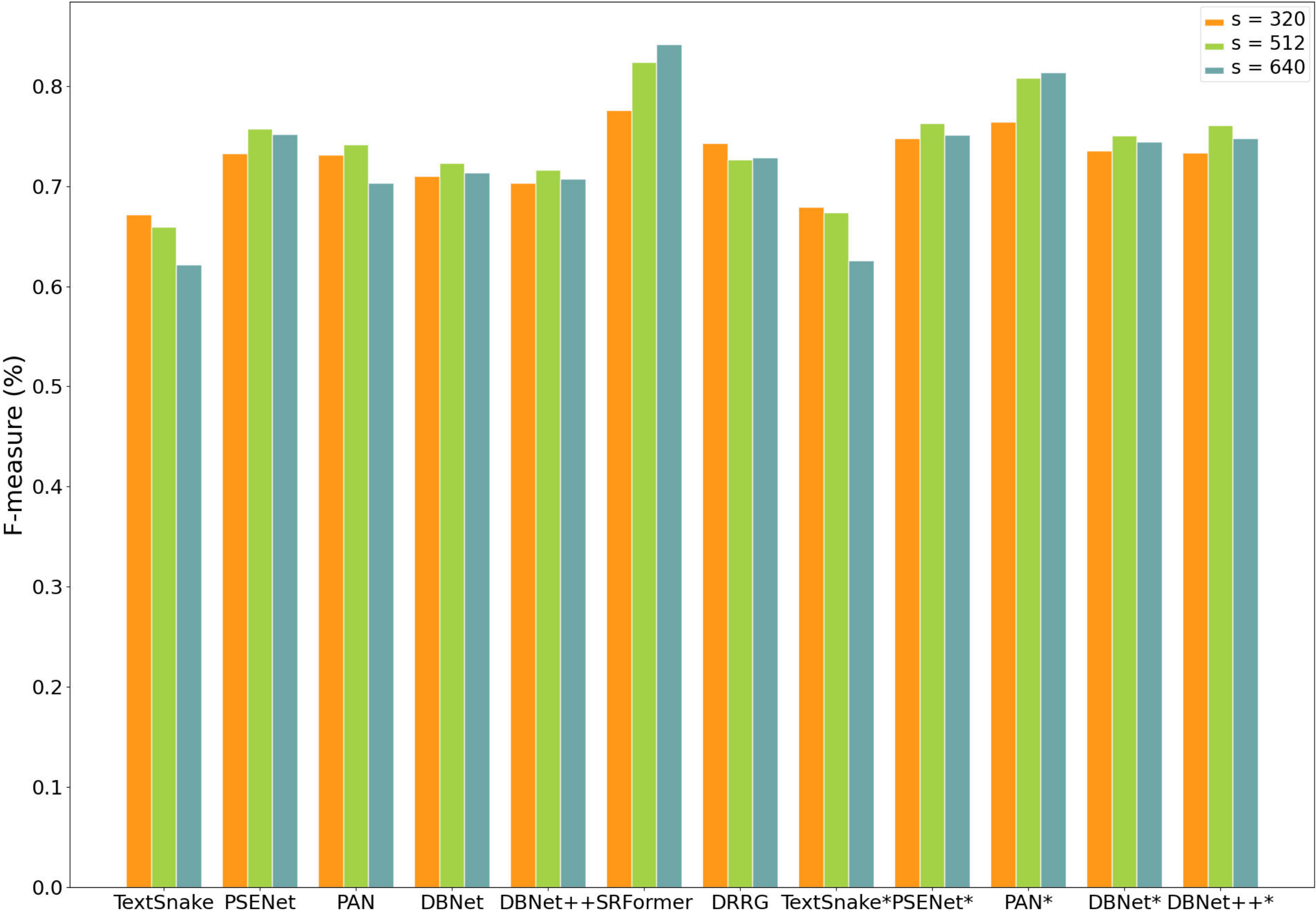}
      \includegraphics[width=0.45\linewidth]{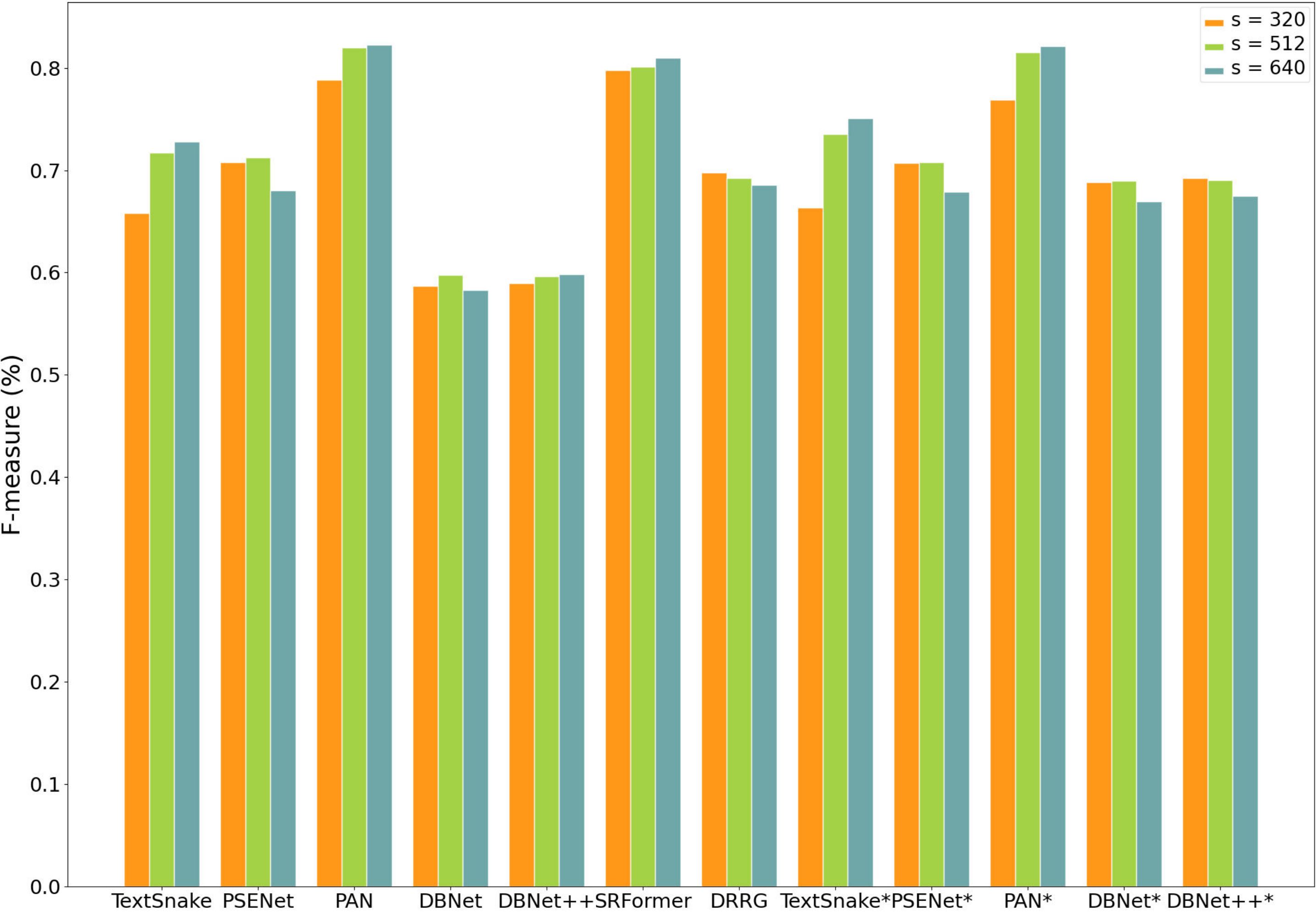}
      \leftline{\small{\hspace{0.78cm} (a) CTW (\textit{IOU@0.5}) \hspace{1.35cm} (b) CTW (\textit{DetEval-v2}) }}
      
      \includegraphics[width=0.45\linewidth]{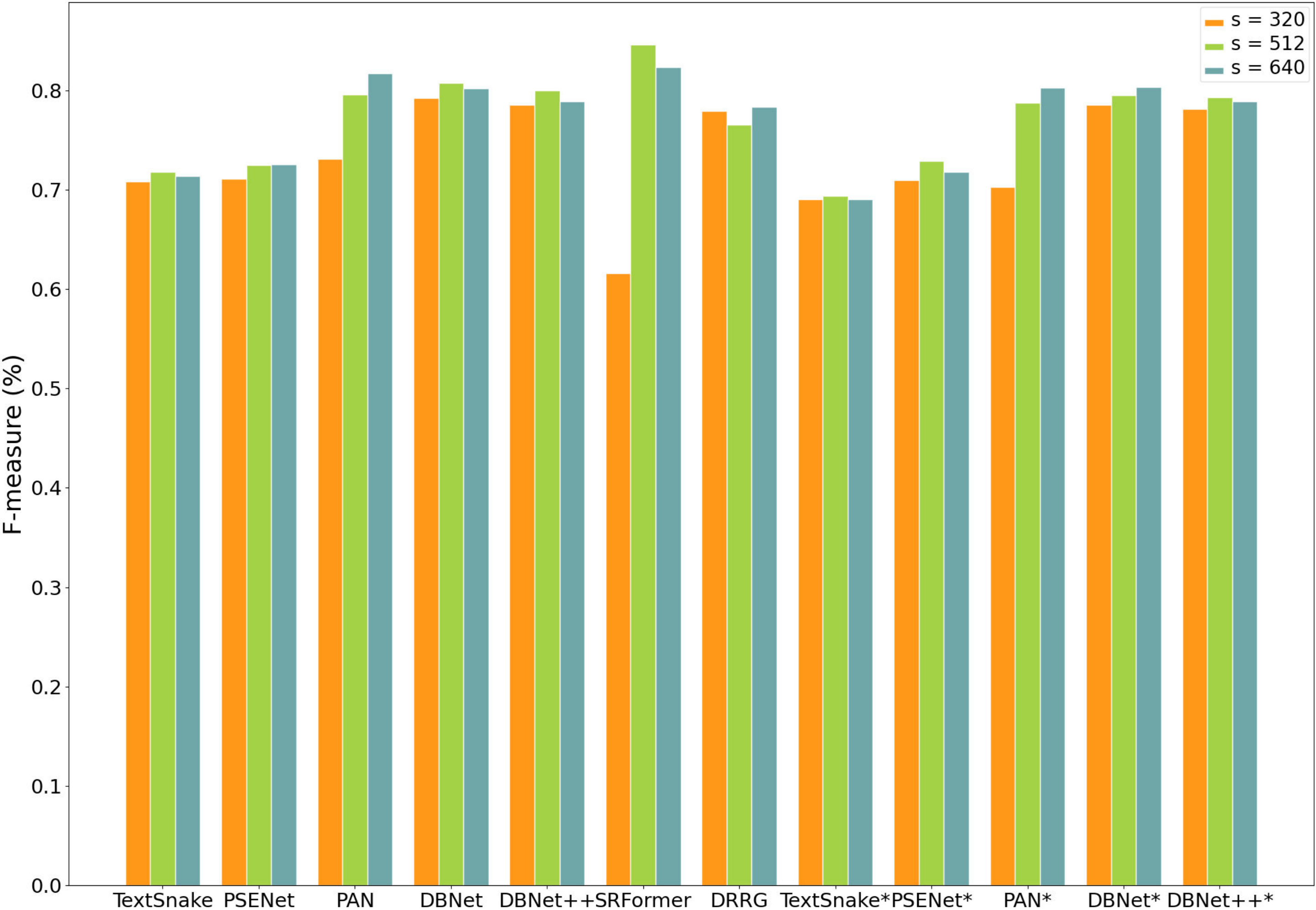}
      \includegraphics[width=0.45\linewidth]{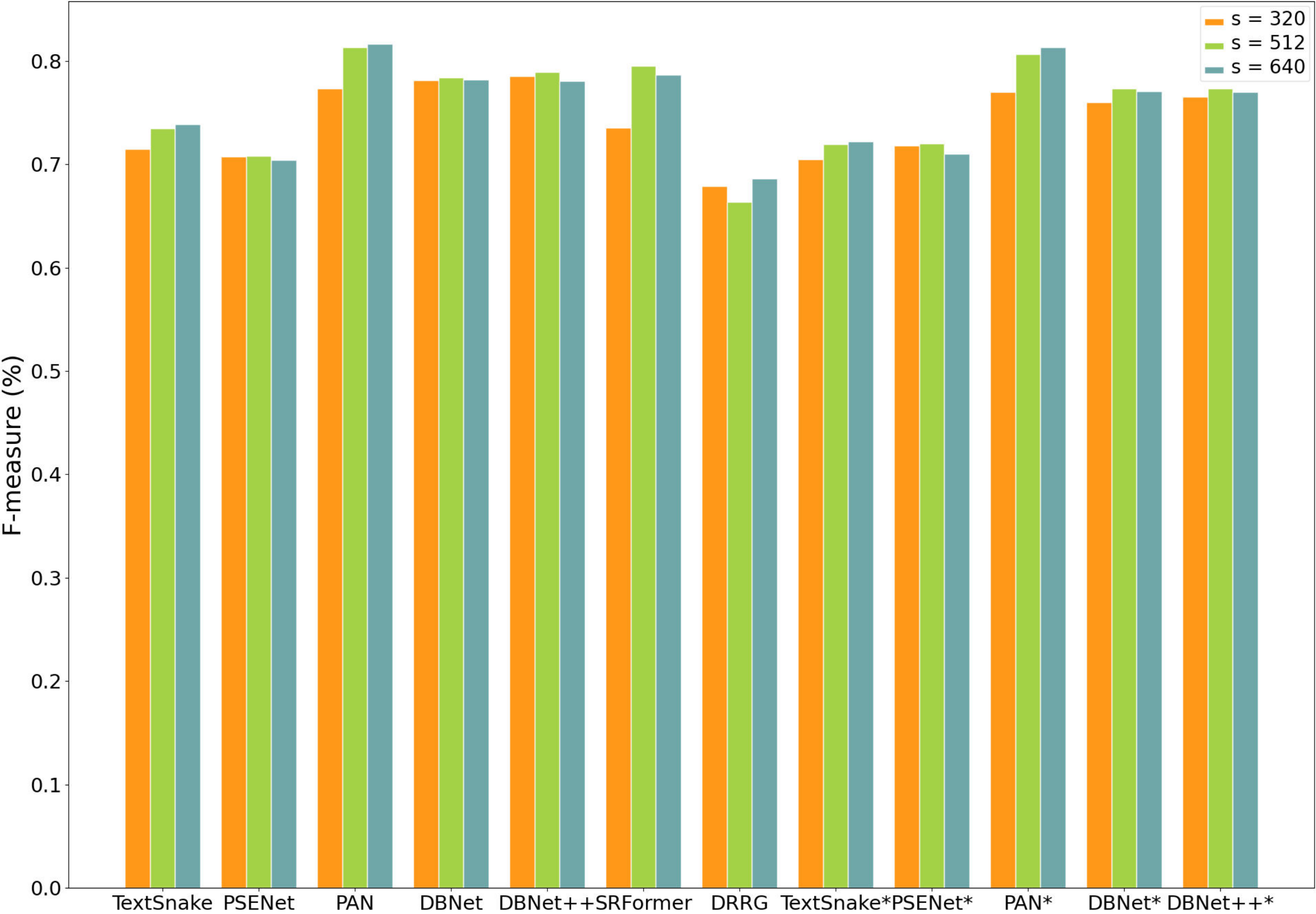}
      \leftline{\small{\hspace{0.78cm} (c) TOT (\textit{IOU@0.5}) \hspace{1.35cm} (d) TOT (\textit{DetEval-v2}) }}
   \end{center}
   \vspace{-0.1cm}
   \caption{Effect of testing image scales. }
   \label{fig:effect_scale}
   \vspace{-0.25cm}
\end{figure}

To investigate the robustness of scales, we validate several kinds of testing image scales like those in \cite{ICCV19/PAN/Wang}. As shown in Figure \ref{fig:effect_scale}, it indicates that different testing image scales would result in obviously different performances under different evaluation protocols for various datasets. It is hard to ensure the robustness against the testing scale for existing methods. Even though the short size $s$ of the test image is consistent with the size ($640\times640$) of the training image, it also can not ensure achieving the optimal performance. Under many conditions, it achieves the best performance, when using a smaller size (e.g., 512). The reason may be ascribed to the domain shift of the scale between the training data and the testing data. In effect, existing methods usually utilize the testing scale with no evidence. Sometimes, when we only change the testing scale, it also brings an obvious performance improvement.  In the past few years, learning a scale-robust detector for arbitrary-shape scene texts with diversified scales and aspect ratios has drawn little attention in the field of scene text detection.  

\subsection{Exploration of Generalization Ability}
To verify the generalization ability of existing methods, we conduct cross-dataset experiments. We train the models on the training set of CTW and then test on the testing set of TOT (CTW $\rightarrow$ TOT), and vice versa (TOT $\rightarrow$ CTW). As shown in Table \ref{table5} and Table \ref{table6}, we find that SRFormer \cite{bu2023srformer} and PAN \cite{ICCV19/PAN/Wang} can achieve the best $\textit{F-measure}$ on CTW $\rightarrow$ TOT, under IOU@0.5 and DetEval-v2. Meanwhile, under the evaluation protocols of IOU@0.5, the older detector TextSnake* \cite{ECCV18/Textsnake/Long} shows better generalization abilities on  TOT$\rightarrow$CTW than other methods. These experiments further reveal that, under the premise of a fair comparison, some older methods may still outperform newly proposed methods in certain aspects due to their superior properties. In addition, existing methods all suffer a significant decrease in performance in cross-dataset testing, which indicates that current scene text detector methods still fail to achieve decent generalization.

\subsection{Analyses of Convergence}
\begin{figure}[t]
   \begin{center} 
   \includegraphics[width=0.4\linewidth,height=0.36\linewidth]{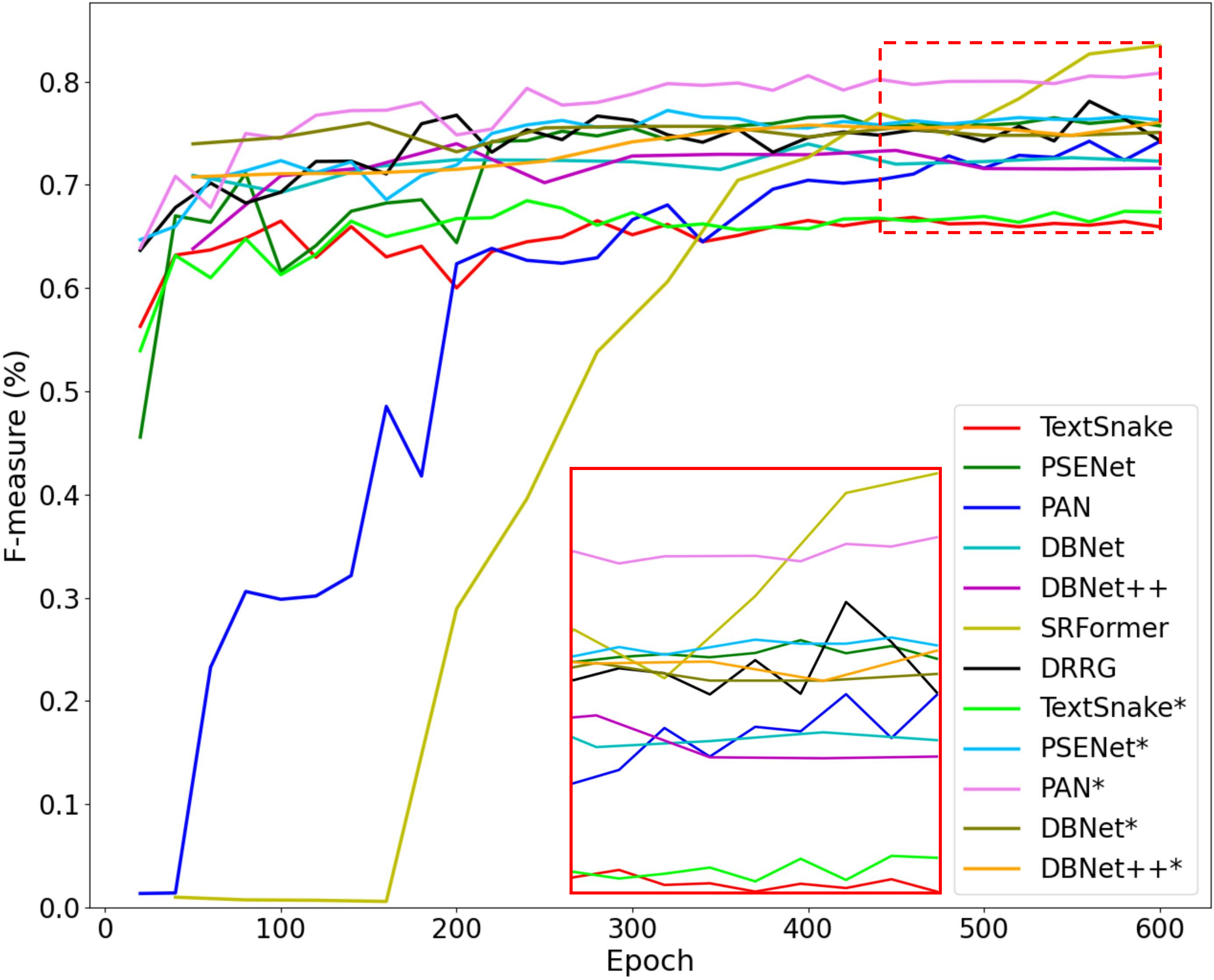}
   \hspace{0.2cm}
   \includegraphics[width=0.4\linewidth,height=0.36\linewidth]{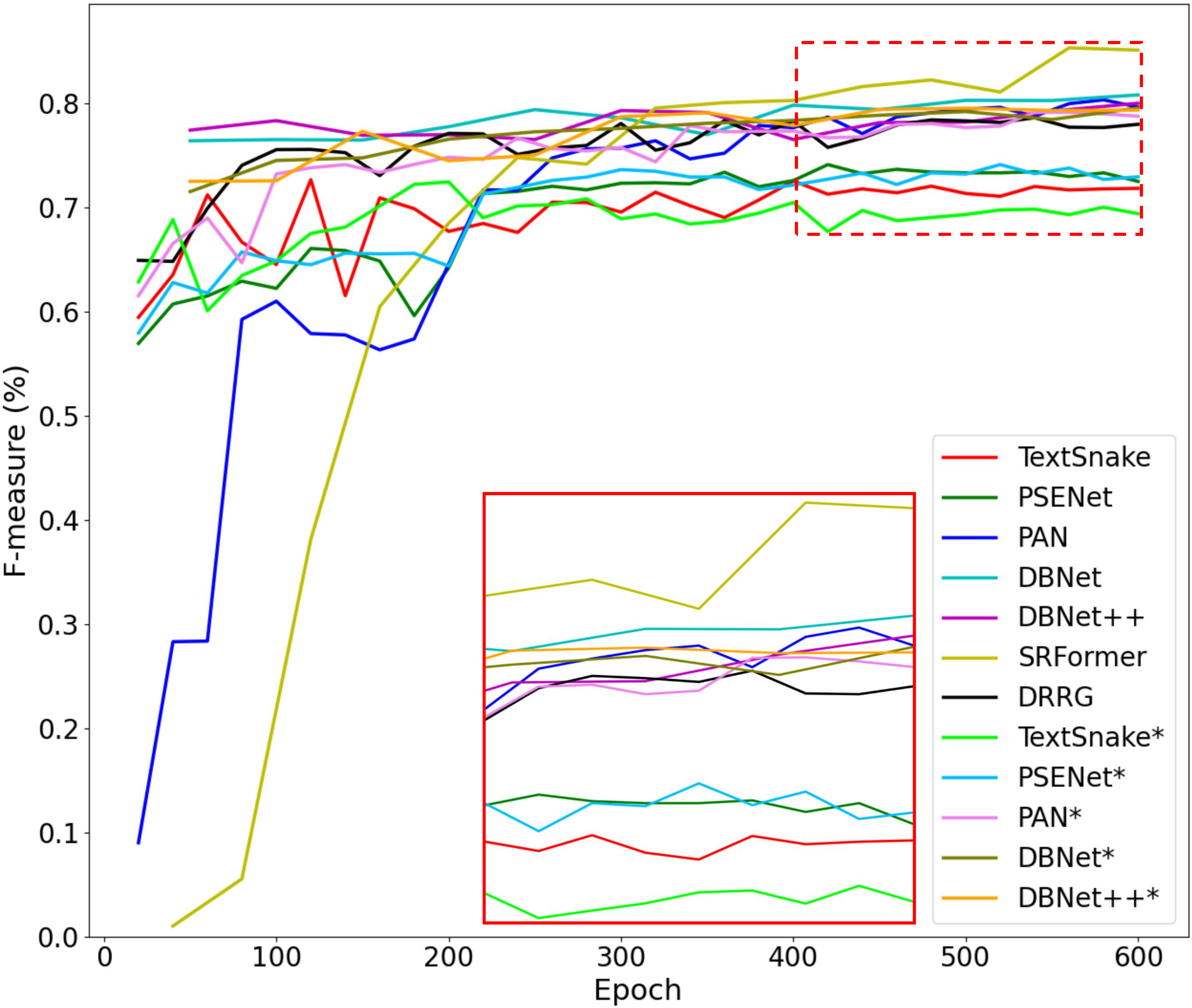}
     \leftline{\small{\hspace{1.60cm} (a) CTW \hspace{3.1cm} (b) TOT}}
   \end{center}
   \caption{Illustration of the model convergence under the evaluation protocol IOU@0.5. The content in the solid line box is the zoom of that in the dashed box.
   }
   \label{fig:effect_convergence}
   \vspace{-0.25cm}
\end{figure}

Figure \ref{fig:effect_convergence} shows that TextSnake \cite{ECCV18/Textsnake/Long} and PAN \cite{ICCV19/PAN/Wang} achieve more stable convergence, when evaluated under both IOU@0.5 and DetEval-v2. Meanwhile, TextSnake  \cite{ECCV18/Textsnake/Long} and TextSnake* \cite{ECCV18/Textsnake/Long} also show faster convergence than other methods. Surprisingly, although the convergence of SRFormer \cite{bu2023srformer} is relatively poor, the final score reaches the highest, which shows that the detection model represented by the visual transformer structure requires a lot of computing power to show excellent performance. Figure \ref{fig:effect_convergence} also indicates that the convergence of the model is dependent on the test datasets. For example, DRRGN \cite{CVPR20/DRRGN/Zhang} can achieve better convergence on both CTW and TOT after 200 epochs.
Generally speaking, the stability and speed of the model convergence also reflect the advantages of the model to some extent, which is usually ignored by existing scene text detectors.

\section{Future Directions} \label{sec:directions}
In the field of arbitrarily-shaped scene text detection, future research directions may involve the following aspects: 

(i) Insufficient research on fine-grained localization of scene text in arbitrary layouts. Current detection methods for arbitrary-shaped scene text are primarily designed to serve downstream tasks such as recognition and deep semantic understanding. When the detected bounding box deviates slightly from the ground truth, or even when some characters of a text instance fall outside the box, the current evaluation metrics may still consider the detection correct, and downstream tasks can often rely on contextual semantic information to recognize or interpret the text correctly. However, when downstream tasks require fine-grained operations such as scene text editing or erasing, imprecise localization can have a significant impact. Although some existing methods achieve fine-grained segmentation of scene text, their capability remains notably insufficient for arbitrary-shaped text instances. 

(ii) Inadequate exploration of cross-domain generalization for arbitrary-shaped scene text detection. Current detection methods mainly focus on in-domain printed text detection, and few studies have systematically investigated the generalization capabilities on out-of-domain data such as handwriting, artistic fonts, ancient scripts, and screen-captured text. A straightforward but crude approach is to mix such data for training, which neglects the inherent semantic associations of text and prior knowledge of character structures. Moreover, while current multimodal large models exhibit good generalization performance, their localization ability for diverse types of arbitrary-shaped scene text remains highly deficient. 

(iii) Insufficient research on weakly supervised learning capabilities. Existing detection methods typically pre-train on fully annotated synthetic datasets and then fine-tune on specific datasets with full annotations. Different synthetic data significantly affect detection performance and lead to long training times. Furthermore, fully annotating arbitrary-shaped scene text is labor-intensive. Although some studies explore weakly supervised scene text detection, their limited performance has prevented them from becoming mainstream. In fact, the inherent properties of scene text—i.e., fragments of a text instance can themselves be considered text, along with their semantic associations with annotations—offer great potential for leveraging self-supervised learning to reduce the demand for large-scale labeled data. However, this direction remains underexplored.

\section{Conclusions} \label{sec:conclusions}
In this paper, we reveal the inconsistencies between existing arbitrary-shape scene text detectors. These inconsistencies bring difficulties in determining the advantages of the newly proposed core techniques compared with previous methods. To clean up the hindrance of fair comparisons, we present a unified framework for bottom-up arbitrary-shape scene text detection models. With this framework, we have provided a fair comparison of several well-known methods. We also provide comprehensive analyses of these methods to better disclose their strengths and weaknesses. Thus, these profound studies suggest multiple directions to explore in the future. Firstly, it is necessary to propose more robust representations for describing arbitrarily-shaped scene texts with various scales and aspect ratios for more accurate localizations. Secondly, it is very meaningful to explore more efficient online data augmentation strategies by considering the intrinsic characteristics of scene texts. Thirdly, it is useful to study the scene text detector with a trade-off between performance and speed in resource-constrained circumstances.

\small
\bibliographystyle{IEEEtran}
\bibliography{ref-page}

@String(CVPR  = {CVPR})

@String(ICCV  = {ICCV})

@String(ECCV  = {ECCV})

@String(IJCAI = {IJCAI})

@String(AAAI = {AAAI})

@String(CVPR= {IEEE Conf. Comput. Vis. Pattern Recog.})

@String(ICCV= {Int. Conf. Comput. Vis.})

@String(ECCV= {Eur. Conf. Comput. Vis.})

@article{TPAMI17/CRNN/Shi,
  author       = {Baoguang Shi and
                  Xiang Bai and
                  Cong Yao},
  title        = {An End-to-End Trainable Neural Network for Image-Based Sequence Recognition and Its Application to Scene Text Recognition},
  journal      = {{IEEE} Trans. Pattern Anal. Mach. Intell.},
  volume       = {39},
  number       = {11},
  pages        = {2298--2304},
  year         = {2017}
}

@article{TPAMI19/ASTER/Shi,
  author       = {Baoguang Shi and
                  Mingkun Yang and
                  Xinggang Wang and
                  Pengyuan Lyu and
                  Cong Yao and
                  Xiang Bai},
  title        = {{ASTER:} An Attentional Scene Text Recognizer with Flexible Rectification},
  journal      = {{IEEE} Trans. Pattern Anal. Mach. Intell.},
  volume       = {41},
  number       = {9},
  pages        = {2035--2048},
  year         = {2019}
}

@inproceedings{ECCV20/DETR/Carion,
  author       = {Nicolas Carion and
                  Francisco Massa and
                  Gabriel Synnaeve and
                  Nicolas Usunier and
                  Alexander Kirillov and
                  Sergey Zagoruyko},
  title        = {End-to-End Object Detection with Transformers},
  booktitle    = {ECCV},
  volume       = {12346},
  pages        = {213--229},
  year         = {2020}
}

@inproceedings{CVPR21/FCENet/Zhu,
  author       = {Yiqin Zhu and
                  Jianyong Chen and
                  Lingyu Liang and
                  Zhanghui Kuang and
                  Lianwen Jin and
                  Wayne Zhang},
  title        = {Fourier Contour Embedding for Arbitrary-Shaped Text Detection},
  booktitle    = {CVPR},
  pages        = {3123--3131},
  year         = {2021}
}

@InProceedings{CVPR22/KMST/Wang,
    author    = {Wang, Hao and Liao, Junchao and Cheng, Tianheng and Gao, Zewen and Liu, Hao and Ren, Bo and Bai, Xiang and Liu, Wenyu},
    title     = {Knowledge Mining With Scene Text for Fine-Grained Recognition},
    booktitle = {CVPR},
    year      = {2022},
    pages     = {4624-4633}
}

@article{TMM22/SADA_SSC/Dai,
  author       = {Pengwen Dai and
                  Yang Li and
                  Hua Zhang and
                  Jingzhi Li and
                  Xiaochun Cao},
  title        = {Accurate Scene Text Detection Via Scale-Aware Data Augmentation and Shape Similarity Constraint},
  journal      = {{IEEE} Trans. Multimedia},
  volume       = {24},
  pages        = {1883--1895},
  year         = {2022}
}

@article{IJCV24/CBNet/Zhao,
  author       = {Xi Zhao and
                  Wei Feng and
                  Zheng Zhang and
                  Jingjing Lv and
                  Xin Zhu and
                  Zhangang Lin and
                  Jinghe Hu and
                  Jingping Shao},
  title        = {CBNet: {A} Plug-and-Play Network for Segmentation-Based Scene Text
                  Detection},
  journal      = {Int. J. Comput. Vis.},
  volume       = {132},
  number       = {8},
  pages        = {3119--3138},
  year         = {2024}
}

@InProceedings{CVPR22/ViSTA/Cheng,
    author    = {Cheng, Mengjun and Sun, Yipeng and Wang, Longchao and Zhu, Xiongwei and Yao, Kun and Chen, Jie and Song, Guoli and Han, Junyu and Liu, Jingtuo and Ding, Errui and Wang, Jingdong},
    title     = {{ViSTA}: Vision and Scene Text Aggregation for Cross-Modal Retrieval},
    booktitle = {CVPR},
    year      = {2022},
    pages     = {5184-5193}
}

@inproceedings{AAAI23/DPText-DETR/Ye,
  author       = {Maoyuan Ye and
                  Jing Zhang and
                  Shanshan Zhao and
                  Juhua Liu and
                  Bo Du and
                  Dacheng Tao},
  title        = {{DPText-DETR}: Towards Better Scene Text Detection with Dynamic Points in Transformer},
  booktitle    = {AAAI},
  pages        = {3241--3249},
  year         = {2023}
}

@article{TPAMI23/ABINet++/Fang,
  author       = {Shancheng Fang and
                  Zhendong Mao and
                  Hongtao Xie and
                  Yuxin Wang and
                  Chenggang Yan and
                  Yongdong Zhang},
  title        = {{ABINet++}: Autonomous, Bidirectional and Iterative Language Modeling
                  for Scene Text Spotting},
  journal      = {{IEEE} Trans. Pattern Anal. Mach. Intell.},
  volume       = {45},
  number       = {6},
  pages        = {7123--7141},
  year         = {2023}
}

@InProceedings{CVPR23/Deepsolo/Ye,
  author={Ye, Maoyuan and Zhang, Jing and Zhao, Shanshan and Liu, Juhua and Liu, Tongliang and Du, Bo and Tao, Dacheng},
  booktitle={CVPR}, 
  title={DeepSolo: Let Transformer Decoder with Explicit Points Solo for Text Spotting}, 
  year={2023},
  pages={19348-19357}
}

@article{IJCV22/I3CL/Du,
  title={I3CL: Intra-and Inter-Instance Collaborative Learning for Arbitrary-shaped Scene Text Detection},
  author={Du, Bo and Ye, Jian and Zhang, Jing and Liu, Juhua and Tao, Dacheng},
  journal={Int. J. Comput. Vis.},
  volume={130},
  number={8},
  pages={1961--1977},
  year={2022}
}

@article{TNNLS23/KPN/Zhang,
  author       = {Shi{-}Xue Zhang and
                  Xiaobin Zhu and
                  Jie{-}Bo Hou and
                  Chun Yang and
                  Xu{-}Cheng Yin},
  title        = {Kernel Proposal Network for Arbitrary Shape Text Detection},
  journal      = {{IEEE} Trans. Neural Networks Learn. Syst.},
  volume       = {34},
  number       = {11},
  pages        = {8731--8742},
  year         = {2023}
}

@inproceedings{MM23/SIR/Qin,
  author       = {Xugong Qin and
                  Pengyuan Lyu and
                  Chengquan Zhang and
                  Yu Zhou and
                  Kun Yao and
                  Peng Zhang and
                  Hailun Lin and
                  Weiping Wang},
  title        = {Towards Robust Real-Time Scene Text Detection: From Semantic to Instance Representation Learning},
  booktitle    = {ACM-MM},
  pages        = {2025--2034},
  year         = {2023}
}

@article{TNNLS24/GLaLT/Zhang,
  author       = {Hui Zhang and
                  Guiyang Luo and
                  Jian Kang and
                  Shan Huang and
                  Xiao Wang and
                  Fei{-}Yue Wang},
  title        = {GLaLT: Global-Local Attention-Augmented Light Transformer for   
                  Scene Text Recognition},
  journal      = {{IEEE} Trans. Neural Networks Learn. Syst.},
  volume       = {35},
  number       = {7},
  pages        = {10145--10158},
  year         = {2024}
}

@inproceedings{AAAI24/LRANet/Su,
  author       = {Yuchen Su and
                  Zhineng Chen and
                  Zhiwen Shao and
                  Yuning Du and
                  Zhilong Ji and
                  Jinfeng Bai and
                  Yong Zhou and
                  Yu{-}Gang Jiang},
  title        = {{LRANet}: Towards Accurate and Efficient Scene Text Detection with Low-Rank Approximation Network},
  booktitle    = {AAAI},
  pages        = {4979--4987},
  year         = {2024}
}

@inproceedings{CVPR24/KACM/Zheng,
  author       = {Jinzhi Zheng and
                  Heng Fan and
                  Libo Zhang},
  title        = {Kernel Adaptive Convolution for Scene Text Detection via Distance Map Prediction},
  booktitle    = {CVPR},
  pages        = {5957--5966},
  year         = {2024}
}

@inproceedings{ICML24/DAT/Wan,
  author       = {Xingyu Wan and
                  Chengquan Zhang and
                  Pengyuan Lyu and
                  Sen Fan and
                  Zihan Ni and
                  Kun Yao and
                  Errui Ding and
                  Jingdong Wang},
  title        = {Towards Unified Multi-granularity Text Detection with Interactive Attention},
  pages = {50012 - 50025},
  booktitle    = {ICML},
  year         = {2024}
}

@article{TIP24/IAST/Zhang,
  author       = {Shi{-}Xue Zhang and
                  Chun Yang and
                  Xiaobin Zhu and
                  Hongyang Zhou and
                  Hongfa Wang and
                  Xu{-}Cheng Yin},
  title        = {Inverse-Like Antagonistic Scene Text Spotting via Reading-Order Estimation and Dynamic Sampling},
  journal      = {{IEEE} Trans. Image Process.},
  volume       = {33},
  pages        = {825--839},
  year         = {2024}
}

@inproceedings{AAAI24/Box2Poly/Chen,
  author       = {Xuyang Chen and
                  Dong Wang and
                  Konrad Schindler and
                  Mingwei Sun and
                  Yongliang Wang and
                  Nicol{\'{o}} Savioli and
                  Liqiu Meng},
  title        = {{Box2Poly}: Memory-Efficient Polygon Prediction of Arbitrarily Shaped and Rotated Text},
  booktitle    = {AAAI},
  pages        = {1219--1227},
  year         = {2024}
}

@inproceedings{CVPR24/ODM/Duan,
  author       = {Chen Duan and
                  Pei Fu and
                  Shan Guo and
                  Qianyi Jiang and
                  Xiaoming Wei},
  title        = {{ODM:} {A} Text-Image Further Alignment Pre-training Approach for Scene Text Detection and Spotting},
  booktitle    = {CVPR},
  pages        = {15587--15597},
  year         = {2024}
}

@inproceedings{CVPR24/LayoutFormer/Liang,
  author       = {Min Liang and
                  Jia{-}Wei Ma and
                  Xiaobin Zhu and
                  Jingyan Qin and
                  Xu{-}Cheng Yin},
  title        = {Layout{F}ormer: Hierarchical Text Detection Towards Scene Text Understanding},
  booktitle    = {CVPR},
  pages        = {15665--15674},
  year         = {2024}
}

@inproceedings{AAAI25/ERRNet/Sun,
  author       = {Yuchen Su and
                  Zhineng Chen and
                  Yongkun Du and
                  Zhilong Ji and
                  Kai Hu and
                  Jinfeng Bai and
                  Xieping Gao},
  title        = {Explicit Relational Reasoning Network for Scene Text Detection},
  booktitle    = {AAAI},
  pages        = {7069--7077},
  year         = {2025}
}

@article{IJCV25/SwinTextSpotterv2/Huang,
  author       = {Mingxin Huang and
                  Dezhi Peng and
                  Hongliang Li and
                  Zhenghao Peng and
                  Chongyu Liu and
                  Dahua Lin and
                  Yuliang Liu and
                  Xiang Bai and
                  Lianwen Jin},
  title        = {{SwinTextSpotter v2}: Towards Better Synergy for Scene Text Spotting},
  journal   = {Int. J. Comput. Vis.},
   volume       = {133},
  number       = {8},
  pages        = {5281--5301},
  year         = {2025}
}

@inproceedings{ICCV19/textRec_fairComp/Baek,
  author       = {Jeonghun Baek and
                  Geewook Kim and
                  Junyeop Lee and
                  Sungrae Park and
                  Dongyoon Han and
                  Sangdoo Yun and
                  Seong Joon Oh and
                  Hwalsuk Lee},
  title        = {What Is Wrong With Scene Text Recognition Model Comparisons? Dataset and Model Analysis},
  booktitle    = {ICCV},
  pages        = {4714--4722},
  year         = {2019}
}

@inproceedings{CVPR20/EST_VQA/Wang,
  author    = {Xinyu Wang and
               Yuliang Liu and
               Chunhua Shen and
               Chun Chet Ng and
               Canjie Luo and
               Lianwen Jin and
               Chee Seng Chan and
               Anton van den Hengel and
               Liangwei Wang},
  title     = {On the General Value of Evidence, and Bilingual Scene-Text Visual Question Answering},
  booktitle   = {CVPR},
  pages     = {10123-10132},
  year      = {2020}
}

@inproceedings{ronen2022glass,
  title={Glass: Global to local attention for scene-text spotting},
  author={Ronen, Roi and Tsiper, Shahar and Anschel, Oron and Lavi, Inbal and Markovitz, Amir and Manmatha, R},
  booktitle={ECCV},
  pages={249--266},
  year={2022}
}

@inproceedings{MM20/TextRay/Wang,
  author    = {Fangfang Wang and
               Yifeng Chen and
               Fei Wu and
               Xi Li},
  title     = {Text{R}ay: Contour-based Geometric Modeling for Arbitrary-shaped Scene Text Detection},
  booktitle = {ACM-MM},
  pages     = {111--119},
  year      = {2020}
}

@inproceedings{CVPR16/FCN_Text/Zhang,
  author    = {Zheng Zhang and
               Chengquan Zhang and
               Wei Shen and
               Cong Yao and
               Wenyu Liu and
               Xiang Bai},
  title     = {Multi-oriented Text Detection with Fully Convolutional Networks},
  booktitle = {CVPR},
  pages     = {4159--4167},
  year      = {2016}
}

@inproceedings{AAAI18/PixelLink/Deng,
  author    = {Dan Deng and
               Haifeng Liu and
               Xuelong Li and
               Deng Cai},
  title     = {Pixel{L}ink: Detecting Scene Text via Instance Segmentation},
  booktitle = {AAAI},
  pages     = {6773--6780},
  year      = {2018},

}

@inproceedings{ECCV18/Border/Xue,
  author    = {Chuhui Xue and
               Shijian Lu and
               Fangneng Zhan},
  title     = {Accurate Scene Text Detection Through Border Semantics Awareness and Bootstrapping},
  booktitle = {ECCV},
  pages     = {370--387},
  year      = {2018}
}

@inproceedings{CVPR19/PSENet/Wang,
  author    = {Wenhai Wang and
               Enze Xie and
               Xiang Li and
               Wenbo Hou and
               Tong Lu and
               Gang Yu and
               Shuai Shao},
  title     = {Shape Robust Text Detection with Progressive Scale Expansion Network},
  booktitle = {CVPR},
  pages     = {9336--9345},
  year      = {2019}
 
}

@inproceedings{CVPR19/SAE/Tian,
  title={Learning Shape-Aware Embedding for Scene Text Detection},
  author={Zhuotao Tian and
        Michelle Shu and
        Pengyuan Lyu and 
        Ruiyu Li and 
        Chao Zhou and 
        Xiaoyong Shen and
        Jiaya Jia},
  booktitle={CVPR},
  pages     = {4234--4243},
  year={2019}
}

@article{TIP19/TextField/Xu,
  author    = {Yongchao Xu and
               Yukang Wang and
               Wei Zhou and
               Yongpan Wang and
               Zhibo Yang and
               Xiang Bai},
  title     = {Text{F}ield: Learning {A} Deep Direction Field for Irregular Scene Text Detection},
  journal   = {{IEEE} Trans. Image Process.},
  volume    = {28},
  number    = {11},
  pages     = {5566--5579},
  year      = {2019},
}

@inproceedings{MM19/SAST/Wang,
  author    = {Pengfei Wang and
               Chengquan Zhang and
               Fei Qi and
               Zuming Huang and
               Mengyi En and
               Junyu Han and
               Jingtuo Liu and
               Errui Ding and
               Guangming Shi},
  title     = {A Single-Shot Arbitrarily-Shaped Text Detector based on Context Attended
               Multi-Task Learning},
  booktitle = {ACM-MM},
  pages     = {1277--1285},
  year      = {2019}
}

@inproceedings{ICCV19/PAN/Wang,
  author    = {Wenhai Wang and
               Enze Xie and
               Xiaoge Song and
               Yuhang Zang and
               Wenjia Wang and
               Tong Lu and
               Gang Yu and
               Chunhua Shen},
  title     = {Efficient and Accurate Arbitrary-Shaped Text Detection with Pixel Aggregation Network},
  booktitle = {ICCV},
  pages     = {8439--8448},
  year      = {2019}
}

@inproceedings{ECCV18/Mask_TextSpotter/Lyu,
  author    = {Pengyuan Lyu and
               Minghui Liao and
               Cong Yao and
               Wenhao Wu and
               Xiang Bai},
  title     = {Mask {T}ext{S}potter: An End-to-End Trainable Neural Network for Spotting
               Text with Arbitrary Shapes},
  booktitle = {ECCV},
  pages     = {71--88},
  year      = {2018}
}

@inproceedings{AAAI19/SPCN/Xie,
  author    = {Enze Xie and
               Yuhang Zang and
               Shuai Shao and
               Gang Yu and
               Cong Yao and
               Guangyao Li},
  title     = {Scene Text Detection with Supervised Pyramid Context Network},
  booktitle = {AAAI},
  pages     = {9038--9045},
  year      = {2019}
}

@article{TPAMI19/Mask_TextSpotter++/Liao,
  author    = {Minghui Liao and
               Pengyuan Lyu and
               Minghang He and
               Cong Yao and
               Wenhao Wu and
               Xiang Bai},
  title     = {Mask {T}ext{S}potter: An End-to-End Trainable Neural Network for Spotting Text with Arbitrary Shapes},
  journal   = {{IEEE} Trans. Pattern Anal. Mach. Intell.},
  volume       = {43},
  number       = {2},
  pages        = {532--548},
  year         = {2021}
}

@article{TMM19/MS-CAFA/Dai,
  author    = {Pengwen Dai and
               Hua Zhang and
               Xiaochun Cao},
  title     = {Deep Multi-Scale Context Aware Feature Aggregation for Curved Scene Text Detection},
  journal   = {{IEEE} Trans. Multimedia},
  volume    = {22},
  number    = {8},
  pages     = {1969--1984},
  year      = {2020}
}

@article{TIP20/MTTD/Liu,
  author    = {Yuliang Liu and
               Lianwen Jin and
               ChuanMing Fang},
  title     = {Arbitrarily Shaped Scene Text Detection With a Mask Tightness Text Detector},
  journal   = {{IEEE} Trans. Image Process.},
  volume    = {29},
  pages     = {2918--2930},
  year      = {2020}
}

@inproceedings{ICCV17/Mask_RCNN/He,
  title={Mask {R-CNN}},
  author={He, Kaiming and Gkioxari, Georgia and Doll{\'a}r, Piotr and Girshick, Ross},
  booktitle={ICCV},
  pages={2980--2988},
  year={2017}
}

@inproceedings{ECCV16/CTPN/Tian,
  title={Detecting text in natural image with connectionist text proposal network},
  author={Tian, Zhi and Huang, Weilin and He, Tong and He, Pan and Qiao, Yu},
  booktitle={ECCV},
  pages     = {56--72},
  year={2016}
}

@article{PR19/ICG/Tang,
  author    = {Jun Tang and
               Zhibo Yang and
               Yongpan Wang and
               Qi Zheng and
               Yongchao Xu and
               Xiang Bai},
  title     = {Seg{L}ink++: Detecting Dense and Arbitrary-shaped Scene Text by Instance-aware Component Grouping},
  journal   = {Pattern Recognit.},
  volume    = {96},
  year      = {2019}
}

@inproceedings{CVPR19/CRAFT/Baek,
  author    = {Youngmin Baek and
               Bado Lee and
               Dongyoon Han and
               Sangdoo Yun and
               Hwalsuk Lee},
  title     = {Character Region Awareness for Text Detection},
  booktitle={CVPR},
  pages     = {9365--9374},
  year={2019}
}

@inproceedings{CVPR19/LOMO/Zhang,
  author={Chengquan Zhang and
  	      Borong Liang and
  	      Zuming Huang and
  	      Mengyi En and
  	      Junyu Han and 
  	      Errui Ding 
  	      and Xinghao Ding},
  title={Look {M}ore {T}han {O}nce: An Accurate Detector for Text of Arbitrary Shapes},
  booktitle={CVPR},
  pages     = {10552--10561},
  year={2019}
}

@inproceedings{CVPR19/CSE/Liu,
  author    = {Zichuan Liu and
               Guosheng Lin and
               Sheng Yang and
               Fayao Liu and
               Weisi Lin and
               Wang Ling Goh},
  title     = {Towards Robust Curve Text Detection with Conditional Spatial Expansion},
  booktitle = {CVPR},
  pages     = {7269--7278},
  year      = {2019}
}

@inproceedings{ICCV19/CharNet/Xing,
  author    = {Linjie Xing and
               Zhi Tian and
               Weilin Huang and
               Matthew R. Scott},
  title     = {Convolutional Character Networks},
  booktitle = {ICCV},
  pages     = {9125--9135},
  year      = {2019}
}

@inproceedings{ICCV19/TextDragon/Feng,
  author    = {Wei Feng and
               Wenhao He and
               Fei Yin and
               Xu{-}Yao Zhang and
               Cheng{-}Lin Liu},
  title     = {Text{D}ragon: An End-to-End Framework for Arbitrary Shaped Text Spotting},
  booktitle = {ICCV},
  pages     = {9075--9084},
  year      = {2019}
}

@inproceedings{AAAI20/DB/Liao,
  author    = {Minghui Liao and
               Zhaoyi Wan and
               Cong Yao and
               Kai Chen and
               Xiang Bai},
  title     = {Real-Time Scene Text Detection with Differentiable Binarization},
  booktitle = {AAAI},
  pages     = {11474--11481},
  year      = {2020}
}

@inproceedings{AAAI20/TextPerceptron/Qiao,
  author    = {Liang Qiao and
               Sanli Tang and
               Zhanzhan Cheng and
               Yunlu Xu and
               Yi Niu and
               Shiliang Pu and
               Fei Wu},
  title     = {Text {P}erceptron: Towards End-to-End Arbitrary-Shaped Text Spotting},
  booktitle = {AAAI},
  pages     = {11899--11907},
  year      = {2020}
}

@inproceedings{AAAI20/AYNIB/Wang,
  author    = {Hao Wang and
               Pu Lu and
               Hui Zhang and
               Mingkun Yang and
               Xiang Bai and
               Yongchao Xu and
               Mengchao He and
               Yongpan Wang and
               Wenyu Liu},
  title     = {All {Y}ou {N}eed {I}s {B}oundary: Toward Arbitrary-Shaped Text Spotting},
  booktitle = {AAAI},
  pages     = {12160--12167},
  year      = {2020}
}

@inproceedings{bu2023srformer,
  title={{SRF}ormer: Empowering regression-based text detection transformer with segmentation},
  author={Bu, Qingwen and Park, Sungrae and Khang, Minsoo and Cheng, Yichuan},
  booktitle = {AAAI},
  pages     = {855--863},
  year      = {2024}
}

@inproceedings{CVPR20/DRRGN/Zhang,
  author    = {Shi{-}Xue Zhang and
               Xiaobin Zhu and
               Jie{-}Bo Hou and
               Chang Liu and
               Chun Yang and
               Hongfa Wang and
               Xu{-}Cheng Yin},
  title     = {Deep Relational Reasoning Graph Network for Arbitrary Shape Text Detection},
  booktitle   = {CVPR},
  pages     = {9696-9705},
  year      = {2020}
}

@inproceedings{CVPR20/ContourNet/Wang,
  author    = {Yuxin Wang and
               Hongtao Xie and
               Zhengjun Zha and
               Mengting Xing and
               Zilong Fu and
               Yongdong Zhang},
  title     = {Contour{N}et: Taking a Further Step toward Accurate Arbitrary-shaped Scene Text Detection},
  booktitle   = {CVPR},            
  pages     = {11750--11759},
  year      = {2020}
}

@inproceedings{AAAI17/TextBox/Liao,
  author    = {Minghui Liao and
               Baoguang Shi and
               Xiang Bai and
               Xinggang Wang and
               Wenyu Liu},
  title     = {Text{B}oxes: {A} Fast Text Detector with a Single Deep Neural Network},
  booktitle = {AAAI},
  pages     = {4161--4167},
  year      = {2017}
}

@inproceedings{ECCV16/SSD/Liu,
  author    = {Wei Liu and
               Dragomir Anguelov and
               Dumitru Erhan and
               Christian Szegedy and
               Scott E. Reed and
               Cheng{-}Yang Fu and
               Alexander C. Berg},
  title     = {{SSD:} Single Shot MultiBox Detector},
  booktitle = {ECCV},
  pages     = {21--37},
  year      = {2016}
}

@article{TIP18/TextBoxes++/Liao,
  author    = {Minghui Liao and
               Baoguang Shi and
               Xiang Bai},
  title     = {Text{B}oxes++: {A} Single-Shot Oriented Scene Text Detector},
  journal   = {{IEEE} Trans. Image Process.},
  volume    = {27},
  number    = {8},
  pages     = {3676--3690},
  year      = {2018}
}

@inproceedings{ICCV17/SSD_RA/HePan,
  title={Single Shot Text Detector with Regional Attention},
  author={He, Pan and Huang, Weilin and He, Tong and Zhu, Qile and Qiao, Yu and Li, Xiaolin},
  booktitle={ICCV},
  pages     = {3066--3074},
  year={2017}
}

@inproceedings{CVPR17/DMPN/Liu,
  author    = {Yuliang Liu and Lianwen Jin},
  title     = {{D}eep {M}atching {P}rior {N}etwork: Toward Tighter Multi-oriented Text Detection},
  booktitle = {CVPR},
  pages     = {3454--3461},
  year      = {2017}
}

@inproceedings{CVPR18/ITN/Wang,
  author    = {Fangfang Wang and
               Liming Zhao and
               Xi Li and
               Xinchao Wang and
               Dacheng Tao},
  title     = {Geometry-Aware Scene Text Detection With Instance Transformation Network},
  booktitle = {CVPR},
  pages     = {1381--1389},
  year      = {2018}
}

@inproceedings{CVPR18/RRD/Liao,
  title={Rotation-Sensitive Regression for Oriented Scene Text Detection},
  author={Liao, Minghui and Zhu, Zhen and Shi, Baoguang and Xia, Gui-song and Bai, Xiang},
  booktitle={CVPR},
  pages     = {5909--5918},
  year={2018}
}

@article{PR19/CTD_CLOC/Liu,
  title={Curved scene text detection via transverse and longitudinal sequence connection},
  author={Liu, Yuliang and Jin, Lianwen and Zhang, Shuaitao and Zhang, Sheng},
  journal={Pattern Recognit.},
  volume    = {90},
  pages     = {337--345},
  year={2019}
}

@inproceedings{ICDAR17/ToT,
  title={Total-{TEXT}: A comprehensive dataset for scene text detection and recognition},
  author={Ch'ng, Chee Kheng and Chan, Chee Seng},
  booktitle={ICDAR},
  pages={935--942},
  year={2017}
}

@article{IJDAR20/TOT/chng,
  author    = {Chee{-}Kheng Ch'ng and
               Chee Seng Chan and
               Cheng{-}Lin Liu},
  title     = {Total-{T}ext: toward orientation robustness in scene text detection},
  journal   = {Int. J. Document Anal. Recognit.},
  volume    = {23},
  number    = {1},
  pages     = {31--52},
  year      = {2020}
}

@inproceedings{CVPR19/ATRR/Wang,
  author    = {Xiaobing Wang and
               Yingying Jiang and
               Zhenbo Luo and
               Cheng{-}Lin Liu and
               Hyunsoo Choi and
               Sungjin Kim},
  title     = {Arbitrary Shape Scene Text Detection with Adaptive Text Region Representation},
  booktitle = {CVPR},
  pages     = {6449--6458},
  year      = {2019}

}

@article{TOMM19/AB_LSTM/Liu,
  title={{AB-LSTM}: Attention-based Bidirectional LSTM Model for Scene Text Detection},
  author={Liu, Zhandong and Zhou, Wengang and Li, Houqiang},
  journal={ACM Trans. Multimedia Computing, Communications, and Applications},
  volume={15},
  number={4},
  pages={1--23},
  year={2019}
}

@inproceedings{IJCAI19/MSR/Xue,
  author    = {Chuhui Xue and
               Shijian Lu and
               Wei Zhang},
  title     = {{MSR:} Multi-Scale Shape Regression for Scene Text Detection},
  booktitle = {IJCAI},
  pages     = {989--995},
  year      = {2019}
}

@inproceedings{ICDAR09/Hanif,
  title={Text detection and localization in complex scene images using constrained adaboost algorithm},
  author={Hanif, Shehzad Muhammad and Prevost, Lionel},
  booktitle={ICDAR},
  pages={1--5},
  year={2009}
}

@inproceedings{CVPR15/Symmetry/Zhang,
  title={Symmetry-based text line detection in natural scenes},
  author={Zhang, Zheng and Shen, Wei and Yao, Cong and Bai, Xiang},
  booktitle={CVPR},
  pages={2558--2567},
  year={2015}
}

@inproceedings{CVPR10/SWT/Boris,
  title={Detecting text in natural scenes with stroke width transform},
  author={Epshtein, Boris and Ofek, Eyal and Wexler, Yonatan},
  booktitle={CVPR},
  pages={2963--2970},
  year={2010}
}

@article{TMM11/SDTL/Bouman,
  author    = {Katherine L. Bouman and
               Golnaz Abdollahian and
               Mireille Boutin and
               Edward J. Delp},
  title     = {A Low Complexity Sign Detection and Text Localization Method for Mobile Applications},
  journal   = {{IEEE} Trans. Multimedia},
  volume    = {13},
  number    = {5},
  pages     = {922--934},
  year      = {2011}
}

@inproceedings{ICCV13/SFT/Huang,
  title={Text localization in natural images using stroke feature transform and text covariance descriptors},
  author={Huang, Weilin and Lin, Zhe and Yang, Jianchao and Wang, Jue},
  booktitle={ICCV},
  pages={1241--1248},
  year={2013}
}

@article{TIP14/Characterness/Li,
  title={Characterness: An indicator of text in the wild},
  author={Li, Yao and Jia, Wenjing and Shen, Chunhua and van den Hengel, Anton},
  journal={IEEE Trans. Image Process. },
  volume={23},
  number={4},
  pages={1666--1677},
  year={2014},
  publisher={IEEE}
}

@inproceedings{ECCV14/CNN_MSER/Huang,
  author    = {Weilin Huang and
               Yu Qiao and
               Xiaoou Tang},
  title     = {Robust Scene Text Detection with Convolution Neural Network Induced
               {MSER} Trees},
  booktitle = {ECCV},
  pages     = {497--511},
  year      = {2014}
}

@article{TMM18/Tang/SSFT,
  title={Scene Text Detection using Superpixel based Stroke Feature Transform and Deep Learning based Region Classification},
  author={Tang, Youbao and Wu, Xiangqian},
  journal={IEEE Trans. Multimedia},
  volume       = {20},
  number       = {9},
  pages        = {2276--2288},
  year={2018}
}

@inproceedings{ICDAR17/DSN/Wu,
  author    = {Dao Wu and
               Rui Wang and
               Pengwen Dai and
               Yueying Zhang and
               Xiaochun Cao},
  title     = {Deep Strip-Based Network with Cascade Learning for Scene Text Localization},
  booktitle = {ICDAR},
  pages     = {826--831},
  year      = {2017}
}

@inproceedings{CVPR16/Canny,
  title={Canny text detector: Fast and robust scene text localization algorithm},
  author={Cho, Hojin and Sung, Myungchul and Jun, Bongjin},
  booktitle={CVPR},
  pages={3566--3573},
  year={2016}
}

@inproceedings{ACMMM16/MOTD_VAM,
  title={Detecting Arbitrary Oriented Text in the Wild with a Visual Attention Model},
  author={Huang, Wenyi and He, Dafang and Yang, Xiao and Zhou, Zihan and Kifer, Daniel and Giles, C Lee},
  booktitle={ACM-MM},
  pages={551--555},
  year={2016},
}

@inproceedings{ICCV17/DR/He,
  title={Deep Direct Regression for Multi-Oriented Scene Text Detection},
  author={He, Wenhao and Zhang, Xu-Yao and Yin, Fei and Liu, Cheng-Lin},
  booktitle={ICCV},
  pages     = {745--753},
  year={2017}
}

@inproceedings{ECCV18/Textsnake/Long,
  author    = {Shangbang Long and
               Jiaqiang Ruan and
               Wenjie Zhang and
               Xin He and
               Wenhao Wu and
               Cong Yao},
  title     = {Text{S}nake: {A} Flexible Representation for Detecting Text of Arbitrary Shapes},
  booktitle = {ECCV},
  pages     = {19--35},
  year      = {2018}
}

@ARTICLE{HGR-Net,
  author={Bi, Hengyue and Xu, Canhui and Shi, Cao and Liu, Guozhu and Zhang, Honghong and Li, Yuteng and Dong, Junyu},
  journal={IEEE Trans. Image Process.}, 
  title={HGR-Net: Hierarchical Graph Reasoning Network for Arbitrary Shape Scene Text Detection}, 
  year={2023},
  volume={32},
  pages={4142-4155}
  }

@ARTICLE{ADNet,
  author={Qu, Yadong and Xie, Hongtao and Fang, Shancheng and Wang, Yuxin and Zhang, Yongdong},
  journal={IEEE Trans. Multimedia}, 
  title={ADNet: Rethinking the Shrunk Polygon-Based Approach in Scene Text Detection}, 
  year={2023},
  volume={25},
  pages={6983-6996}
 }

@ARTICLE{FSAS,
  author={Wang, Fangfang and Xu, Xiaogang and Chen, Yifeng and Li, Xi},
  journal={IEEE Trans. Image Process.}, 
  title={Fuzzy Semantics for Arbitrary-Shaped Scene Text Detection}, 
  year={2023},
  volume={32},
  pages={1-12}
  }

@ARTICLE{STSEA,
  author={Tang, Zhengmi and Miyazaki, Tomo and Omachi, Shinichiro},
  journal={IEEE Trans. Image Process.}, 
  title={A Scene-Text Synthesis Engine Achieved Through Learning From Decomposed Real-World Data}, 
  year={2023},
  volume={32},
  number={},
  pages={5837-5851}
 }

@ARTICLE{RNet,
  author={Wang, Yuxin and Xie, Hongtao and Zha, Zhengjun and Tian, Youliang and Fu, Zilong and Zhang, Yongdong},
  journal={IEEE Trans. Multimedia}, 
  title={R-Net: A Relationship Network for Efficient and Accurate Scene Text Detection}, 
  year={2021},
  volume={23},
  number={},
  pages={1316-1329}
  }

@ARTICLE{TextPMs,
  author={Zhang, Shi-Xue and Zhu, Xiaobin and Chen, Lei and Hou, Jie-Bo and Yin, Xu-Cheng},
  journal={IEEE Trans. Pattern Anal. Mach. Intell.}, 
  title={Arbitrary Shape Text Detection via Segmentation With Probability Maps}, 
  year={2023},
  volume={45},
  number={3},
  pages={2736-2750}
  }

@INPROCEEDINGS{UNITS,
  author={Kil, Taeho and Kim, Seonghyeon and Seo, Sukmin and Kim, Yoonsik and Kim, Daehee},
  booktitle={CVPR}, 
  title={Towards Unified Scene Text Spotting Based on Sequence Generation}, 
  year={2023},
  pages={15223-15232}
}

@ARTICLE{NLIDC,
  author={Shivakumara, Palaiahnakote and Banerjee, Ayan and Pal, Umapada and Nandanwar, Lokesh and Lu, Tong and Liu, Cheng-Lin},
  journal={IEEE Trans. Image Process.}, 
  title={A New Language-Independent Deep CNN for Scene Text Detection and Style Transfer in Social Media Images}, 
  year={2023},
  volume={32},
  pages={3552-3566},
  }

@inproceedings{ICCV15/Tian/TextFlow,
  author    = {Shangxuan Tian and
               Yifeng Pan and
               Chang Huang and
               Shijian Lu and
               Kai Yu and
               Chew Lim Tan},
  title     = {Text {F}low: {A} Unified Text Detection System in Natural Scene Images},
  booktitle = {ICCV},
  pages     = {4651--4659},
  year      = {2015}
}

@inproceedings{MLT17/competition/Nayef,
  author    = {Nibal Nayef and
               Fei Yin and
               Imen Bizid and
               et al.},
  title     = {{ICDAR2017} Robust Reading Challenge on Multi-Lingual Scene Text Detection and Script Identification - {RRC-MLT}},
  booktitle = {ICDAR},
  pages     = {1454--1459},
  year      = {2017}
}

@inproceedings{ye2022dptext,
  title={DPText-DETR: Towards Better Scene Text Detection with Dynamic Points in Transformer},
  author={Ye, Maoyuan and Zhang, Jing and Zhao, Shanshan and Liu, Juhua and Du, Bo and Tao, Dacheng},
  booktitle={AAAI},
  volume={37},
  number={3},
  pages={3241--3249},
  year={2023}
}

@ARTICLE{tmm23adnet,
  author={Qu, Yadong and Xie, Hongtao and Fang, Shancheng and Wang, Yuxin and Zhang, Yongdong},
  journal={IEEE Trans. Multimedia}, 
  title={ADNet: Rethinking the Shrunk Polygon-Based Approach in Scene Text Detection}, 
  year={2023},
  volume={25},
  number={},
  pages={6983-6996},
 }

@ARTICLE{tip23fsa,
  author={Wang, Fangfang and Xu, Xiaogang and Chen, Yifeng and Li, Xi},
  journal={IEEE Trans. Image Process.}, 
  title={Fuzzy Semantics for Arbitrary-Shaped Scene Text Detection}, 
  year={2023},
  volume={32},
  number={},
  pages={1-12}
  }

@ARTICLE{22dbnet++,
  author={Liao, Minghui and Zou, Zhisheng and Wan, Zhaoyi and Yao, Cong and Bai, Xiang},
  journal={IEEE Trans. Pattern Anal. Mach. Intell}, 
  title={Real-Time Scene Text Detection With Differentiable Binarization and Adaptive Scale Fusion}, 
  year={2023},
  volume={45},
  number={1},
  pages={919-931},
  }

@inproceedings{wu2022decoupling,
  title={Decoupling recognition from detection: Single shot self-reliant scene text spotter},
  author={Wu, Jingjing and Lyu, Pengyuan and Lu, Guangming and Zhang, Chengquan and Yao, Kun and Pei, Wenjie},
  booktitle={ACM-MM},
  pages={1319--1328},
  year={2022}
}

@inproceedings{SIRACMM23,
author = {Qin, Xugong and Lyu, Pengyuan and Zhang, Chengquan and Zhou, Yu and Yao, Kun and Zhang, Peng and Lin, Hailun and Wang, Weiping},
title = {Towards Robust Real-Time Scene Text Detection: From Semantic to Instance Representation Learning},
year = {2023},
booktitle = {ACM-MM},
pages = {2025–2034},
numpages = {10},
}

@article{TextBPN++,
  title={Arbitrary shape text detection via boundary transformer},
  author={Zhang, Shi-Xue and Yang, Chun and Zhu, Xiaobin and Yin, Xu-Cheng},
  journal={IEEE Trans. Multimedia},
  year={2024},
  volume       = {26},
  pages        = {1747--1760},
  publisher={IEEE}
}

@article{DeepSolo++,
  author       = {Maoyuan Ye and
                  Jing Zhang and
                  Shanshan Zhao and
                  Juhua Liu and
                  Tongliang Liu and
                  Bo Du and
                  Dacheng Tao},
  title        = {DeepSolo++: Let Transformer Decoder with Explicit Points Solo for Text Spotting},
  journal      = {CoRR},
  volume       = {abs/2305.19957},
  year         = {2023},
  eprinttype    = {arXiv}
}

@INPROCEEDINGS{PCR21,
  author={Dai, Pengwen and Zhang, Sanyi and Zhang, Hua and Cao, Xiaochun},
  booktitle={CVPR}, 
  title={Progressive Contour Regression for Arbitrary-Shape Scene Text Detection}, 
  year={2021},
  pages={7389-7398}
  }

@INPROCEEDINGS{TextBPN,
  author={Zhang, Shi-Xue and Zhu, Xiaobin and Yang, Chun and Wang, Hongfa and Yin, Xu-Cheng},
  booktitle={ICCV}, 
  title={Adaptive Boundary Proposal Network for Arbitrary Shape Text Detection}, 
  year={2021},
  pages={1285-1294}
  }

@ARTICLE{OPMP,
  author={Zhang, Sheng and Liu, Yuliang and Jin, Lianwen and Wei, Zhongrong and Shen, Chunhua},
  journal={IEEE Trans. Multimedia}, 
  title={OPMP: An Omnidirectional Pyramid Mask Proposal Network for Arbitrary-Shape Scene Text Detection}, 
  year={2021},
  volume={23},
  number={},
  pages={454-467}
  }

@ARTICLE{Boundary-TextSpotter22,
  author={Lu, Pu and Wang, Hao and Zhu, Shenggao and Wang, Jing and Bai, Xiang and Liu, Wenyu},
  journal={IEEE Trans. Image Process.}, 
  title={Boundary TextSpotter: Toward Arbitrary-Shaped Scene Text Spotting}, 
  year={2022},
  volume={31},
  number={},
  pages={6200-6212}
  }

@INPROCEEDINGS{SwinTextSpotter,
  author={Huang, Mingxin and Liu, Yuliang and Peng, Zhenghao and Liu, Chongyu and Lin, Dahua and Zhu, Shenggao and Yuan, Nicholas and Ding, Kai and Jin, Lianwen},
  booktitle={CVPR}, 
  title={SwinTextSpotter: Scene Text Spotting via Better Synergy between Text Detection and Text Recognition}, 
  year={2022},
  pages={4583-4593}
  }

@inproceedings{gupta2016synthetic,
  title={Synthetic data for text localisation in natural images},
  author={Gupta, Ankush and Vedaldi, Andrea and Zisserman, Andrew},
  booktitle={CVPR},
  pages={2315--2324},
  year={2016}
}

@inproceedings{TPSNet2022,
author = {Wang, Wei and Zhou, Yu and Lv, Jiahao and Wu, Dayan and Zhao, Guoqing and Jiang, Ning and Wang, Weipinng},
title = {{TPSNet}: Reverse Thinking of Thin Plate Splines for Arbitrary Shape Scene Text Representation},
year = {2022},
booktitle = {ACM-MM},
pages = {5014–5025},
numpages = {12},
location = {, Lisboa, Portugal, },
}

@ARTICLE{EMA,
  author={Zhao, Mengbiao and Feng, Wei and Yin, Fei and Zhang, Xu-Yao and Liu, Cheng-Lin},
  journal={IEEE Trans. Image Process.}, 
  title={Mixed-Supervised Scene Text Detection With Expectation-Maximization Algorithm}, 
  year={2022},
  volume={31},
  pages={5513-5528}
  }

@INPROCEEDINGS{TESTR,
  author={Zhang, Xiang and Su, Yongwen and Tripathi, Subarna and Tu, Zhuowen},
  booktitle={CVPR}, 
  title={Text Spotting Transformers}, 
  year={2022},
  pages={9509-9518}
  }

@INPROCEEDINGS{TTS22cvpr,
  author={Kittenplon, Yair and Lavi, Inbal and Fogel, Sharon and Bar, Yarin and Manmatha, R. and Perona, Pietro},
  booktitle={CVPR}, 
  title={Towards Weakly-Supervised Text Spotting using a Multi-Task Transformer}, 
  year={2022},
  pages={4594-4603}
}

@article{du2022i3cl,
  title={I3CL: Intra-and Inter-Instance Collaborative Learning for Arbitrary-shaped Scene Text Detection},
  author={Du, Bo and Ye, Jian and Zhang, Jing and Liu, Juhua and Tao, Dacheng},
  journal={Int. J. Comput. Vis.},
  volume={130},
  number={8},
  pages={1961--1977},
  year={2022},
  publisher={Springer}
}

@INPROCEEDINGS{FSGNet,
  author={Tang, Jingqun and Zhang, Wenqing and Liu, Hongye and Yang, MingKun and Jiang, Bo and Hu, Guanglong and Bai, Xiang},
  booktitle={CVPR}, 
  title={Few Could Be Better Than All: Feature Sampling and Grouping for Scene Text Detection}, 
  year={2022},
  pages={4553-4562}
  }

@article{KPN,
  title={Kernel proposal network for arbitrary shape text detection},
  author={Zhang, Shi-Xue and Zhu, Xiaobin and Hou, Jie-Bo and Yang, Chun and Yin, Xu-Cheng},
  journal={IEEE Trans. Neural Networks and Learning Systems},
  volume       = {34},
  number       = {11},
  pages        = {8731--8742},
  year         = {2023}
}

@InProceedings{ESTextSpotter,
    author    = {Huang, Mingxin and Zhang, Jiaxin and Peng, Dezhi and Lu, Hao and Huang, Can and Liu, Yuliang and Bai, Xiang and Jin, Lianwen},
    title     = {ESTextSpotter: Towards Better Scene Text Spotting with Explicit Synergy in Transformer},
    booktitle = {ICCV},
    year      = {2023},
    pages     = {19495-19505}
}

@INPROCEEDINGS{deepsolo,
  author={Ye, Maoyuan and Zhang, Jing and Zhao, Shanshan and Liu, Juhua and Liu, Tongliang and Du, Bo and Tao, Dacheng},
  booktitle={CVPR}, 
  title={DeepSolo: Let Transformer Decoder with Explicit Points Solo for Text Spotting}, 
  year={2023},
  pages={19348-19357}
}

@ARTICLE{DBNet++,
  author={Liao, Minghui and Zou, Zhisheng and Wan, Zhaoyi and Yao, Cong and Bai, Xiang},
  journal={IEEE Trans. Pattern. Anal. Mach. Intell.}, 
  title={Real-Time Scene Text Detection With Differentiable Binarization and Adaptive Scale Fusion}, 
  year={2023},
  volume={45},
  number={1},
  pages={919-931}
  }

@article{TIP20/Mask-TTD/Liu,
  author    = {Yuliang Liu and
               Lianwen Jin and
               ChuanMing Fang},
  title     = {Arbitrarily Shaped Scene Text Detection With a Mask  
               Tightness Text Detector},
  journal   = {{IEEE} Trans. Image Process.},
  volume    = {29},
  pages     = {2918--2930},
  year      = {2020}
}

@inproceedings{IJCAI19/DSRN/Wang,
  author    = {Yuxin Wang and
               Hongtao Xie and
               Zilong Fu and
               Yongdong Zhang},
  title     = {{DSRN:} {A} Deep Scale Relationship Network for Scene Text Detection},
  booktitle = {IJCAI},
  pages     = {947--953},
  year      = {2019}
}

@inproceedings{CVPR17/EAST/Zhou,
  title={{EAST}: An Efficient and Accurate Scene Text Detector},
  author={Zhou, Xinyu and Yao, Cong and Wen, He and Wang, Yuzhi and Zhou, Shuchang and He, Weiran and Liang, Jiajun},
  booktitle={CVPR},
  pages     = {2642--2651},
  year={2017}
}

@inproceedings{MM20/CRNet/Zhou,
  author    = {Yu Zhou and
               Hongtao Xie and
               Shancheng Fang and
               Yan Li and
               Yongdong Zhang},
  title     = {{CRN}et: {A} Center-aware Representation for Detecting Text of Arbitrary Shapes},
  booktitle = {ACM-MM},
  pages     = {2571--2580},
  year      = {2020}
}

@inproceedings{ECCV20/SDM/Xiao,
  author    = {Shanyu Xiao and
               Liangrui Peng and
               Ruijie Yan and
               Keyu An and
               Gang Yao and
               Jaesik Min},
  title = {Sequential Deformation for Accurate Scene Text Detection},
  booktitle = {ECCV},
  pages     = {108--124},
  year      = {2020}
}

@inproceedings{IJCAI20/TextFuseNet/Ye,
  author    = {Jian Ye and
               Zhe Chen and
               Juhua Liu and
               Bo Du},
  title     = {Text{F}use{N}et: Scene Text Detection with Richer Fused Features},
  booktitle = {IJCAI},
  pages     = {516--522},
  year      = {2020}
}

@article{PR21/ReLaText/Ma,
  author    = {Chixiang Ma and
               Lei Sun and
               Zhuoyao Zhong and
               Qiang Huo},
  title     = {Re{L}a{T}ext: Exploiting visual relationships for arbitrary-shaped scene text detection with graph convolutional networks},
  journal   = {Pattern Recognit.},
  volume    = {111},
  pages     = {107684},
  year      = {2021}
}

@inproceedings{ECCV20/TextCaps/Sidorov,
  author    = {Oleksii Sidorov and
               Ronghang Hu and
               Marcus Rohrbach and
               Amanpreet Singh},
  title     = {Text{C}aps: {A} Dataset for Image Captioning with Reading Comprehension},
  booktitle = {ECCV},
  pages     = {742--758},
  year      = {2020}
}

@inproceedings{ECCV16/ResNet_v2/He,
  author    = {Kaiming He and
               Xiangyu Zhang and
               Shaoqing Ren and
               Jian Sun},
  title     = {Identity Mappings in Deep Residual Networks},
  booktitle = {ECCV},
  pages     = {630--645},
  year      = {2016}
}

@inproceedings{CVPR16/ResNet_v1/He,
  author    = {Kaiming He and
               Xiangyu Zhang and
               Shaoqing Ren and
               Jian Sun},
  title= {Deep Residual Learning for Image Recognition},
  booktitle = {CVPR},
  pages= {770--778},
  year= {2016}
}

@article{IJCV15/ImageNet/Russakovsky,
  author    = {Olga Russakovsky and
               Jia Deng and
               Hao Su and
               Jonathan Krause and
               Sanjeev Satheesh and
               Sean Ma and
               Zhiheng Huang and
               Andrej Karpathy and
               Aditya Khosla and
               Michael S. Bernstein and
               Alexander C. Berg and
               Fei{-}Fei Li},
  title     = {ImageNet Large Scale Visual Recognition Challenge},
  journal   = {Int. J. Comput. Vis.},
  volume    = {115},
  number    = {3},
  pages     = {211--252},
  year      = {2015}
}

@inproceedings{ICDAR17/COCO_Text/Gomez,
  author    = {Raul Gomez and
               Baoguang Shi and
               Lluis Gomez{-}Bigorda and
               Lukas Neumann and
               Andreas Veit and
               Jiri Matas and
               Serge J. Belongie and
               Dimosthenis Karatzas},
  title     = {{ICDAR2017} Robust Reading Challenge on {COCO-T}ext},
  booktitle = {ICDAR},
  pages     = {1435--1443},
  year      = {2017}
}

@INPROCEEDINGS{LayoutFormer,
  author={Liang, Min and Ma, Jia-Wei and Zhu, Xiaobin and Qin, Jingyan and Yin, Xu-Cheng},
  booktitle={CVPR}, 
  title={Layout{F}ormer: Hierarchical Text Detection Towards Scene Text Understanding}, 
  year={2024},
  pages={15665-15674}
  }

@inproceedings{FCENet,
  title={Fourier contour embedding for arbitrary-shaped text detection},
  author={Zhu, Yiqin and Chen, Jianyong and Liang, Lingyu and Kuang, Zhanghui and Jin, Lianwen and Zhang, Wayne},
  booktitle={CVPR},
  pages={3123--3131},
  year={2021}
}

@inproceedings{su2024lranet,
  title={LRANet: towards accurate and efficient scene text detection with low-rank approximation network},
  author={Su, Yuchen and Chen, Zhineng and Shao, Zhiwen and Du, Yuning and Ji, Zhilong and Bai, Jinfeng and Zhou, Yong and Jiang, Yu-Gang},
  booktitle={AAAI},
  volume={38},
  number={5},
  pages={4979--4987},
  year={2024}
}

@inproceedings{chen2024box2poly,
  title={Box2Poly: memory-efficient polygon prediction of arbitrarily shaped and rotated text},
  author={Chen, Xuyang and Wang, Dong and Schindler, Konrad and Sun, Mingwei and Wang, Yongliang and Savioli, Nicolo and Meng, Liqiu},
  booktitle={AAAI},
  volume={38},
  number={2},
  pages={1219--1227},
  year={2024}
}

@inproceedings{zheng2024kernel,
  title={Kernel adaptive convolution for scene text detection via distance map prediction},
  author={Zheng, Jinzhi and Fan, Heng and Zhang, Libo},
  booktitle={CVPR},
  pages={5957--5966},
  year={2024}
}

@article{IAST2024,
  author       = {Shi{-}Xue Zhang and
                  Chun Yang and
                  Xiaobin Zhu and
                  Hongyang Zhou and
                  Hongfa Wang and
                  Xu{-}Cheng Yin},
  title        = {Inverse-Like Antagonistic Scene Text Spotting via Reading-Order Estimation and Dynamic Sampling},
  journal      = {{IEEE} Trans. Image Process.},
  volume       = {33},
  pages        = {825--839},
  year         = {2024}
}

@article{TNNLS23/ZTD/Yang,
  title={Zoom text detector},
  author={Yang, Chuang and Chen, Mulin and Yuan, Yuan and Wang, Qi},
  journal={IEEE Trans. on Neural Net. and Learn. Sys.},
volume       = {35},
  number       = {11},
  pages        = {15745--15757},
  year         = {2024}
}

@article{TMM23/RP-Text/Wang,
  title={Region-aware arbitrary-shaped text detection with progressive fusion},
  author={Wang, Qitong and Fu, Bin and Li, Ming and He, Junjun and Peng, Xi and Qiao, Yu},
  journal={{IEEE} Trans. Multimedia},
  volume={25},
  pages={4718--4729},
  year={2022},
  publisher={IEEE}
}

@article{TMM23/Morphtext/Xu,
  title={MorphText: Deep morphology regularized accurate arbitrary-shape scene text detection},
  author={Xu, Chengpei and Jia, Wenjing and Wang, Ruomei and Luo, Xiaonan and He, Xiangjian},
  journal={{IEEE} Trans. Multimedia},
  volume={25},
  pages={4199--4212},
  year={2022},
  publisher={IEEE}
}

@article{TMM23/RSMTD/Yang,
  title={Reinforcement shrink-mask for text detection},
  author={Yang, Chuang and Chen, Mulin and Yuan, Yuan and Wang, Qi},
  journal={{IEEE} Trans. Multimedia},
  volume={25},
  pages={6458--6470},
  year={2022},
  publisher={IEEE}
}

@article{TITS23/HFENet/Liang,
  title={{HFENet}: Hybrid feature enhancement network for detecting texts in scenes and traffic panels},
  author={Liang, Min and Zhu, Xiaobin and Zhou, Hongyang and Qin, Jingyan and Yin, Xu-Cheng},
  journal={IEEE Trans. on Intell. Transport. Syst.},
  volume={24},
  number={12},
  pages={14200--14212},
  year={2023},
  publisher={IEEE}
}

@article{TMM23/TextDCT/Su,
  title={Textdct: Arbitrary-shaped text detection via discrete cosine transform mask},
  author={Su, Yuchen and Shao, Zhiwen and Zhou, Yong and Meng, Fanrong and Zhu, Hancheng and Liu, Bing and Yao, Rui},
  journal={{IEEE} Trans. Multimedia},
  volume={25},
  pages={5030--5042},
  year={2022},
  publisher={IEEE}
}

@article{TCSVT24/CT-Net/Shao,
  title={CT-Net: Arbitrary-shaped text detection via contour transformer},
  author={Shao, Zhiwen and Su, Yuchen and Zhou, Yong and Meng, Fanrong and Zhu, Hancheng and Liu, Bing and Yao, Rui},
  journal={IEEE Trans. Circuit Syst. Video Technol.},
  volume={34},
  number={3},
  pages={1815--1826},
  year={2023},
  publisher={IEEE}
}

@article{TMM24/Zhangetal/Zhang,
  title={Arbitrary shape text detection via boundary transformer},
  author={Zhang, Shi-Xue and Yang, Chun and Zhu, Xiaobin and Yin, Xu-Cheng},
  journal={{IEEE} Trans. Multimedia},
  volume={26},
  pages={1747--1760},
  year={2023},
  publisher={IEEE}
}

@article{ECCV22/ATTR/Zhou,
  author       = {Zhao Zhou and
                  Xiangcheng Du and
                  Yingbin Zheng and
                  Cheng Jin},
  title        = {Aggregated Text Transformer for Scene Text Detection},
  journal      = {CoRR},
  volume       = {abs/2211.13984},
  year         = {2022},
  eprinttype   = {arXiv},
}

@article{CVPR20/ABCNet/Liu,
  title={ABCNet: Real-Time Scene Text Spotting With Adaptive Bezier-Curve Network},
  author={Yuliang Liu and Hao Chen and Chunhua Shen and Tong He and Lianwen Jin and Liangwei Wang},
  journal={CVPR},
  year={2020},
  pages={9806-9815},
}

@article{TPAMI21/ABCNetV2/Liu,
  title={ABCNet v2: Adaptive Bezier-Curve Network for Real-Time End-to-End Text Spotting},
  author={Yuliang Liu and Chunhua Shen and Lianwen Jin and Tong He and Peng Chen and Chongyu Liu and Hao Chen},
  journal={{IEEE} Trans. Pattern Anal. Mach. Intell.},
  year={2021},
  volume={44},
  pages={8048-8064}
}

@inproceedings{CVPR25/DG-Brigde-Spotter/Huang,
  title={Bridging the Gap Between End-to-End and Two-Step Text Spotting},
  author={Mingxin Huang and Hongliang Li and Yuliang Liu and Xiang Bai and Lianwen Jin},
  booktitle={CVPR},
  year={2025},
  pages={15608-15618}
}

@ARTICLE{TIP22/CM-Net/Yang,
  author={Yang, Chuang and Chen, Mulin and Xiong, Zhitong and Yuan, Yuan and Wang, Qi},
  journal={{IEEE} Trans. Image Process.}, 
  title={CM-Net: Concentric Mask Based Arbitrary-Shaped Text Detection}, 
  year={2022},
  volume={31},
  pages={2864-2877}
 }

@ARTICLE{TIPS24/CR2-Net/Wang,
  author={Wang, Runmin and Hei, Jielei and Liu, Minghao and Wan, Zukun and Xu, Juan and Cao, Xiaofei and He, Xuan and Ding, Yajun and Gao, Changxin and Sang, Nong},
  journal={IEEE Transactions on Intelligent Vehicles}, 
  title={CR2-Net: Component Relationship Reasoning Network for Traffic Text Detection}, 
  year={2025},
  volume={10},
  number={1},
  pages={190-204}
  }

@article{TPAMI21/PAN++/Wang,
  title={PAN++: Towards Efficient and Accurate End-to-End Spotting of Arbitrarily-Shaped Text},
  author={Wenhai Wang and Enze Xie and Xiang Li and Xuebo Liu and Ding Liang and Yang Zhibo and Tong Lu and Chunhua Shen},
  journal={{IEEE} Trans. Pattern Anal. Mach. Intell.},
  year={2021},
  volume={44},
  pages={5349-5367}
}

@article{PR19/SegLink++/Tang,
  title={SegLink++: Detecting Dense and Arbitrary-shaped Scene Text by Instance-aware Component Grouping},
  author={Jun Tang and Zhibo Yang and Yongpan Wang and Qi Zheng and Yongchao Xu and Xiang Bai},
  journal={Pattern Recognit.},
  year={2019},
  volume={96}
}

@article{TMM25/STD/Han,
  title={Spotlight Text Detector: Spotlight on Candidate Regions Like a Camera},
  author={Xu Han and Junyu Gao and Chuan Yang and Yuan Yuan and Qi Wang},
  journal={{IEEE} Trans. Multimedia},
  year={2025},
  volume={27},
  pages={1937-1949}
}

@inproceedings{AAAI24/SRFormer/Bu,
  title={SRFormer: Text Detection Transformer with Incorporated Segmentation and Regression},
  author={Qingwen Bu and Sungrae Park and Minsoo Khang and Yi-Bin Cheng},
  booktitle={AAAI},
  pages={855--863},
  year={2024}
}

@ARTICLE{TMM23/LVT/Yang,
  author={Yang, Chuang and Chen, Mulin and Yuan, Yuan and Wang, Qi},
  journal={{IEEE} Trans. Multimedia}, 
  title={Text Growing on Leaf}, 
  year={2023},
  volume={25},
  number={},
  pages={9029-9043}
  }

@article{PR21/TextMountain/Zhu,
  title={TextMountain: Accurate Scene Text Detection via Instance Segmentation},
  author={Yixing Zhu and Jun Du},
  journal={Pattern Recognit.},
  year={2021},
  volume={110},
  pages={107336}
}

@article{ICCV19/TUTS/Qin,
  title={Towards Unconstrained End-to-End Text Spotting},
  author={Siyang Qin and A. Bissacco and Michalis Raptis and Yasuhisa Fujii and Ying Xiao},
  journal={ICCV},
  year={2019},
  pages={4703-4713}
}

@article{CVPR17/CIAS/He,
  title={Multi-scale FCN with Cascaded Instance Aware Segmentation for Arbitrary Oriented Word Spotting in the Wild},
  author={Dafang He and X. Yang and Chen Liang and Zihan Zhou and Alexander Ororbia and Daniel Kifer and C. Lee Giles},
  journal={CVPR},
  year={2017},
  pages={474-483}
}

@inproceedings{AAAI18/FEN/Zhang,
  title={Feature Enhancement Network: A Refined Scene Text Detector},
  author={Sheng Zhang and Yuliang Liu and Lianwen Jin and Canjie Luo},
  booktitle={AAAI},
  year={2018}
}

@inproceedings{CVPR18/cornerRS/Yao,
  title={Multi-oriented Scene Text Detection via Corner Localization and Region Segmentation},
  author={Pengyuan Lyu and Cong Yao and Wenhao Wu and Shuicheng Yan and Xiang Bai},
  booktitle={CVPR},
  year={2018},
  pages={7553-7563}
}

@inproceedings{CVPR18/MCN/Liu,
  title={Learning Markov Clustering Networks for Scene Text Detection},
  author={Zichuan Liu and Guosheng Lin and Sheng Yang and Jiashi Feng and Weisi Lin and Wang Ling Goh},
  booktitle={CVPR},
  year={2018},
  pages={6936-6944}
}

@article{TCSVT20/PSS/Cheng,
  title={A Direct Regression Scene Text Detector With Position-Sensitive Segmentation},
  author={Peirui Cheng and Yuanqiang Cai and Weiqiang Wang},
  journal={IEEE Trans. Circuit Syst. Video Technol.},
  year={2020},
  volume={30},
  pages={4171-4181}
}

@article{TIP20/HAM/Hou,
  title={HAM: Hidden Anchor Mechanism for Scene Text Detection},
  author={Jie-Bo Hou and Xiaobin Zhu and Chang Liu and Kekai Sheng and Long-Huang Wu and Hongfa Wang and Xu-Cheng Yin},
  journal={{IEEE} Trans. Image Process.},
  year={2020},
  volume={29},
  pages={7904-7916}
}

@article{TCSVT21/SLN/Cai,
  title={Scale-Residual Learning Network for Scene Text Detection},
  author={Yuanqiang Cai and Chang Liu and Peirui Cheng and Dawei Du and Libo Zhang and Weiqiang Wang and Qixiang Ye},
  journal={IEEE Trans. Circuit Syst. Video Technol.},
  year={2021},
  volume={31},
  pages={2725-2738}
}

@inproceedings{MM22/SRTS/Wu,
  title={Decoupling Recognition from Detection: Single Shot Self-Reliant Scene Text Spotter},
  author={Jingjing Wu and Pengyuan Lyu and Guangming Lu and Chengquan Zhang and Kun Yao and Wenjie Pei},
  booktitle={ACM-MM},
  pages        = {1319--1328},
  year={2022}
}

@article{MM22/TPSNet/Wang,
  title={TPSNet: Reverse Thinking of Thin Plate Splines for Arbitrary Shape Scene Text Representation},
  author={Wei Wang and Yu ZHOU and Jiahao Lv and Dayan Wu and Guoqing Zhao and Ning Jiang and Weiping Wang},
  journal={ACM-MM},
  year={2022}
}

@article{TMM22/TPF/Yang,
  title={Text-Pass Filter: An Efficient Scene Text Detector},
  author={Chuan Yang and Haozhao Ma and Xu Han and Yuan Yuan and Qi Wang},
  journal={ArXiv},
  year={2026},
  volume={abs/2601.18098}
}

@article{TCSVT23/LOAD/Cheng,
  title={Detect Arbitrary-Shaped Text via Adaptive Thresholding and Localization Quality Estimation},
  author={Peirui Cheng and Yuzhong Zhao and Weiqiang Wang},
  journal={IEEE Trans. Circuit Syst. Video Technol.},
  year={2023},
  volume={33},
  pages={7480-7490}
}

@article{TCSVT24/TPPAN/Xu,
  title={Text Position-Aware Pixel Aggregation Network With Adaptive Gaussian Threshold: Detecting Text in the Wild},
  author={Jiayu Xu and Ailiang Lin and Jinxing Li and Guangming Lu},
  journal={IEEE Trans. Circuit Syst. Video Technol.},
  year={2024},
  volume={34},
  pages={286-298}
}

@ARTICLE{TPAMI/TCM/Yu,
  author={Yu, Wenwen and Liu, Yuliang and Zhu, Xingkui and Cao, Haoyu and Sun, Xing and Bai, Xiang},
  journal={{IEEE} Trans. Pattern Anal. Mach. Intell.}, 
  title={Turning a CLIP Model Into a Scene Text Spotter}, 
  year={2024},
  volume={46},
  number={9},
  pages={6040-6054}
  }

@article{TCSVT25/CIC/Han,
  title={Compensating for the Incomplete With the Complete: An Efficient Scene Text Detector},
  author={Xu Han and Qi Wang},
  journal={IEEE Trans. Circuit Syst. Video Technol.},
  year={2025},
  volume={35},
  pages={12096-12108}
}

@article{TIP25/MTD/Han,
  title={Pull Pole Points to Text Contour by Magnetism: A Real-Time Scene Text Detector},
  author={Xu Han and Chuan Yang and Qi Wang},
  journal={{IEEE} Trans. Image Process.},
  year={2025},
  volume={34},
  pages={6374-6385}
}

@ARTICLE{TMM25/MTPT/Xie,
  author={Xie, Hongtao and Wang, Keran and Zhou, Bangbang and Wang, Yuxin and Qi, Weigang and Wu, Jie and Qu, Yadong and Gao, Zuan and Zhang, Dongming and Liu, Yizhi},
  journal={{IEEE} Trans. Multimedia}, 
  title={Masked Text Pre-Training for Scene Text Detection}, 
  year={2025},
  volume={27},
  number={},
  pages={9429-9443}
  }

@article{PR22/DText/Cai,
  title={Arbitrarily shaped scene text detection with dynamic convolution},
  author={Ying Cai and Yuliang Liu and Chunhua Shen and Lianwen Jin and Yidong Li and Daji Ergu},
  journal={Pattern Recognit.},
  year={2022},
  volume={127},
  pages={108608}
}

@article{IJCA15/TextLocSurvey/Chavre,
  title={A Survey on Text Localization Method in Natural Scene Image},
  author={Pooja Chavre and Archana Ghotkar},
  journal={International Journal of Computer Applications},
  year={2015},
  volume={112},
  pages={15-19}
}

@inproceedings{MLICOM16/TextDetSurvey/Wang,
  title={Text Detection in Natural Scene Image: A Survey},
  author={Shupeng Wang and Chenglin Fu and Qi Li},
  booktitle={MLICOM},
  year={2016}
}

@article{AIR21/DeepTextDetReview/Khan,
  title={Deep learning approaches to scene text detection: a comprehensive review},
  author={Tauseef Khan and Ram Sarkar and Ayatullah Faruk Mollah},
  journal={Artificial Intelligence Review},
  year={2021},
  pages={1-60}
}

@article{MTAP24/MSERDetSurvey/Dutta,
  title={Natural scene text localization and detection using MSER and its variants: a comprehensive survey},
  author={Kalpita Dutta and Ritesh Sarkhel and Mahantapas Kundu and Mita Nasipuri and N. Das},
  journal={Multimedia Tools and Applications},
  year={2023},
  volume={83},
  pages={55773 - 55810}
}

@article{CSUR21/TextRecWild/Chen,
  title={Text Detection and Recognition in the Wild: A Review},
  author={Z. Raisi and Mohamed A. Naiel and Paul W. Fieguth and Steven Wardell and John S. Zelek},
  journal={ArXiv},
  year={2020},
  volume={abs/2006.04305}
}

@InProceedings{SIST26/SegFreeTextRec/Deepak,
  author="Deepak, C. R.
  and Padmavathi, S.",
  title="A Review of Deep Learning-Based Segmentation-Free Scene Text Recognition Methods",
  booktitle={SIST},
  year="2026",
  pages="417--433",
}

@article{TPAMI15/TextDetRecSurvey/Ye,
  title={Text Detection and Recognition in Imagery: A Survey},
  author={Qixiang Ye and David S. Doermann},
  journal={{IEEE} Trans. Pattern Anal. Mach. Intell.},
  year={2015},
  volume={37},
  pages={1480-1500}
}

@article{FCS16/TextDetRecAdv/Zhu,
  title={Scene text detection and recognition: recent advances and future trends},
  author={Yingying Zhu and Cong Yao and Xiang Bai},
  journal={Frontiers of Computer Science},
  year={2015},
  volume={10},
  pages={19 - 36}
}

@article{IJCV21/TextDetRecDeep/Long,
  title={Text Detection, Tracking and Recognition in Video: A Comprehensive Survey},
  author={Xu-Cheng Yin and Ze-Yu Zuo and Shu Tian and Chenglin Liu},
  journal={{IEEE} Trans. Image Process.},
  year={2016},
  volume={25},
  pages={2752-2773}
}

@article{TIP16/TextDetRecVideo/Author,
  title={Scene Text Detection and Recognition: The Deep Learning Era},
  author={Shangbang Long and Xin He and Cong Yao},
  journal={International Journal of Computer Vision},
  year={2018},
  volume={129},
  pages={161 - 184}
}

@article{JFCST24/NaturalTextRecDL/Fanzhi,
  title={Survey on Natural Scene Text Recognition Methods of Deep Learning},
  author={Zeng, Fanzhi and Feng, Wenjie and Zhou, Yan},
  journal={Journal of Frontiers of Computer Science and Technology},
  volume={18},
  number={5},
  pages={1160--1181},
  year={2024}
}

@inproceedings{CVPR15/FCN/Shelhamer,
  title={Fully convolutional networks for semantic segmentation},
  author={Evan Shelhamer and Jonathan Long and Trevor Darrell},
  booktitle={CVPR},
  year={2014},
  pages={3431-3440}
}

@inproceedings{IJCAI18/IncepText/Yang,
  title={Incep{T}ext: A New Inception-Text Module with Deformable PSROI Pooling for Multi-Oriented Scene Text Detection},
  author={Yang, Qiangpeng and Cheng, Mengli and Zhou, Wenmeng and Chen, Yan and Qiu, Minghui and Lin, Wei},
  booktitle={IJCAI},
  pages        = {1071--1077},
  year={2018}
}

@inproceedings{CVPR17/FCIS/Li,
  title={Fully Convolutional Instance-Aware Semantic Segmentation},
  author={Yi Li and Haozhi Qi and Jifeng Dai and Xiangyang Ji and Yichen Wei},
  booktitle={CVPR},
  year={2016},
  pages={4438-4446}
}

@InProceedings{ECCV20/MaskTextSpotterv3/Liao,
    author={Liao, Minghui and Pang, Guan and Huang, Jing and Hassner, Tal and Bai, Xiang},
    title={Mask TextSpotter v3: Segmentation Proposal Network for Robust Scene Text Spotting},
    booktitle={ECCV},
    year={2020},
    pages={706--722}
}

@article{PR20/QRPN/Wang,
  title={A quadrilateral scene text detector with two-stage network architecture},
  author={Siwei Wang and Yudong Liu and Zheqi He and Yongtao Wang and Zhi Tang},
  journal={Pattern Recognit.},
  year={2020},
  volume={102},
  pages={107230}
}

@article{TMM17/RRPN/Ma,
  title={Arbitrary-Oriented Scene Text Detection via Rotation Proposals},
  author={Jianqi Ma and Weiyuan Shao and Hao Ye and Li Wang and Hong Wang and Yingbin Zheng and X. Xue},
  journal={{IEEE} Trans. Multimedia},
  year={2017},
  volume={20},
  pages={3111-3122}
}

@article{PR24/DIPCYT/Roy,
  title={A novel domain independent scene text localizer},
  author={Ayush Roy and Palaiahnakote Shivakumara and Umapada Pal and Cheng-Lin Liu},
  journal={Pattern Recognit.},
  year={2024},
  volume={158},
  pages={111015}
}

@ARTICLE{TIP22/Boundary-TextSpotter/Lu,
  author={Lu, Pu and Wang, Hao and Zhu, Shenggao and Wang, Jing and Bai, Xiang and Liu, Wenyu},
  journal={{IEEE} Trans. Image Process.}, 
  title={Boundary TextSpotter: Toward Arbitrary-Shaped Scene Text Spotting}, 
  year={2022},
  volume={31},
  number={},
  pages={6200-6212}
  }

@ARTICLE{TMM21/R-Net/Wang,
  author={Wang, Yuxin and Xie, Hongtao and Zha, Zhengjun and Tian, Youliang and Fu, Zilong and Zhang, Yongdong},
  journal={{IEEE} Trans. Multimedia}, 
  title={R-Net: A Relationship Network for Efficient and Accurate Scene Text Detection}, 
  year={2021},
  volume={23},
  number={},
  pages={1316-1329}
 }

@article{TIP18/DDR++/He,
  title={Multi-Oriented and Multi-Lingual Scene Text Detection With Direct Regression},
  author={Wenhao He and Xu-Yao Zhang and Fei Yin and Chenglin Liu},
  journal={{IEEE} Trans. Image Process.},
  year={2018},
  volume={27},
  pages={5406-5419}
}

@inproceedings{NeurIPS21/CentripetalText/Sheng,
  title={CentripetalText: An Efficient Text Instance Representation for Scene Text Detection},
  author={Tao Sheng and Jie Chen and Zhouhui Lian},
  booktitle={NeurIPS},
  pages={335--346},
  year={2021}
}

@article{TMM21/LEMNet/Xing,
  title={Boundary-Aware Arbitrary-Shaped Scene Text Detector With Learnable Embedding Network},
  author={Mengting Xing and Hongtao Xie and Qingfeng Tan and Shancheng Fang and Yuxin Wang and Zhengjun Zha and Yongdong Zhang},
  journal={{IEEE} Trans. Multimedia},
  year={2021},
  volume={24},
  pages={3129-3143}
}

@article{TMM26/TextDCTv2/Pan,
  title={TextDCTv2: Scene Text Detection with Patch-Based Discrete Cosine Transform Representation},
  author={Zhaolong Pan and Yuchen Su and Lina Tan and Peng Gao and Wei Wang and Zhineng Chen},
  journal={{IEEE} Trans. Multimedia},
  year={2026}
}

@article{TCSVT25/EdgeText/Yang,
  title={Edge Approximation Text Detector},
  author={Chuan Yang and Xue Han and Tao Han and Han Han and Bingxuan Zhao and Qi Wang},
  journal={IEEE Trans. Circuit Syst. Video Technol.},
  year={2025},
  volume={35},
  pages={9234-9245}
}

@article{IVC04/MSER/Matas,
  title={Robust wide-baseline stereo from maximally stable extremal regions},
  author={Matas, Jiri and Chum, Ondrej and Urban, Martin and Pajdla, Tom{\'a}s},
  journal={Image Vis. Comput.},
  volume={22},
  number={10},
  pages={761--767},
  year={2004},
  publisher={Elsevier}
}

@inproceedings{CVPR15/FASText/Busta,
  title={{FASText}: Efficient Unconstrained Scene Text Detector},
  author={Michal Busta and Luk{\'a}s Neumann and Jiri Matas},
  booktitle={ICCV},
  year={2015},
  pages={1206-1214}
}

@ARTICLE{TMM21/OPMP/Zhang,
  author={Zhang, Sheng and Liu, Yuliang and Jin, Lianwen and Wei, Zhongrong and Shen, Chunhua},
  journal={{IEEE} Trans. Multimedia}, 
  title={OPMP: An Omnidirectional Pyramid Mask Proposal Network for Arbitrary-Shape Scene Text Detection}, 
  year={2021},
  volume={23},
  number={},
  pages={454-467}
  }

@article{CVPR21/MOST/He,
  title={{MOST}: A Multi-Oriented Scene Text Detector with Localization Refinement},
  author={Minghang He and Minghui Liao and Zhibo Yang and Humen Zhong and Jun Tang and Wenqing Cheng and Cong Yao and Yongpan Wang and Xiang Bai},
  journal={CVPR},
  year={2021},
  pages={8809-8818}
}

@article{TMM26/TextRSR/Shao,
  title={TextRSR: Enhanced Arbitrary-Shaped Scene Text Representation via Robust Subspace Recovery},
  author={Zhiwen Shao and Shengtian Jiang and Hancheng Zhu and Xuehuai Shi and Canlin Li and Lizhuang Ma and D. Y. Yeung},
  journal={{IEEE} Trans. Multimedia},
  year={2026},
  volume={28},
  pages={2550-2563}
}

@ARTICLE{TNNLS25/S3INet/Wang,
  author={Wang, Runmin and Chen, Hua and Zhu, Yanbin and Xu, Juan and Cao, Xiaofei and Zhu, Zhenlin and Qian, Shengyou and Gao, Changxin and Liu, Li and Sang, Nong},
  journal={IEEE Trans. on Neural Net. and Learn. Syst.}, 
  title={{S3INet}: Semantic-Information Space Sharing Interaction Network for Arbitrary Shape Text Detection}, 
  year={2025},
  volume={36},
  number={8},
  pages={14691-14705}
  }

@INPROCEEDINGS{CVPR17/ResNet50-FPN/Lin,
  author={Lin, Tsung-Yi and Dollár, Piotr and Girshick, Ross and He, Kaiming and Hariharan, Bharath and Belongie, Serge},
  booktitle={CVPR}, 
  title={Feature Pyramid Networks for Object Detection}, 
  year={2017},
  volume={},
  number={},
  pages={936-944}
  }

@ARTICLE{TPAMI26/handwritten,
  author={Garrido-Munoz, Carlos and Rios-Vila, Antonio and Calvo-Zaragoza, Jorge},
  journal={{IEEE} Trans. Pattern Anal. Mach. Intell.}, 
  title={Handwritten Text Recognition: A Survey}, 
  year={2026},
  volume={48},
  number={4},
  pages={4367-4387},
  }

\vfill

\end{document}